\documentclass[twoside,11pt]{article}

\usepackage{jmlr2e}
\usepackage{mathtools}
\usepackage{amsmath}
\usepackage{amssymb}
\usepackage{sansmath}
\usepackage{amsfonts}
\usepackage{float}
\usepackage{subcaption}

\usepackage{tabularx}
\usepackage{rotating}
\usepackage{booktabs}
\usepackage{multirow}
\usepackage{makecell}
\newcommand{\sfw}{\textsf{W}}
\newcommand{\sfnw}{\textsf{NW}}

\newcommand\norm[1]{\left\lVert#1\right\rVert}

\usepackage{xcolor}

\newcommand{\R}{\mathbb{R}}
\newcommand{\La}{\mathcal{L}}
\newcommand{\D}{\mathcal{D}}

\newcommand{\vvvPBoW}{\mathbf{v^{PBoW}}}
\newcommand{\vPBoW}{v^{PBoW}}
\newcommand{\vvvWPBoW}{\mathbf{v^{wPBoW}}}
\newcommand{\vWPBoW}{v^{wPBoW}}
\newcommand{\vvvSPBoW}{\mathbf{v^{sPBoW}}}
\newcommand{\vSPBoW}{v^{sPBoW}}
\newcommand{\vvvPVLAD}{\mathbf{v^{PVLAD}}}
\newcommand{\vPVLAD}{v^{PVLAD}}
\newcommand{\vvvSPVLAD}{\mathbf{v^{sPVLAD}}}
\newcommand{\vSPVLAD}{v^{sPVLAD}}
\newcommand{\vvvPFV}{\mathbf{v^{PFV}}}

\ShortHeadings{Persistence Codebooks}{Zieli\'nski et al.}
\firstpageno{1}

\begin{document}

\title{Persistence Codebooks for Topological Data Analysis}

\author{\name Bartosz Zieli\'nski \email bartosz.zielinski@uj.edu.pl\\
\name Micha\l{} Lipi\'nski \email michal.lipinski@uj.edu.pl\\
\name Mateusz Juda \email mateusz.juda@uj.edu.pl\\
\addr Institute of Computer Science and Computer Mathematics\\
Faculty of Mathematics and Computer Science\\
Jagiellonian University\\
\L{}ojasiewicza 6, 30-348 Krak\'ow, Poland
\AND
\name Matthias Zeppelzauer \email m.zeppelzauer@fhstp.ac.at\\
\addr Media Computing Group, Institute of Creative Media Technologies\\
St. P\"{o}lten University of Applied Sciences\\
Matthias Corvinus-Strasse 15, 3100 St. P\"{o}lten, Austria
\AND
\name Pawe\l{} D\l{}otko \email p.t.dlotko@swansea.ac.uk\\
\addr Department of Mathematics and Swansea Academy of Advanced Computing\\
Swansea University,\\
Fabian Way SA1 8EP, Swansea, UK\\
}

\editor{TODO}

\maketitle

\begin{abstract}
Persistent homology (PH) is a rigorous mathematical theory that provides a robust descriptor of data in the form of persistence diagrams (PDs) which are 2D multisets of points. Their variable size makes them, however, difficult to combine with typical machine learning workflows. 
In this paper we introduce persistence codebooks, a novel expressive and discriminative fixed-size vectorized representation of PDs. To this end, we adapt bag-of-words (BoW), vectors of locally aggregated descriptors (VLAD) and Fischer vectors (FV) for the quantization of PDs. Persistence codebooks represent PDs in a convenient way for machine learning and statistical analysis and have a number of favorable practical and theoretical properties including 1-Wasserstein stability.
We evaluate the presented representations on several heterogeneous datasets and show their (high) discriminative power. Our approach achieves state-of-the-art performance and beyond in much less time than alternative approaches. 
\end{abstract}


\section{Introduction}
\label{intro}

Topological data analysis (TDA) provides a powerful framework for the structural analysis of high-dimensional data. An important tool in TDA is persistent homology, PH~\citep{EdLeZo2002}.
It provides a comprehensive, multiscale summary of the underlying data's shape and currently gains an increasing importance in data science~\citep{FerriSurvey2017}. Recently, it has been successfully applied to computer vision problems, such as shape and texture analysis~\citep{li2014persistence,reininghaus2014stable}, 3D surface analysis~\citep{adams2017persistence,zeppelzauer2017study}, 3D shape matching~\citep{carriere2015stable}, mesh segmentation~\citep{SkrabaSegmentation2010}, and motion analysis~\citep{VejdemoJohansson2015}. 
Further application areas include time series analysis~\citep{seversky2016time}, music tagging~\citep{liu2016applying} and social-network analysis~\citep{hofer2017deep} as well as applications from the bio-medical domain, e.g. biomolecular analysis~\citep{cang2017topologynet}, brain network analysis~\citep{lee2012persistent}, protein investigation~\citep{gameiro2015topological} and material science~\citep{nakamura2015persistent}.

Persistent homology can be efficiently computed using various currently available tools~\citep{bauer2017phat,chen2011persistent,de2011dualities,DBLP:conf/esa/DeySW16,edelsbrunner2010computational,maria2014gudhi}. A basic introduction to PH is given in Section~\ref{background_and_rw} and the more detailed one in the Appendix.
The common representation of PH are \emph{persistence diagrams} (PDs) which are multisets of points in $\mathbb{R}^2$. Due to their variable size, which varies depending on the input data, PDs are not easy to integrate within common data analysis, statistics and machine learning workflows. To alleviate this problem, a number of kernel functions defined on PDs and vectorization methods for PDs have been introduced. 

Kernel-based approaches have a strong theoretical background but in practice they often become inefficient when the number of training samples is large. As typically the entire kernel matrix must be computed explicitly (like in case of SVMs), this leads to roughly quadratic complexity in computation time and memory with respect to the size of the training set. Furthermore, vector-based approaches are limited to kernelized methods, such as SVM and kernel PCA.
Vectorized representations, in contrast, are compatible with a much wider range of methods and do not suffer from complexity constraints of kernels. They, however, often lack in representational power, as they require the spatial quantization of the PDs, which is unsually non-adaptive and thus does not cope well with the sparseness of PDs.

In this work we present a novel adaptive representation of PDs which aims at combining the large representational power of kernel-based approaches with the general applicability of vectorized representations. To this end, we adapt the popular bag-of-words (BoW) encoding~\citep{mccallum1998comparison,sivic2003video}, as well as its more comprehensive extensions, such as VLAD~\citep{jegou2010aggregating} and Fisher vectors~\citep{perronnin2007fisher} to cope with the inherent sparsity of PDs. The proposed persistent codebooks provide universally applicable fixed-sized feature vectors. They are, under mild assumptions, stable with respect to a standard metric in PDs and thus, also theoretically, built upon a solid basis. Experiments show that the new representations achieve state-of-the-art performance and even outperform numerous competitive methods while being more compact and requiring orders of magnitude less time. 

This paper builds upon previous work of \citep{zielinski2018persistence}. The additional contribution includes: (i) two new persistence codebook representations (PVLAD and PFV) building upon vectors of locally aggregated descriptors (VLAD) and Fisher vectors (FV); (ii) the investigation of their stability; (iii) the introduction of stable variant of PVLAD algorithm together with the proofs of its stability; (iv) a significant number of additional  experiments on an extended collection of datasets; and (v) an extended discussion of results.

The paper is structured as follows. Section~\ref{background_and_rw} gives a basic introduction to PH and reviews related approaches. In Section~\ref{approaches} we introduce persistence codebooks and investigate their stability. Sections~\ref{experiments} and \ref{results} present the experimental setup and results. We conclude the~work in Section~\ref{sec:conclusion}.

\section{Background and Related Work}
\label{background_and_rw}

\subsection{Background on Persistent Homology}

In this section, we first introduce persistent homology, and then describe related state-of-the-art approaches, both kernel- and vectorization-based, that aim at making PH compatible with machine learning methods.

\begin{figure*}[ht]
\centering
\includegraphics[width=0.98\linewidth]{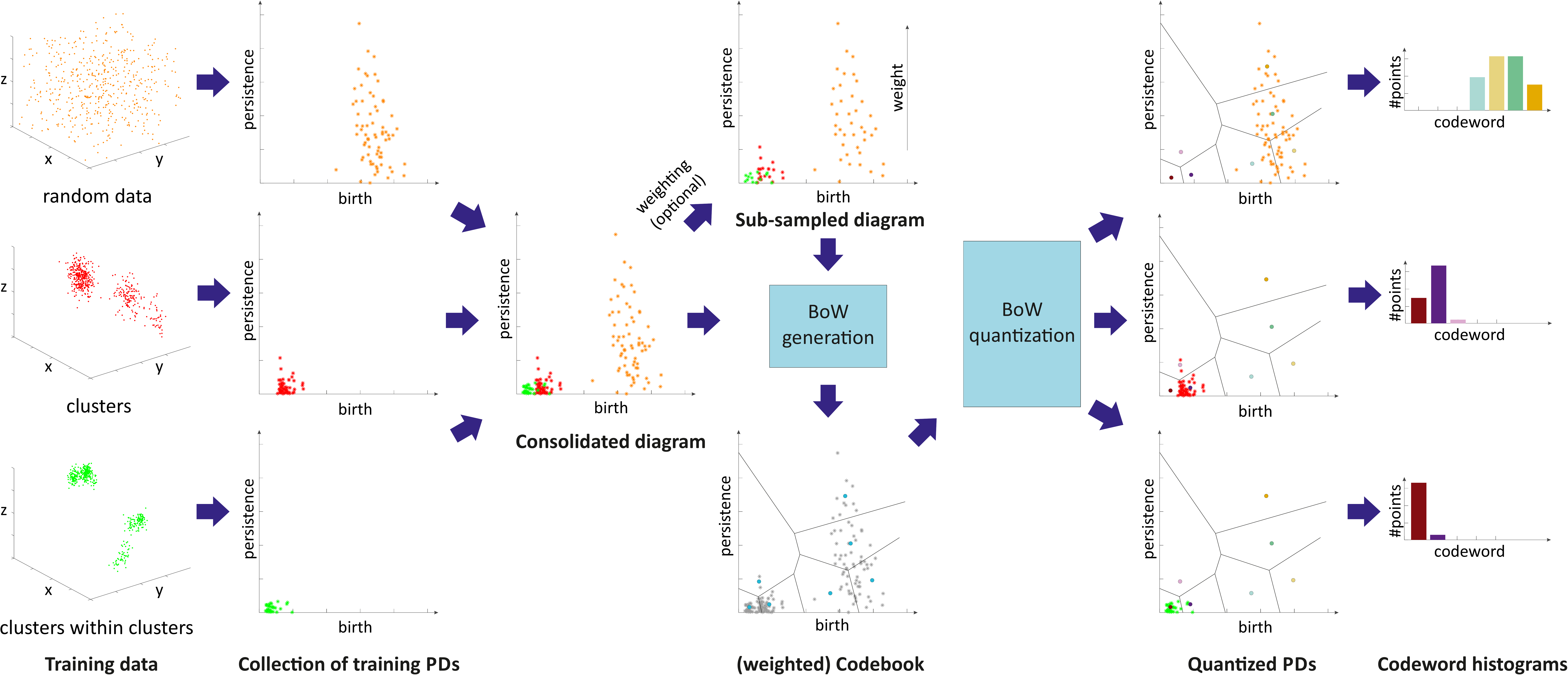}
\caption{The principle behind persistent codebooks on the example of the computational workflow of the persistence bag-of-words representation (PBoW): From the input data we compute PDs in dimension $1$ in birth-persistence coordinates and combine them into one consolidated diagram (for the entire dataset). Next, a subset of points is obtained from this diagram by either a weighted or unweighted sub-sampling. Subsequently, we cluster the~sub-sampled consolidated diagram to retrieve the codewords which will form our codebook. Finally, the points for each input PD are encoded by the codewords (BoW quantization). In this illustration a hard assignment of points to words (PBoW) is performed. The result is a codeword histogram for each input PD that represents how many points fall into which cluster of the codebook, i.e. codeword cardinalities. These codeword histograms are a compact and fixed-size vectorial representation. It is worth mentioning that while the hard assignment presented here gives the idea of the procedure, in practice we often employ soft assignment for stability reasons. Please, note further that the workflows for other persistent codebook encodings (e.g. based on VLAD or Fisher Vectors) are structurally similar, but partly use different codeword generation, quantization, and assignment schemes.}
\label{fig:approachPipeline}
\end{figure*}

Under mild assumptions, persistent homology (PH) can be defined for a continuous function $f :X \rightarrow \mathbb{R}$, where $X \subset R^n$.
Typically $f$ is a distance function from a collection of points, or a scalar value function defined on a grid of points, but in principle it can be an arbitrary function that satisfies a tameness assumption specified below. 
Focusing on sub-level sets $L_x = f^{-1}( (-\infty,x] )$, we let $x$ grow from $-\infty$ to $+\infty$. While this happens, we can observe a whole hierarchy of events. In dimension zero, connected components of $L_x$ will be created and merged together. One dimensional cycles that are not bounded, or higher dimensional voids, will appear in $L_x$ at critical points of $f$. The value of $x$ on which a connected component, cycle or a higher dimensional void appears is refereed to as \emph{birth time}.
They will subsequently either become identical (up to a deformation) to other cycles and voids (created earlier), or they will be glued-in and become trivial. The value of $x$ on which that happens is refereed to as \emph{death time}. 
Every connected component, a cycle, or a~higher dimensional void can, therefore, be characterized by a pair of numbers, $b$ and $d$, its birth and death time. The difference between the death and the birth, $p = d-b$, is the so-called {\em persistence value}. In this paper, we will use the \emph{birth-persistence} pair $[b,p]$ to encode the feature. The multi-set of birth-persistence pairs makes up a \emph{persistence diagram} (PD). The set of all persistence diagrams will be denoted as $\D$. Example PDs for three different input point clouds are shown in Fig.~\ref{fig:approachPipeline}.

The persistence coordinate is often an indicator of whether a cycle is structurally relevant or more likely to be related to noise. This observation is justified by many \emph{stability theorems} for persistence~\citep{EdLeZo2002}, which state that a small change in the space $X$, or in a function $f$, implies only a small change in the resulting persistence diagram. Consequently, points in the PD with low persistence can be removed by a small perturbation of the data; and therefore, are not considered \emph{stable features}. Those stability results make PDs a robust tool in data analysis.  

Throughout this paper we assume that the given function $f$ is \emph{tame}, i.e. it induces a finite number of birth-persistence points. There are various metrics on finite PDs. To define them, the finite diagrams have to be enriched with an infinite collection of points $(b,0)$, which represent features that are born and immediately die. Having the enriched PDs  $B$ and $B'$ let us consider all possible matchings $\eta: B \rightarrow B'$. The 1-Wasserstein distance is defined as:
\[ W_1(B,B') = inf_{\eta : B \rightarrow B'} ||x-\eta (x)||_{\infty} \]
In this paper, when considering stability of the representations, we will consider the stability with respect to 1-Wasserstein distance. A more in-depth introduction to PH is provided in Appendix.

\subsection{Kernels and Vectorized Representations of PDs}

Numerous kernel-based and vectorized approaches have been introduced to make PDs compatible with statistical analysis and machine learning methods. The goal of \textbf{kernel-based approaches} is to define dissimilarity measures (also known as kernel functions) on PDs to compare them, and thereby make them compatible with kernel-based machine learning methods, such as Support Vector Machines (SVMs), and kernel Principal Component Analysis (kPCA). \cite{li2014persistence} combine the traditional bag-of-features (BoF) approach with PDs by using various distance functions between $0$-dimensional PDs (bottleneck and Wasserstein distances for PDs, $L^p$ distance functions for persistence landscapes of PDs) to generate kernels. On different datasets (SHREC 2010, TOSCA, hand gestures, Outex) they show that topological information is complementary to the information of traditional BoF.
\cite{reininghaus2014stable} propose a kernel for persistence diagrams by turning PDs into a continuous distribution by appropriate placement of Gaussian distributions in $\mathbb{R}^2$. Subsequently, they define a kernel as a scalar product of the two corresponding distributions. They apply topological descriptors together with the novel kernel to shape retrieval and~texture classification.
\cite{Carrire2017SlicedWK} propose another kernel based on sliced approximation of the Wasserstein distance. The authors show that the kernel is not only stable, but also mimics bottleneck distances between PDs. They subsequently develop an~approximation technique to reduce the kernel computation time. They apply it to 3D shape segmentation, texture classification, and orbit recognition in dynamical systems. 
Another approach for the representation of PDs are persistence landscapes, PL~\citep{bubenik2015statistical}. PL is a stable functional representation of a PD obtained from transforming it into a sequence of real-valued piece-wise linear functions. To compare two landscapes, the authors use standard $L^p$ distance. This distance can be used to define a kernel function. Note that PL can also be transformed into a fixed-length vectorized representation by sampling the values of the landscape function. The authors, however, are not reporting results for vectorized PLs; therefore, we compare to kernels derived from PLs in our experiments.

\textbf{Vectorized representations} aim at deriving fixed-size encodings of PDs that can be used directly as input to current machine learning methods.
One approach is the persistence image (PI) proposed by \cite{adams2017persistence} which is built upon earlier work on size functions~\citep{Frosini1998,Ferri1997}. It is a vectorized representation of PD, which can be employed directly as a feature vector in conventional machine learning techniques. 
\cite{anirudh2016riemannian} propose an alternative approach based on the reconstruction of a certain Riemannian manifold (RM) based on PDs and its subsequent representation by a fixed-size vector.

Recently, a third type of approach has been introduced, which aims at learning which points in the PD are of particular importance for the given task in a supervised manner~\citep{hofer2017deep}. Despite promising properties this approach is limited to cases where supervisory information (labels) is available. The employed deep learning methodology requires large training sets and considerable training time. A particular challenge of the approach is to find adequate network architectures for the given task and input data type. Kernels and vectorized representations can be used directly off-the-shelf and do not require such design considerations.

\section{Persistence Codebooks}
\label{approaches}
In this section, we adapt the bag-of-words (BoW) model~\citep{mccallum1998comparison,sivic2003video} as well as its more comprehensive extensions, such as VLAD~\citep{jegou2010aggregating} and Fisher vector~\citep{perronnin2007fisher}, introduced originally in text and image retrieval, for adaptive quantization of PDs into a fixed length vectorial representation. The idea behind BoW is to quantize variable length input data into a fixed-size representation by using a common dictionary, also called \textit{codebook} of constant size. The~codebook is generated from the input data in an unsupervised manner by extracting centers of clusters obtained from data clustering. The basic assumption behind BoW is that the clusters (i.e. codewords) capture the intrinsic structure of the data and, thereby represent an efficient vocabulary for the quantization of the data. 

The overall approach of bag-of-words for persistence diagrams is visualized in Fig.~\ref{fig:approachPipeline}. The input is a set of PDs extracted from all instances of a given dataset. First, all PDs are merged into one diagram. This consolidated diagram is then sub-sampled to reduce the influence of noise. In this paper, we consider two types of sub-sampling. A standard one which does not consider the persistence of the points, and one where points of higher persistence are more likely to be sampled, see Section \ref{wf} (we refer to those two types of sub-sampling as \emph{without} and \emph{with weighting}, respectively). From the (sub-sampled) consolidated diagram, the codebook $C$ is generated using clustering. Given a codebook $C$, every input point $P$ is encoded by assigning it to the nearest codeword from $C$. In traditional BoW this encoding leads to a codeword histogram, i.e. a histogram for which each codeword from $C$ counts how many points from $P$ are closest to this codeword. Further encodings investigated include vector of locally aggregated descriptors (VLAD) and Fisher vector (FV), see below.

For the proposed approaches, three important hyperparameters need to be identified: (1) the clustering algorithm used to generate the codebook, (2) the size of the codebook, i.e., the number of clusters, and (3) the type of proximity encoding which is used to obtain the final descriptors, i.e. hard and soft assignment. In this paper, we use k-means and Gaussian mixture models (GMM) for clustering. The codebook size is investigated empirically. In the following sections, we introduce persistent codebook approaches based on different quantizations and encodings, such as standard BoW, VLAD and FV. Consult Table~\ref{tab:boCombinations} for an overview of representations introduced and evaluated in this paper.

\begin{figure}
\begin{center}
\begin{sc}
\begin{tiny}
\begin{tabularx}{0.8\textwidth}{l|l|l|l|l|c}
\toprule
\thead{ \textbf{Sampling}\\\textbf{consolidated}\\\textbf{PD} } & \thead{ \textbf{Codebook}\\ \textbf{generation} } & \thead{ \textbf{Codebook}\\ \textbf{size} } & \thead{ \textbf{Histogram}\\ \textbf{assignment} } & \thead{ \textbf{Abbrev.} } & \thead{ \textbf{Equation} } \\
\hline
\midrule
\multirow{6}{*}{\thead{\tiny no weighting\\ \tiny weighting}} & \multirow{3}{*}{k-means} & \multirow{6}{*}{see Table~\ref{tab:kernelParameters}-\ref{tab:vectorParameters}} & hard & PBoW & (\ref{eq:PBoW}) \\
\cline{4-6}
{} & {} & {} & hard with weights & wPBoW & (\ref{eq:wPBoW}) \\
\cline{4-6}
{} & {} & {} & hard & PVLAD & (\ref{PVLAD}) \\
\cline{2-2}\cline{4-6}
{} & \multirow{3}{*}{GMM} & {} & \multirow{3}{*}{soft} & sPBoW & (\ref{eq:sPBoW}) \\
\cline{5-6}
{} & {} & {} & {} & sPVLAD & (\ref{eq:sPVLAD}) \\
\cline{5-6}
{} & {} & {} & {} & sFV & (\ref{eq:sFV}) \\
\hline
\end{tabularx}
\caption{The design space of persistent codebook approaches introduced in this paper together with their abbreviations for reference. Each resulting representation can use either weighting of no weighting in codebook generation (see experiments in Section \ref{subsec:resultAccuracyvsCodebookSize} for a~direct comparison). For variants with weighting, we add ``-w" to the abbreviation, e.g. ``PBoW-w" for clarity.}
\label{tab:boCombinations}
\end{tiny}
\end{sc}
\end{center}
\end{figure}

For all approaches presented in these sections, we show if they are stable with respect to 1-Wasserstein distance. We would like to indicate that since the representations presented here are additive (consider the definition of additivity from~\citep{reininghaus2014stable}), they are not stable for a $p$-Wasserstein distance for any $p > 1$ as indicated in Theorem~3 in~\citep{reininghaus2014stable}.

\subsection{Persistence Bag of Words (PBoW)}
\label{pbow}

Let us first consider a direct adaptation of BoW~\citep{baeza1999modern,sivic2003video} to PDs. Given a collection of persistence diagrams $B_1 , B_2, \ldots, B_n$, they are consolidated into $D = B_1 \cup B_2 \cup \ldots \cup B_n$ and a codebook of size $N$ is obtained by using $k$-means clustering on $D$. Let $\{\mu_i \in \R^2, i=1,\ldots,N\}$ denote the centers of obtained clusters (the codewords). Moreover, for a PD $B=\{x_t \in \R^2\}_{t=1}^T$, let us denote $NN(x_t)$ as the index of the codeword nearest to $x_t$, $NN(x_t) = i |\ d(x_t,\mu_i) \leq d(x_t,\mu_j)$ for all $j \in \{1,\ldots,N\}$. For every codeword $\mu_i$, $\vPBoW_i(B)=card\{x_t\in B\ |\ NN(x_t)=i\}$ captures the number of points from $B$, which are closer to $\mu_i$ than to any other $\mu_j$. Then the \emph{persistence bag of words} (PBoW) is defined as a vector:
\begin{equation}
\vvvPBoW(B) = \left(\vPBoW_i(B)\right)_{i=1,\ldots,N},
\label{eq:PBoW}
\end{equation}

Subsequently, $\vvvPBoW(B)$ is normalized by taking the square root of each component (preserving the initial sign) and dividing it by the norm of the whole vector. This is a~standard normalization for BoW~\citep{perronnin2010large}, which reduces the influence of outliers.

\medskip
\noindent {\bf Remark} Let $B,B'\in\D$ be persistence diagrams containing only finitely many off-diagonal points. The persistence bag of words, PBoW with $N$ words is \textit{not stable} with respect to \mbox{1-Wasserstein} distance.

\medskip
\noindent
{\bf Proof.}
Let us assume that we have two clusters with centers $\mu_1=(0,0),\mu_2 = (1,0)\in\R^2$, and PD $B$ containing only one point $x_1=(\frac{1}{2}+\epsilon,0)$, for some small $\epsilon>0$. 
Then, $\vvvPBoW(B)=[0, 1]$, because $x_1$ is closer to $\mu_2$ than $\mu_1$. However, a small perturbation in $B$, e.g. by $-2\epsilon$, changes the assignment of $x_1$ from $\mu_2$ to $\mu_1$. 
In this case $B'=\{ y_1=(\frac{1}{2}-\epsilon,0) \}$ and $\vvvPBoW(B')=[1, 0]$. 
In order to be stable in 1-Wasserstein sense, PBoW should fulfill the following condition:
\[
2=|\vvvPBoW(B)-\vvvPBoW(B'))|<C|x_1-y_1|<2C\epsilon,
\]
therefore $C>1/\epsilon$. As $\epsilon>0$ can be arbitrarily small, there does not exist a constant $C$ that meets this condition. Hence, the direct adaption of BoW to PDs (PBoW) is not stable.
\begin{flushright} $\square$\end{flushright}

\subsection{Weighted Subsampling for Codebook Generation}
\label{wf}

Aside from being unstable, the straight-forward application of BoW to PD would neglect an~important property of persistence diagrams, i.e. that points in a PD with higher persistence are typically considered more important than points with lower persistence. It is a consequence of a stability theorem~\cite{edelsbrunner2010computational} indicating that points with low persistence are more likely to originate from a noise than the points of high persistence. 

In order to integrate this property into the codebook generation procedure, we perform $k$-means clustering on a subset of points obtained by a \emph{weighted} sampling described below. This results in extended procedure of codebook generation which is as follows:
\begin{enumerate}
\item Place all the persistence diagrams (or all diagrams of a certain dimension) on to a~single consolidated persistence diagram $D$.
\item Subsample $n$ points from $D$ in a way that points of higher persistence are more likely to be sampled. In the experiments presented in this paper we set it to $n=10000$\footnote{Preliminary experiments have shown that this number is insensitive and has little influence on the results (evaluated value range: 1000 to 100000).}.
\item Perform $k$-means clustering on the obtained subset of $n$ points to extract the centers of the clusters (the codewords). 
\end{enumerate}

For the weighted sampling of points from a persistence diagram we define a piecewise linear weighting function $w_{a,b}: \R \rightarrow \R$ as:
\begin{equation}
w_{a,b}(t) = \left\{ \begin{array}{ll}
0 & \textrm{if $t < a$}\\
(t-a)/(b-a) & \textrm{if $a \leq t < b$}\\
1 & \textrm{if $t \geq b$}
\end{array} \right. ,
\label{eq:weighting_function}
\end{equation}
and use it to weight second coordinates (persistence) of points in PD. In our experiments we set $a$ and $b$ to the persistence values corresponding to $0.05$ and $0.95$ quantiles of the persistence coordinate of the points in $D$. In the performed sub-sampling persistence points having longer values of the function $w$ are more likely to be sampled. 

We want to highlight that in this case we have selected a \emph{linear weighting} with respect to persistence, i.e. the probability of sampling of a point is \emph{proportional to its persistence}. It works well in the cases considered in this paper, however in the case of very noisy data with just a few dominant persistent points, the points of high persistence may not be sampled at all. In such case, we suggest to consider the weighting to be a higher degree polynomial or an exponential function to boost the probability of capturing the high persistence points. 

\begin{figure}[ht]
\begin{center}
\includegraphics[width=0.35\columnwidth]{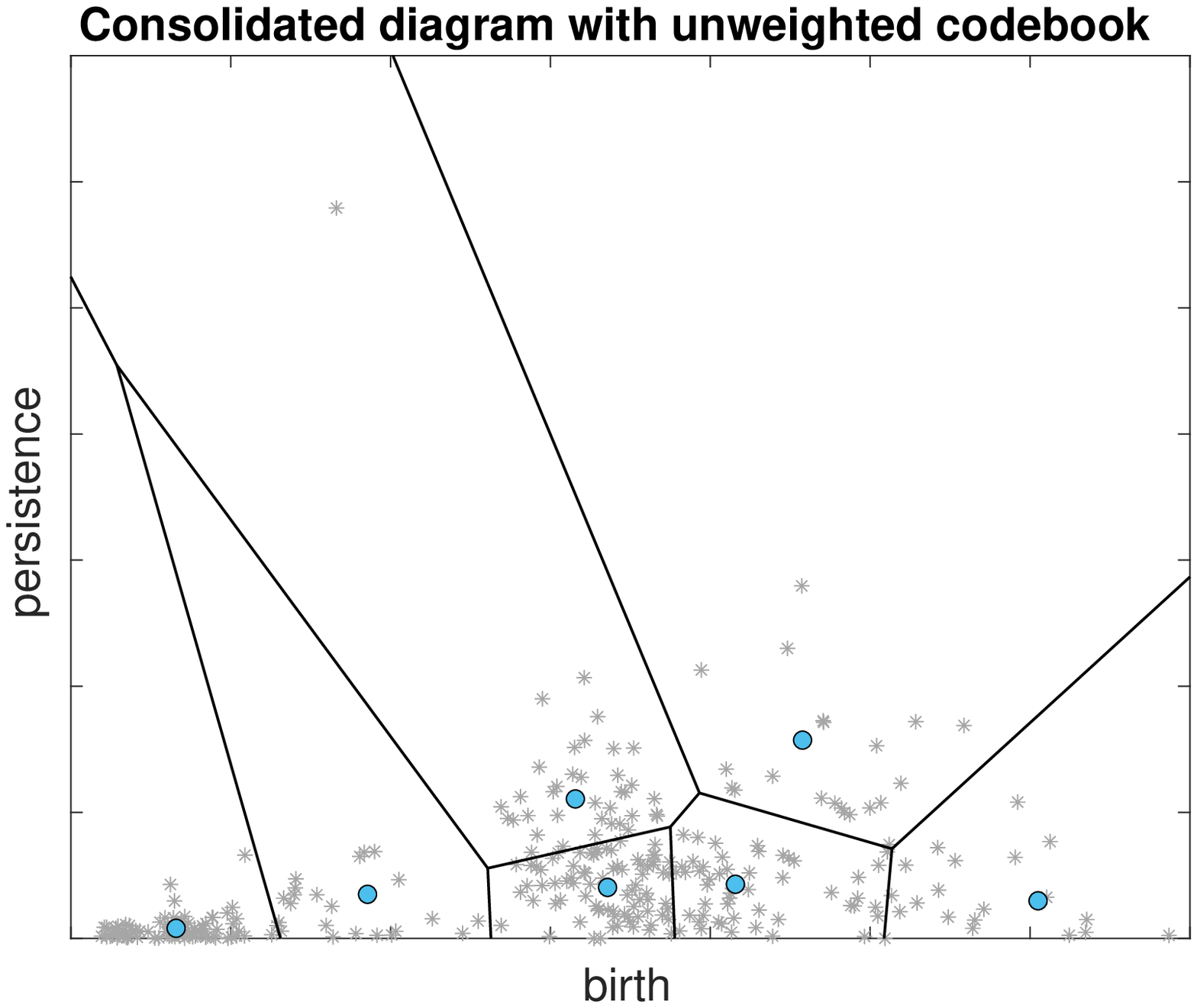}
\includegraphics[width=0.35\columnwidth]{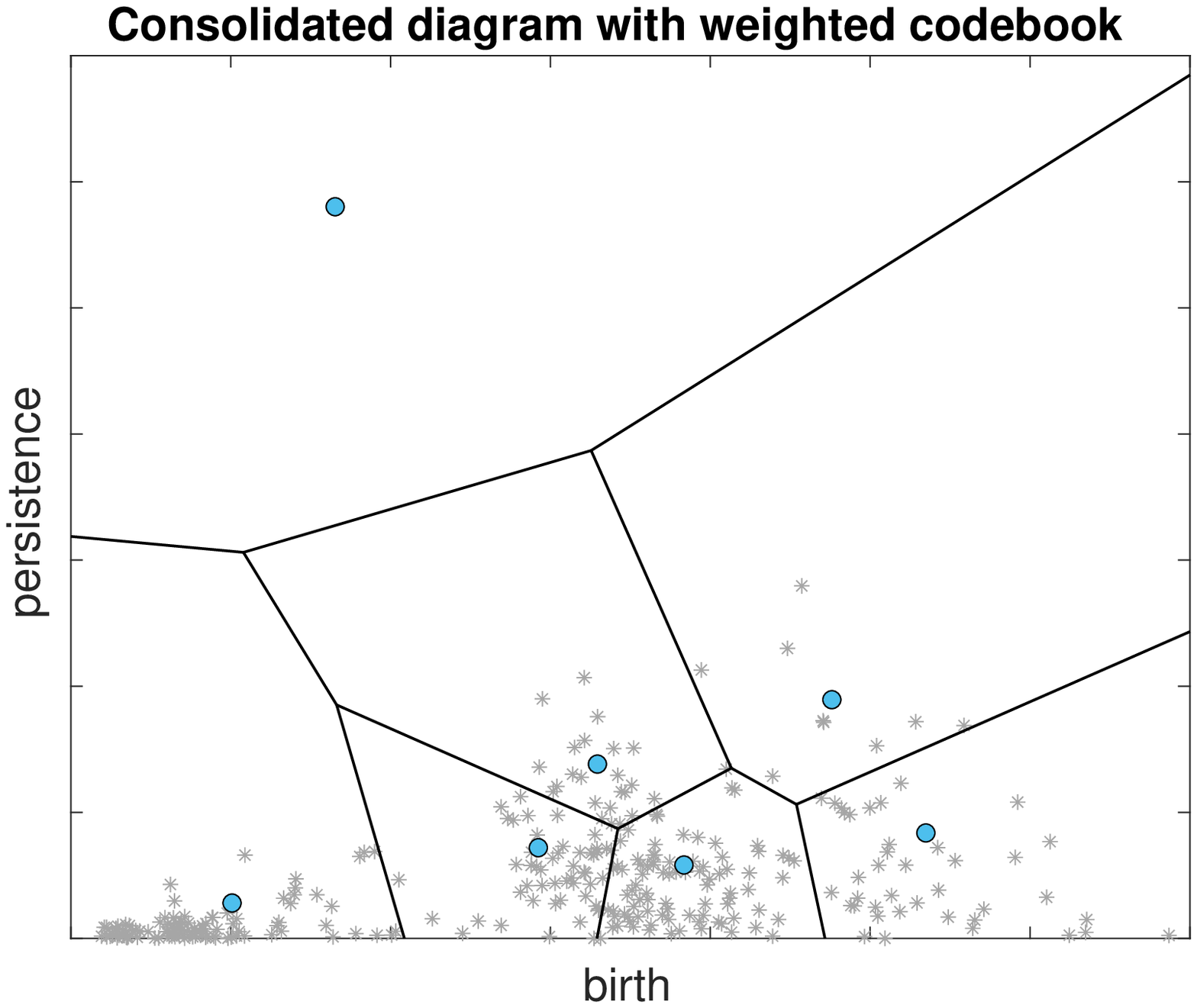}\\
\includegraphics[width=0.35\columnwidth]{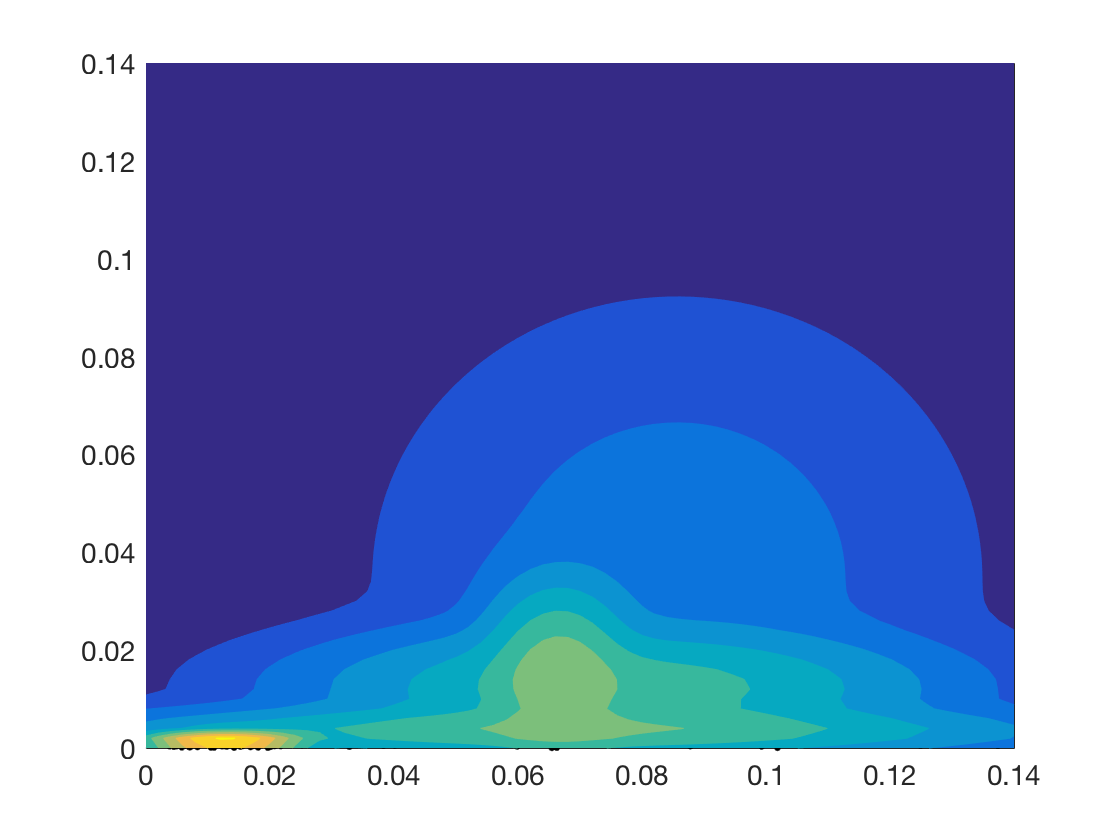}
\includegraphics[width=0.35\columnwidth]{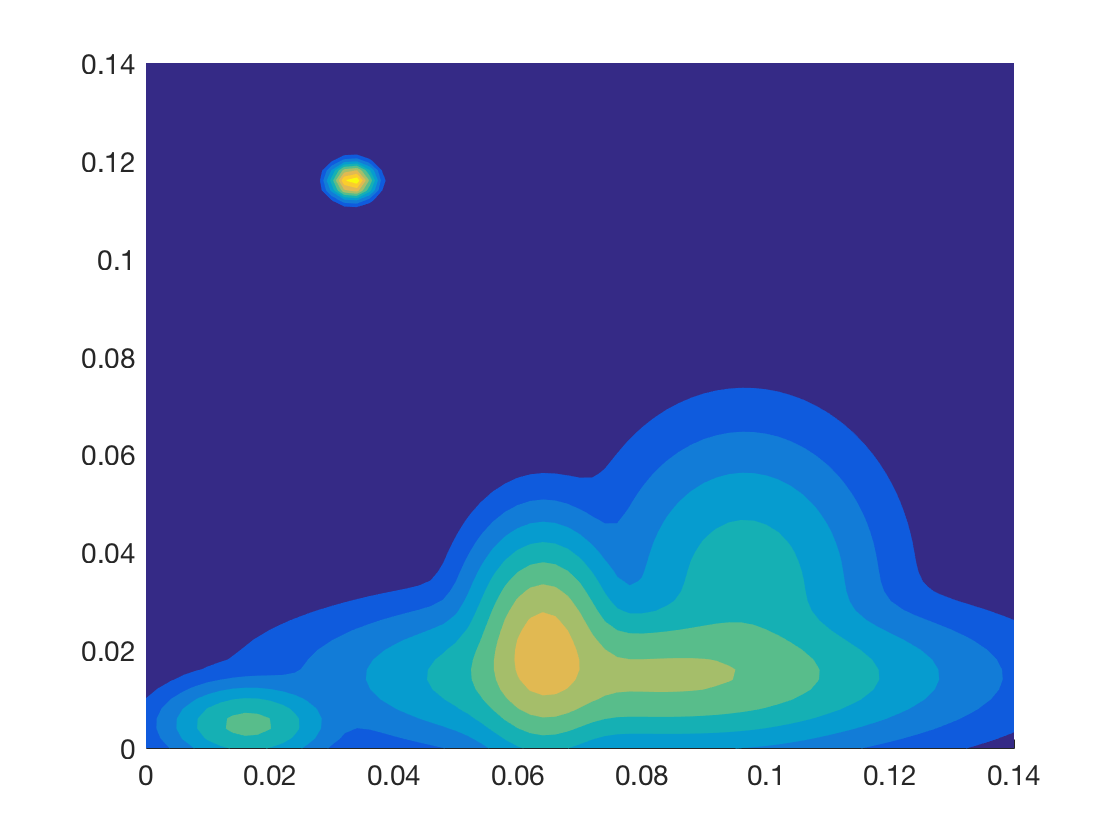}

\caption{Codebook generation based on the consolidated PD with $N=7$ codewords (top: k-means, bottom: GMM, left: no weighting, right: weighting). Using the language of computational geometry; one may tell that the cells in the top diagrams form a Voronoi diagram of the codewords. Equivalently all points in the same Voronoi cell have the same closest codeword. 
Weighting allows to sample more points of higher persistent from the~diagram, representing more stable topological structures and yields more balanced cluster weights (even though there are many more points on the bottom of consolidated PD). Note that a~cluster can (but does not have to) be created for a single high persistence point which is well separated from the others, as is the case here with the most persistent point (top-left quadrant).}
\label{fig:bowClusters}
\end{center}
\end{figure}

Please, note further that the sub-sampling does not directly enforce the points of the highest persistence to be automatically selected as the centers of clusters, but it makes the~probability of such an event considerably larger. Examples of birth-persistence distributions with standard (unweighted) and weighted codebooks obtained with k-means and GMM are presented in~Fig.~\ref{fig:bowClusters}. The unweighted clustering produces larger clusters, which are less adaptive to the strongest topological structures. At the same time, the weighted clustering yields a more adaptive codebook with a more uniform sampling of the space. 

\subsection{Weighted Codeword Assignment for Persistence Bag of Words (wPBoW)}
\label{wpbow}

The weighting function from Section~\ref{wf} can be similarly used to weight the histogram assignments to give points with higher persistence more influence in the final representation. For this purpose, instead of counting the number of points, we sum up the weights of their persistent coordinates. 
\begin{equation}
\vvvWPBoW(B)=\left(\vWPBoW_i=\sum_{(b,p)\in B :NN((b,p))=i}w_{a,b}(p)\right)_{i=1..N},
\label{eq:wPBoW}
\end{equation}
where $B \in \mathcal{D}$. We will refer to this representation as \emph{weighted persistence bag of words} (wPBoW) in the following. Similary to standard PBoW, wPBoW is not stable with respect to 1-Wasserstein distance. The couterexample is identical with the one in Section~\ref{pbow}, when we assume that function $w_{a,b}$ is identity.

\subsection{Stable Persistence Bag of Words (sPBoW)}
\label{spbow}

After having integrated persistence-based weighting into codebook generation and also into histogram assignment, we aim at making the representation stable. To this end we adapt soft assignment of points to clusters and prove that such an approach guarantees stability of the resulting representation. \emph{Stable persistence bag of words} (sPBoW) similarly to PBoW (and wPBoW) first consolidates PDs in the initial step of construction, and then generates a GMM based on the sub-sampled points (e.g. by expectation maximization algorithm~\citep{nasrabadi2007pattern}). This approach was originally introduced by~\cite{van2008kernel}. 

Let the parameters of the fitted GMM be $\lambda=\{w_i,\mu_i,\Sigma_i, i=1,\ldots,N\}$, where $w_i$, $\mu_i$ and $\Sigma_i$ denote the weight, mean vector and covariance matrix of $i$-th Gaussian, and $N$ denotes the number of Gaussians. Given a PD $B$, the stable PBoW is defined as:
\begin{equation}
\vvvSPBoW(B)=\left(\vSPBoW_i=w_i\sum_{x_t\in B}p_i(x_t|\lambda)\right)_{i=1,\ldots,N},
\label{eq:sPBoW}
\end{equation}
where $w_i>0$, $\sum_{i=1}^N w_i=1$, and $p_i(x_t|\lambda)$ is the likelihood that observation $x_t$ was generated by Gaussian $i$: 
\begin{equation}
p_i(x_t|\lambda) = \frac{exp\{-\frac{1}{2}(x_t-\mu_i)'\Sigma_i^{-1}(x_t-\mu_i)\}}{2\pi|\Sigma_i|^{\frac{1}{2}}}.\nonumber
\end{equation}

The intuition behind this approach is to assign each point to \emph{all} codewords, but with weight inversely proportional to the distance to the codewords.

\medskip
\noindent
{\bf Theorem} Let $B$ and $B'$ be persistence diagrams with a finite number of non-diagonal points. Stable persistence bag of words, sPBoW with $N$ words is stable with respect to \mbox{1-Wasserstein} distance between the diagrams, that is
\[
\norm{\vvvSPBoW(B)-\vvvSPBoW(B')}_{\infty}\leq C\cdot W_1(B, B'),
\]
where $C$ is a constant.

\medskip
\noindent
{\bf Proof.} 
Let $\eta : B \rightarrow B'$ be the optimal matching in the definition of 1-Wasserstein distance. For a fixed $i\in\{1,\ldots,N\}$ we have:
\begin{multline}
\norm{\vSPBoW_i(B)-\vSPBoW_i(B')}_\infty
= \norm{w_i\sum_{x \in B}( p_i(x|\lambda)- p_i(\eta(x)|\lambda)}_\infty \\
\begin{aligned}
&\leq |w_i|\sum_{x \in B}\norm{\left(p_i(x|\lambda)-p_i(\eta(x)|\lambda)\right)}_\infty\nonumber
\end{aligned}
\end{multline}
As $p_i: \R^2 \rightarrow \R$ are Lipschitz continuous with the Lipschitz constants $L_i$, we get
\begin{multline}
|w_i|\sum_{x \in B}\norm{\left(p_i(x|\lambda)-p_i(\eta(x)|\lambda)\right)}_\infty
\leq  |w_i|\sum_{x \in B}\norm{\left( L_i(x - \eta(x))\right)}_\infty \\
\begin{aligned}
&= |w_i\ L_i|\sum_{x \in B}\norm{\left( x - \eta(x)\right)}_\infty = |w_i\ L_i|\ W_1(B, B')\nonumber
\end{aligned}
\end{multline}
\begin{flushright} $\square$\end{flushright}

\subsection{Persistence VLAD}
\label{pvlad}
\emph{Persistence VLAD} (PVLAD) is based on vector of locally aggregated descriptors (VLAD) by~\cite{jegou2010aggregating}, an extension of the bag-of-words concept, which accumulates the~residual of each descriptor with respect to its assigned cluster. The first computation step is similar to PBoW: a codebook $\{\mu_i \in \R^2, i=1..N\}$ is obtained from a training set using $k$-means clustering. Given a new PD $B$, each point $x_t \in B$ is associated with its nearest codeword $NN(x_t)$. 
In the second step, for each codeword $\mu_i$, we compute a sum of differences between $\mu_i$ and all $x_t \in B$ for which $NN(x_t) = i$. 
This results in:
\begin{equation}
\vvvPVLAD=\left(\vPVLAD_i=\sum_{x_t:NN(x_t)=i}x_t-\mu_i\right)_{i=1}^N.
\label{PVLAD}
\end{equation}
The dimension of $\vPVLAD_i$ equals $2$ (differences on two coordinates), therefore $\vvvPVLAD$ is of size $2N$. Intuitively, this vector should capture more information than PBoW alone, because it encodes the first order moments of the points assigned to a codeword instead of simply counting those points.

Similarly to PBoW, PVLAD is not stable with respect to 1-Wasserstein distance. Therefore, in Section~\ref{spvlad}, we propose to adapt a stable variant of VLAD, called soft VLAD.

\medskip
\noindent {\bf Remark} Let $B,B'\in\D$ be persistence diagrams containing only finite off-diagonal points. The persistence vector of locally aggregated descriptors, PVLAD with $N$ words is not stable with respect to \mbox{1-Wasserstein} distance.

\medskip
\noindent {\bf Proof.} 
Starting from two clusters with centers $\mu_1=(0,0),\mu_2 = (1,0)\in\R^2$ and a~persistence diagram $B = \{ (\frac{1}{2}+\epsilon,0) \}$, for small $\epsilon>0$, we get $\vPVLAD_1=[0,0]$ and $\vPVLAD_2=[\epsilon-\frac{1}{2},0]$. However, similarly to the case of PBoW, a small perturbation of $B$, e.g. by $[-2\epsilon,0]$ will change $B$ to $B' = \{ (\frac{1}{2}-\epsilon,0) \}$ and the corresponding components of PVLADs to $[\frac{1}{2}-\epsilon,0]$ and $[0,0]$. Calculating the difference between $\vvvPVLAD(B)$ and $\vvvPVLAD(B')$: 
\[
|\vvvPVLAD(B)-\vvvPVLAD(B')| = |[[0,0],[\epsilon-\frac{1}{2},0]]-[[\frac{1}{2}-\epsilon,0],[0,0]]| = 
|[ [\frac{1}{2}-\epsilon,0] , [\epsilon-\frac{1}{2},0] ]| = 1
\]
\noindent In order to be stable in 1-Wasserstein sense, PVLAD should fulfill the following condition: $1=|\vvvPVLAD(B)-\vvvPVLAD(B'))|<C|x_1-y_1|<2C\epsilon$, therefore $C>\frac{1}{2\epsilon}$. As $\epsilon>0$ can be arbitrarily small, and there does not exist a constant $C$ that meets this condition. Therefore, PVLAD is not stable.
\begin{flushright} $\square$\end{flushright}

\subsection{Stable Persistence VLAD}
\label{spvlad}

Similarly to PBoW, the hard association with codewords can be replaced by soft association in VLAD~\citep{jegou2012aggregating}, to account for instability. To this end, we define \emph{stable persistence VLAD} (sPVLAD) as follows:
\begin{equation}
\vvvSPVLAD(B)=\left(\vSPVLAD_i=\sum_{x_t\in B}\gamma_i(x_t)(x_t-\mu_i)\right)_{i=1..N},
\label{eq:sPVLAD}
\end{equation}
where $\gamma_i(x_t)$ is the soft assignment of descriptor $x_t$ to $i$th Gaussian:
\begin{equation}
\gamma_i(x_t)=p(i|x_t,\lambda)=\frac{w_i p_i(x_t|\lambda)}{\sum_{j=1}^{N}w_j p_j(x_t|\lambda)}, \nonumber
\end{equation}

In the stability theorem for stable persistence VLAD (presented below) we assume that coordinates of the points in the considered persistence diagrams are limited to a certain compact subset of $\mathbb{R}^2$. This limitation is crucial to prove the stability and it is a reasonable assumption in case of TDA. Moreover, this theorem is true for any $\mathbb{R}^n$ (not only for $\mathbb{R}^2$).

\medskip
\noindent
{\bf Theorem} Let $B,B'\in\D$ be persistence diagrams, such that $B, B'\subset [a,b]\times[a,b]$. 
The~stable persistence VLAD with $N$ words is stable with respect to \mbox{1-Wasserstein} distance, that is:
\[
\norm{\vvvSPVLAD(B) - \vvvSPVLAD(B')}_{\infty} \leq C \cdot W_1(B, B'),
\]
where $C$ is a constant depending on $[a,b]\times[a,b]$.

\medskip
\noindent
{\bf Proof.} Let us first consider the following difference:
\begin{multline}
|\gamma_i(x_t) - \gamma_i(y_t)| = \left|\frac{w_i p_i(x_t)}{\sum_{j=1}^N w_j p_j(x_t)}-\frac{w_i p_i(y_t)}{\sum_{j=1}^N w_j p_j(y_t)}\right|\\
\begin{aligned}
&=\left|\frac{w_i p_i(x_t)\sum_{j=1}^N w_j p_j(y_t) - w_i p_i(y_t)\sum_{j=1}^N w_j p_j(x_t)}{\sum_{j=1}^N w_j p_j(x_t)\sum_{j=1}^N w_j p_j(y_t)}\right|\\
&=\left|\frac{w_i \sum_{j=1}^N \left(w_j p_i(x_t) p_j(y_t) - w_j p_i(y_t) p_j(x_t)\right)}{\sum_{j=1}^N\sum_{k=1}^N w_j w_k p_j(x_t) p_k(y_t)}\right|\\
&\leq C_1 \left| \sum_{j=1}^N \left(w_j p_i(x_t) p_j(y_t) - w_j p_i(y_t) p_j(x_t)\right) \right|\\
&\stackrel{(i)}{=}C_1 \left| \sum_{j=1}^N \left(w_j p_i(x_t) \left(p_j(y_t) - p_j(x_t) + p_j(x_t)\right) - w_j p_i(y_t) p_j(x_t)\right) \right|\\
&=C_1 \left| \sum_{j=1}^N w_j\left(p_i(x_t) \left(p_j(y_t) - p_j(x_t)\right) + p_j(x_t)\left( p_i(x_t) - p_i(y_t) \right)\right) \right|\\
&\stackrel{}{\leq}C_1 \left| \sum_{j=1}^N w_j\left(M_i L_j\norm{y_t - x_t}_\infty + M_j L_i\norm{x_t - y_t}_\infty \right) \right|\\
&\stackrel{}{\leq}C_1 N \max_j\{w_j\left(M_i L_j + M_j L_i \right)\}\norm{y_t - x_t}_\infty = C_2 \norm{y_t - x_t}_\infty,
\end{aligned}
\label{eq:estimation1_proof_svlad}
\end{multline}
where:
\begin{multline}
\begin{aligned}
&C_1 = \frac{\max_i\{w_i\}}{N^2 \min_j\{w_j^2\} \min_j\{\min \{p_j^2(z)\,|\,z\in[a,b]\times[a,b]\}\}},\nonumber\\
&C_2 = C_1  N \max_j\{w_j\left(M_i L_j + M_j L_i \right)\},\nonumber
\end{aligned}
\end{multline}
while $M_i$ and $L_i$ are the maximal value and Lipschitz constant of $i$-th Gaussian $p_i$.
Note that the constant $C_1$ exists because diagrams are supported in a compact subset of $\mathbb{R}^2$. Therefore, the Gaussians achieve a minimum value, which is bounded away from zero. When it comes to assignment $(i)$, we simply put $p_j(y_t)= p_j(y_t) - p_j(x_t) + p_j(x_t)$.

\medskip
\noindent For a fixed $i$ we can estimate:
\begin{multline}
\norm{\vSPVLAD_i(B) - \vSPVLAD_i(B')}_\infty
= \norm{\sum_{t}\gamma_i(x_t)(x_t-\mu_i) - \sum_{t}\gamma_i(y_t)(y_t-\mu_i)}_\infty \\
\begin{aligned}
&=\norm{\sum_{t}\gamma_i(x_t)(x_t-y_t+y_t-\mu_i) - \sum_{t}\gamma_i(y_t)(y_t-\mu_i)}_\infty \\
&=\norm{\sum_{t}\left(\gamma_i(x_t)-\gamma_i(y_t)\right)(y_t-\mu_i) + \sum_{t}\gamma_i(x_t)(x_t-y_t)}_\infty \\
&\stackrel{(ii)}{\leq}\norm{\sum_{t} C_2 (y_t-\mu_i) \norm{y_t - x_t}_\infty + \sum_{t}(x_t-y_t)}_\infty \\
&\stackrel{(iii)}{\leq} \sum_{t}{(C_3+1)\norm{y_t - x_t}_\infty} = C_4\sum_{t}{\norm{y_t - x_t}_\infty} \leq C_4 W_1(B, B'),
\end{aligned}
\label{eq:sPVLAD_stability}
\end{multline}
where:
\begin{multline}
\begin{aligned}
&C_3 = C_2 \max_i\{\max\{\norm{p-\mu_i}_\infty\ |\ p\in[a,b]\times[a,b] \}\},\nonumber\\
&C_4 = C_3 + 1.\nonumber
\end{aligned}
\end{multline}

\medskip
\noindent In $(ii)$ we used the estimation (\ref{eq:estimation1_proof_svlad}) and the fact that $\sup_y{\gamma_i(y)}=1$. The boundaries of $\mathbb{R}^2$ compact subset allow to determine the maximal distance between diagram points and the~Gaussian centers, which is used in $(iii)$ to estimate $C_3$. Note that $C_4$ is independent of $i$, hence:
\begin{multline}
\norm{\vvvSPVLAD(B)-\vvvSPVLAD(B')}_{\infty} = \max_i\{\norm{\vSPVLAD_i(B) - \vSPVLAD_i(B')}_\infty\}\\ \begin{aligned}
&\leq C_4 \cdot W_1(B, B'),\nonumber
\end{aligned}
\end{multline}
and
\begin{multline}
\norm{\vvvSPVLAD(B)-\vvvSPVLAD(B')}_{p} = \sqrt[\uproot{4}\scriptstyle p]{ \sum_i (\norm{\vSPVLAD_i(B) - \vSPVLAD_i(B')}_\infty)^p } \\
\begin{aligned}
& \leq \sqrt[\uproot{2}\scriptstyle p]{N\ C_4} \cdot W_1(B, B').\nonumber
\end{aligned}
\end{multline}
\begin{flushright} $\square$\end{flushright}

\medskip
\noindent We would like to point out that in the estimation of $C_1$ and $(iii)$ we use an assumption that diagrams are supported in the compact $\mathbb{R}^2$ subset $[a,b] \times [a,b]$. As a result, if the support of a persistence diagram sequence diverges to $\infty$, then the corresponding sequence of $C_1$ also diverges to infinity. Therefore $C_4$ is not a \emph{global} constant and the persistence VLAD is not \emph{globally} stable. We want to indicate, however, that for any practical case, the assumption about compact support of diagrams is always satisfied.

\subsection{Persistence Fisher Vector}
\label{pfv}

The idea of \emph{persistence Fisher vector} (PFV) is based on Fisher vectors introduced by~\cite{perronnin2007fisher} and relies on the gradient of the log-likelihood with respect to the parameters of a Gaussian mixture model. Compared to the traditional BoW model, it captures first and second order moments. It can be extended to PDs as follows.
Given a PD $B\in\D$ we aim a characterizing it with a gradient vector derived from a generative probability model (obtained for all PDs used for codebook generation). 
This is similar to sPVLAD, however in case of PFV we compute not only the first order, but also the second order moments of the points assigned to a codeword (i.e. not only the gradient for $\mu_i$ but also for $\Sigma_i$).

Let $\La(B|\lambda)=log\:p(B|\lambda)$, where under the independence assumption:
\[
\La(B|\lambda) = log\ \Pi_{x_t \in B} p(x_t|\lambda)) = \sum_{x_t\in B}log\:p(x_t|\lambda),
\]
where:
\[
p(x_t|\lambda)=\sum_{i=1}^{N}w_i p_i(x_t|\lambda),
\]
is the likelihood that point $x_t$ was generated by the GMM.

\noindent Assuming that the covariance matrices are diagonal (for ease of calculation), the derivations $\frac{\partial \La(B|\lambda)}{\partial \mu_i^d}$ and $\frac{\partial \La(B|\lambda)}{\partial \sigma_i^d}$ (where $\sigma_i^d = diag(\Sigma_i)$ and superscript $d$ denotes the $d$-th dimension of a vector) can be effectively computed as~\citep{perronnin2007fisher}:

\begin{equation}
\label{eq:dLdm}
\frac{\partial \La(B|\lambda)}{\partial \mu_i^d}=\sum_{x_t\in B}\gamma_i(x_t)\left[\frac{x_t^d-\mu_i^d}{(\sigma_i^d)^2}\right],
\end{equation}
\begin{equation}
\label{eq:dLds}
\frac{\partial \La(B|\lambda)}{\partial \sigma_i^d}=\sum_{x_t\in B}\gamma_i(x_t)\left[\frac{(x_t^d-\mu_i^d)^2}{(\sigma_i^d)^3}-\frac{1}{\sigma_i^d}\right].
\end{equation}
The gradient vector is just a concatenation of the partial derivatives with respect to all the~parameters.

To normalize the dynamic range of the different dimensions of the gradient vectors, the~diagonal of the Fisher information matrix $F_\lambda$ is computed as:
\[
F_\lambda=E_B[\nabla_\lambda \La(B|\lambda) \nabla_\lambda \La(B|\lambda)'].
\]
applied to partial derivatives, resulting in the final definition of Fisher vector:
\begin{equation}
\vvvPFV=\left( f_{\mu_i^d}^{-1/2} \frac{\partial \La(B|\lambda)}{\partial \mu_i^2}, f_{\sigma_i^d}^{-1/2} \frac{\partial \La(B|\lambda)}{\partial \sigma_i^d} \right)_{i=1..N},
\label{eq:sFV}
\end{equation}
where $f_{\mu_i^d}$ and $f_{\sigma_i^d}$ are the corresponding terms on the diagonal of $F_\lambda$. Vector $\vvvPFV$ is the concatenation of $N$ pairs of components containing $D=2$ values for every Gaussian component, therefore it is of size $4N$.

\medskip
\noindent
{\bf Theorem}
Let $B,B'\in\D$ be persistence diagrams, such that $B, B'\subset [a,b]\times[a,b]$. 
The~persistence Fisher vector with $N$ words is stable with respect to \mbox{1-Wasserstein} distance, that is:
\[
\norm{FV(B) - FV(B')}_{\infty} \leq C \cdot W_1(B, B'),
\]
where $C$ is a constant depending on $[a,b]\times[a,b]$.

\medskip
\noindent
{\bf Proof.}
Persistence Fisher Vector is a concatenation of the two components presented in equations (\ref{eq:dLdm}) and (\ref{eq:dLds}). In order to be stable, both components have to be stable with respect to 1-Wasserstein distance; therefore, we estimate them separately (we skip $d$ superscript from the original notation for clarity).

\noindent
The first FV component (\ref{eq:dLds}) can be estimated using the theorem about sPVLAD stability (\ref{eq:sPVLAD_stability}):
\begin{multline}
\norm{
\sum_{t}\gamma_i(x_t)\left[\frac{x_t-\mu_i}{(\sigma_i)^2}\right] 
-\sum_{t}\gamma_i(y_t)\left[\frac{y_t-\mu_i}{(\sigma_i)^2}\right]
}\\
\begin{aligned}
&= \frac{1}{(\sigma_i)^2}\norm{
\sum_{t}\gamma_i(x_t)(x_t-\mu_i) 
-\sum_{t}\gamma_i(y_t)(y_t-\mu_i)
}\\
&= \frac{1}{(\sigma_i)^2}\norm{\vSPVLAD_i(B) - \vSPVLAD_i(B')}_\infty \stackrel{(\ref{eq:sPVLAD_stability})}{\leq} \frac{C_4}{(\sigma_i)^2} W_1(B, B'),\nonumber
\end{aligned}
\end{multline}

\noindent The second component (\ref{eq:dLds}) can be estimated as follows:
\begin{multline}
\norm{
\sum_{t}\gamma_i(x_t)\left[\frac{(x_t-\mu_i)^2}{(\sigma_i)^3}-\frac{1}{\sigma_i}\right]
-
\sum_{t}\gamma_i(y_t)\left[\frac{(y_t-\mu_i)^2}{(\sigma_i)^3}-\frac{1}{\sigma_i}\right]
}\\
\begin{aligned}
&= \norm{
\sum_{t}\gamma_i(x_t)\left[\frac{(x_t-\mu_i)^2}{(\sigma_i)^3}-\frac{1}{\sigma_i}\right]
-
\sum_{t}\gamma_i(y_t)\left[\frac{(y_t-\mu_i)^2}{(\sigma_i)^3}-\frac{1}{\sigma_i} - 
\frac{(x_t-\mu_i)^2}{(\sigma_i)^3}+\frac{1}{\sigma_i}
+
\frac{(x_t-\mu_i)^2}{(\sigma_i)^3}-\frac{1}{\sigma_i}
\right]
}\\
&= \norm{
\sum_{t}(\gamma_i(x_t)-\gamma_i(y_t))\left[\frac{(x_t-\mu_i)^2}{(\sigma_i)^3}-\frac{1}{\sigma_i}\right]
-
\sum_{t}\gamma_i(y_t)\left[\frac{(y_t-\mu_i)^2}{(\sigma_i)^3}-
\frac{(x_t-\mu_i)^2}{(\sigma_i)^3}
\right]
}\\
&\stackrel{(iv)}{\leq} \norm{
\sum_{t}D_1 L_i(x_t-y_t)
-
\sum_{t}\frac{\gamma_i(y_t)}{(\sigma_i)^3} \left[(y_t-\mu_i)^2-(x_t-\mu_i)^2
\right]
}\\
&= \norm{
\sum_{t}D_1 L_i(x_t-y_t)
-
\sum_{t}
\frac{\gamma_i(y_t)}{(\sigma_i)^3}
\left[
y_t^2-x_t^2-2\ \mu_i(y_t-x_t)
\right]
}\\
&= \norm{
\sum_{t}D_1 L_i(x_t-y_t)
-
\sum_{t}\frac{\gamma_i(y_t)}{(\sigma_i)^3}
\left[
(y_t-x_t)(x_t+y_t)-2\ \mu_i(y_t-x_t)
\right]
}\\
&= \norm{
\sum_{t}D_1 L_i(x_t-y_t)
-
\sum_{t}\frac{\gamma_i(y_t)}{(\sigma_i)^3}\ (y_t-x_t)
\left[
(x_t+y_t)-2\ \mu_i
\right]
}\\
&\stackrel{(v)}{\leq} \norm{
\sum_{t}D_1 L_i(x_t-y_t)
-
\sum_{t}D_2\ (y_t-x_t)
}
= \norm{
\sum_{t}D_1 L_i(x_t-y_t)
+
\sum_{t}D_2\ (x_t-y_t)
}\\
&\leq D_3 \norm{
 \sum_{t}(x_t-y_t)
}
\leq 
D_3 W_1(B,B'),\nonumber
\end{aligned}
\end{multline}
where $D_1$ is an upper bound for $\frac{(x_t-\mu_i)^2}{(\sigma_i)^3}-\frac{1}{\sigma_i}$ $(iv)$, $D_2$ is a bound for $\frac{\gamma_i(y_t)}{(\sigma_i)^3}$ $(v)$, both on the domain $[a,b] \times [a,b]$, and $D_3 = D_1\ max\{L_i\} + D_2$.

\medskip
\noindent
Summing up the two estimates above, we conclude that persistence Fisher vector is stable with respect to 1-Wasserstein distance with a constant $D_3+\frac{C_4}{(\sigma_i)^2}$, where $D_3$ is defined above and $C_4$ is defined in Section~\ref{spvlad}.

\section{Experimental Setup}
\label{experiments}

To evaluate the proposed persistence BoW representations (PBoW, sPBoW, wPBoW, PVLAD, sPVLAD and PFV), we compare them with a number of state-of-the-art approaches including kernel-based methods and vectorized PD representations. The evaluation is performed on classification tasks involving different datasets representing heterogeneous data including, among others, 3D shapes, textures, and social media graphs. In the following, we describe the datasets used in our experiments, list the state-of-the-art approaches we compare with, and discuss the setup of the experiments.

\subsection{Datasets}
\label{subsec:datasets}

For the evaluation we incorporate various datasets which cover a wide range of different data types. Firstly, to provide a proof-of-concept, we evaluate all the approaches on a~synthetically generated shape classes from~\cite{adams2017persistence}. Next, the approaches are evaluated on real-world datasets for 3D shape segmentation~\citep{Carrire2017SlicedWK},  activity recognition in 3D motion capture data~\citep{conf/iccv/AliBS07}, geometry-informed material recognition~\citep{degol2016geometry}, classification of social network graphs~\citep{hofer2017deep} and analysis of 3D surface texture~\citep{zeppelzauer2017study}. The datasets are described in detail in the following sections. Where possible, we have used pre-computed PDs available with the datasets to foster reproducibility and comparability. As the computation times for some of the considered methods, especially for kernel-based approaches, do not scale well with the sizes of datasets, we have decided to randomly sub-sample some of the datasets (see details below).

\subsubsection{Synthetic Dataset}

\begin{figure}[ht]
\begin{center}
\includegraphics[width=0.7\columnwidth]{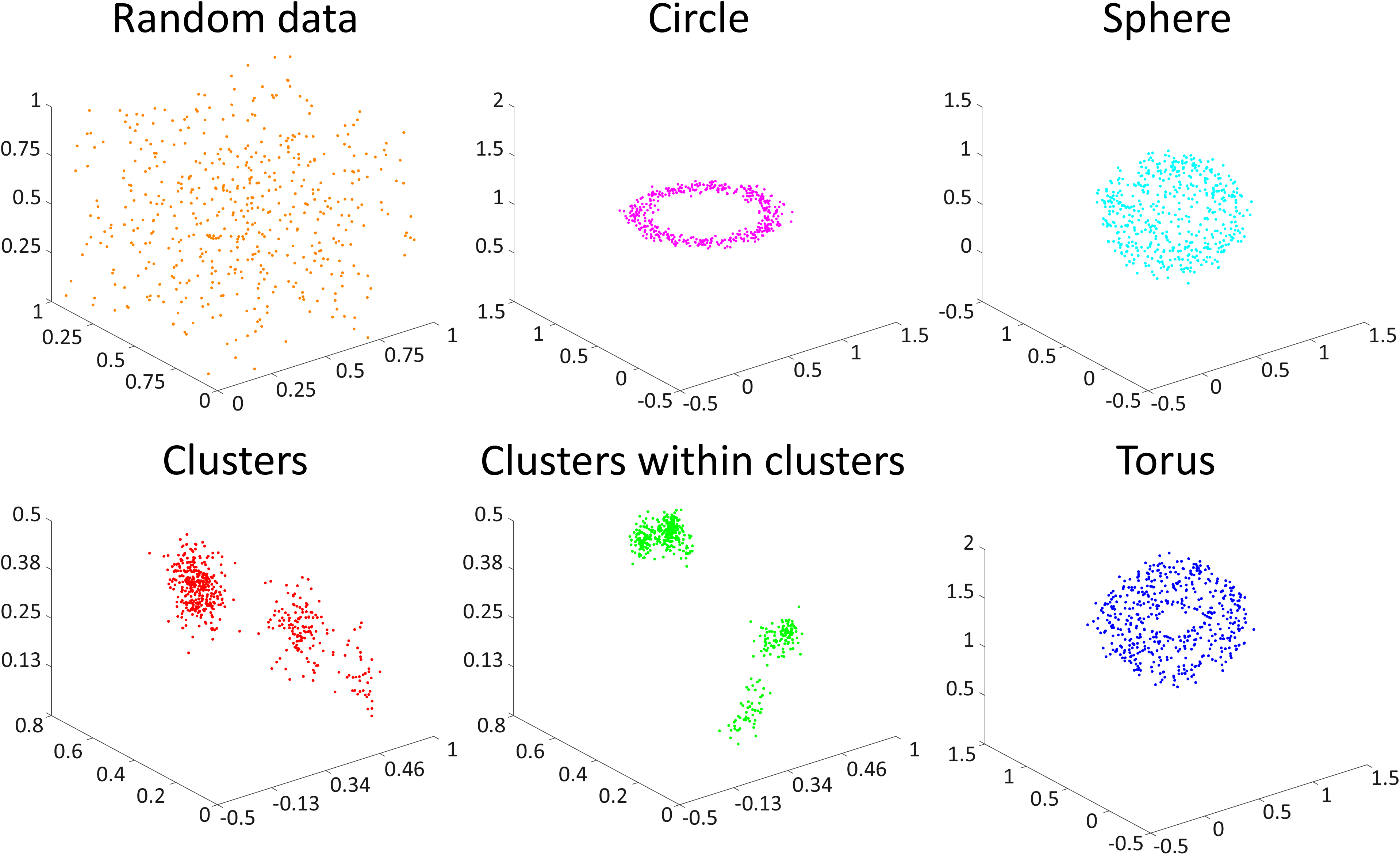}
\caption{Example shapes from the six shape classes of the synthetic dataset.}
\label{fig:syntheticData}
\end{center}
\end{figure}

The first dataset is a synthetic dataset introduced by Adams et al.~\citeyearpar{adams2017persistence}. It consists of seven shape classes represented by point clouds in $\R^3$ of the following geometrical objects: unit cube, circle of diameter one, sphere of diameter one, three clusters with centers randomly chosen from unit cube, hierarchical structure of three minor clusters within three major clusters (where the centers of the minor clusters are chosen as small perturbations from the major cluster centers), and a torus (see Fig.~\ref{fig:syntheticData} for example shapes). Each point cloud is randomly perturbed by positioning a Gaussian distribution of standard deviation $0.1$ at this point and sampling novel points from the distribution. Overall, this dataset contains $50$ point clouds for each of the six classes, each containing $500$ 3D points. This gives $300$ point clouds in total. 

From each point cloud we compute the PDs in dimension $1$ for a Vietoris-Rips filtration for a radius parameter equal to the maximal distance between points in the point cloud. We employ the approximation method proposed by~\cite{DBLP:conf/esa/DeySW16} and the SimBa implementation based on the work of Dayu Shi\footnote{\url{http://web.cse.ohio-state.edu/~dey.8/SimBa/Simba.html}, last visited April, 2019}.

\subsubsection{Geometry-Informed Material Recognition Dataset (GeoMat)}

The GeoMat dataset provides geometry information (point clouds) as well as visual images of 19 different materials, such as ``brick", ``grass" and ``gravel"~\citep{degol2016geometry}. The~GeoMat dataset contains patches sampled from larger photographs of surfaces from buildings and grounds. Each patch predominantly represents only one material, while each class consists of $600$ images, each of size $100\times 100$ pixels. Among them, there are pictures of different scales, i.e. $100\times 100$, $200\times 200$, $400\times 400$ and $800\times 800$. 

For each patch, the dataset provides a depth image\footnote{Source: \url{http://web.engr.illinois.edu/~degol2/pages/MatRec_CVPR16.html}, last visited April, 2019}, containing the local (fine-grained) surface texture and the global surface curvature. To filter out the global curvature, we transform each depth image into a point cloud in 3D space, consisting of $10000$ points (every point represents one of the $100\times 100$ pixels). Then, the resulting point cloud is rotated in a~way that the Z axis represents depth and the global surface curvature is removed by fitting a second degree function (paraboloid) to the point cloud and subtracting the approximated values from the Z coordinates of the original points. The values of the Z-coordinates are then centered at 0 and the point cloud is projected back into a bitmap (depth map). Ultimately, PDs are computed by gray-scale filtration.

\subsubsection{Social Network Graph Datasets (Reddit)}

To extend the range of different data types in our evaluation, we further incorporate graph-based datasets. To this end we employ the reddit-5k and reddit-12k datasets from~\cite{yanardag2015deep}, which contain discussion graphs from the reddit platform\footnote{Reddit is a content-aggregation website: \url{http:\\reddit.com}}. Nodes in the graphs correspond to users, and edges between users exist if one user has commented a posting of the other user. Different graphs are labeled by subreddits, which refer to different topics. The dataset reddit-5k contains overall 4999 graphs for 5 popular subreddits. The larger dataset, reddit-12k, contains 11929 graphs for 11 subreddits including topics like, e.g. ``worldnews", ``videos" and ``atheism". The task for both datasets is to predict the subreddit (topic) from the input graph. For both datasets we use the pre-computed PDs available online\footnote{Source: \url{https://github.com/c-hofer/nips2017}, last visited April, 2019}.

\subsubsection{3D Surface Texture Dataset (PetroSurf3D)}

A further dataset in our experiments is the recently released \emph{PetroSurf3D} dataset, which contains high-resolution 3D surface reconstructions from the archaeological domain with a~resolution of approximately 0.1 mm~\citep{PetroSurf3D}. 
The reconstructions represent 26 natural rock surfaces that exhibit human-made engravings (so-called rock art), and thereby exhibit complex 3D surface textures. The classification task for PetroSurf3D is to automatically predict which areas of the surface have been manipulated by tools (engravings) and~which have not, i.e. there are two classes of surface topographies: engraved areas and~the natural rock surface. Engraved areas represent approximately 19\% of the data. For~each surface, a precise pixel-accurate ground truth exists together with a depth map of the~surface. The depth maps are analyzed in a patch-wise manner. 
Overall, there are 754.386 square patches to classify from 26 surfaces. 
In order to keep the number of training samples in a practical range, we randomly subsampled each surface. Overall, a balanced set (equal class cardinalities) of 600 patches per surface ($26*600=15600$ samples) is used in each repetition of an experiment.

For each patch, a PD is computed by grayscale filtration over the surface depth ranges (depth maps) as a basis for our experiments. To normalize the values for different shaped surfaces, the depth value range is z-standardized before filtration.

\subsubsection{3D Shape Segmentation Dataset}

We further employ the 3D shape dataset from~\cite{chen2009benchmark} which was preprocessed by ~\cite{carriere2015stable} for topological data analysis. The preprocessed dataset contains PDs for $5700$ 3D points from airplane models. Each point is assigned to one sub-part (segment) of an airplane, e.g., 'wing', 'vertical stabilizer' and 'horizontal stabilizer'. For our experiments we use the PDs computed by~\cite{carriere2015stable}, available in their repository\footnote{Source: \url{https://github.com/MathieuCarriere/sklearn_tda}, last visited April, 2019}. The PDs were generated by tracking topology evolution of a geodesic ball centered at the individual points of the input 3D model. Thereby, the radius grows from 0 to infinity. We focus on PDs of dimension $1$ as the considered 3D shapes are connected. 
The task is to classify each point according to the segment it belongs to.

\subsubsection{Motion Capture Dataset}

Another real-world dataset represents 3-dimensional motion capture sequences of body joints~\citep{conf/iccv/AliBS07}. The dataset describes the following five activities: dancing, jumping, running, sitting and walking with $31$, $14$, $30$, $35$ and $48$ instances, respectively. For~each activity, a set of 19 3D motion trajectories (each corresponding to the motion of one tracked joint) is extracted. This corresponds to $3 \cdot 19=57$ curves of individual ($x$, $y$, and $z$) components for which 57 separate PDs are computed by~\cite{conf/iccv/AliBS07}. For our experiments we employ the original pre-computed PDs\footnote{Source: \url{https://github.com/rushilanirudh/pdsphere}, last visited April, 2019}.

Experiments with this dataset are performed only on the vectorized representations. For~kernel-based approaches we would have to compute 57 full kernel matrices, which is computationally expensive and would further require an adequate method for the combination of the kernels. For vectorized representations, the proceedure is much more efficient and straight-forward. We simply compute one vector per PD and concatenate them into a final feature vector for classification. For the BoW approaches, we compute 57 codebooks, one for each 3D motion component, and concatenate the corresponding codeword histograms. In case of the Riemannian manifold representation, RM~\citep{anirudh2016riemannian}, we generate vectorial features by PCA and concatenate them as proposed by the authors. We also use original procedure for PI~\citep{adams2017persistence}.

\subsection{Compared Approaches}

We compare our bag-of-word approaches with both kernel-based techniques and vectorized representations. Kernel-based approaches include: 2-Wasserstein distance\footnote{Source: \url{https://bitbucket.org/grey_narn/hera}, last visited April, 2019}, 2Wd~\citep{kerber2017geometry}, the multi-scale kernel\footnote{Source: \url{https://github.com/rkwitt/persistence-learning}, last visited April, 2019}, MK~\citep{reininghaus2014stable}, and sliced Wasserstein kernel\footnote{Code obtained from Mathieu Carri{\`e}re}, SWK~\citep{Carrire2017SlicedWK}. Furthermore, we employ the persistence landscape\footnote{Source: \url{https://www.math.upenn.edu/~dlotko/persistenceLandscape.html}, last visited April, 2019}$^,$\footnote{Source: \url{https://github.com/queenBNE/Persistent-Landscape-Wrapper}, last visited April, 2019} (PL) representation and generate a kernel matrix by the distance metric defined in~\citep{bubenik2015statistical}. Vectorized PD representations include: persistence image\footnote{Source: \url{https://github.com/CSU-TDA/PersistenceImages}, last visited April, 2019}, PI~\citep{adams2017persistence} and the Riemannian manifold approach\footnote{Source: \url{https://github.com/rushilanirudh/pdsphere}, last visited April, 2019}, RM~\citep{anirudh2016riemannian}. 

\subsection{Setup}
\label{subsec:setupOfExperiments}

For all datasets, except Reddit, in Section~\ref{subsec:datasets} we consider the PDs of dimension 1 as a~common input (cycles) since they best express the internal structure in the data and~yielded the most promising results in related works~\citep{adams2017persistence,carriere2015stable}. In~case of Reddit database we use PDs of dimension 0 (connected components), since graphs are considered as 1-complex; thus, first dimensional homology generators never die. In the considered datasets no infinite intervals of dimension 1 occur. In cases where infinite intervals are present, there are different ways to proceed: (1) ignoring them, (2) substituting infinity with some (large) number or (3) building separate representations for finite and infinite intervals. In the general case, we recommend to compute persistence codebooks for PDs of all available dimensions separately and to combine them before classification.

The classification pipeline is as follows. For the kernel-based approaches, we take the PDs as input and compute the explicit kernel matrices for the training and test samples. Next, we train an SVM from the explicitly computed kernel-matrix and evaluate it on the test set. For the vectorized representations we compute the respective feature vectors from the~PDs and feed them into a linear SVM for training. This procedure allows direct comparison between kernel-based approaches and vectorized representations.

For all datasets, we aim at solving a supervised classification task. In order to enhance the comparability with results obtained in original experiments, if available, we employ train/test divisions of samples based on the original procedures. To find optimal parameters for each evaluated approach, we run a grid search over their respective hyperparameters. The hyperparameters and their evaluated values for each approach are listed in Table~\ref{tab:kernelParameters} (for the kernel-based approaches) and in Table~\ref{tab:vectorParameters} (for the vectorized representations). The~optimal parameters are highlighted in bold. For each parameter combination, we run a complete experiment including cross-validation on the training set to evaluate its performance. The~number of repetitions for each parameter combination in the grid search depends on the~dataset and is provided in Tables~\ref{tab:kernelResults} and~\ref{tab:vectorResults}.

Our evaluation is partitioned into two sets of experiments. EXP-A uses all related approaches on a sub-sampled version of the datasets, while EXP-B operates only on the~vectorized representations and uses larger datasets. The reason for this is that for the larger datasets in our study it is inconvenient to compute the explicit kernel matrices for the kernel-based approaches for computational reasons. Nevertheless, to still enable a fair comparison of all approaches, we sub-sample the datasets in EXP-A to reduce their size, i.e. by randomly selecting 30 (GeoMat), 100 (reddit-5k), 50 (reddit-12k) and 390 (PetroSurf3D) samples for each class. In EXP-B we solely evaluate the vectorized representations on the~datasets as described in Section~\ref{subsec:datasets}.

The evaluation procedure for each dataset is as follows. For the synthetic dataset, we subsample 80\% of the samples as training data and use the remaining samples for testing. For GeoMat dataset, we use the original train/test partition with 400/200 samples per class. For Reddit experiments, we employ the original ratio of 90\% graphs in the training set and~the remaining 10\% in the test set. For the 3D shape segmentation dataset, we employ the~original 50/50  split. The train/test split ratio of the motion capture dataset is 80/20. For all datasets, we average the achieved performance of grid search over 5 repetitions of random selected training and test partitions. The only exception is PetroSurf3D, where we divided the set of all surfaces into 4 folds (resulting in four repetitions) according to original work of \cite{PetroSurf3D}.

Ultimately, we run a Wilcoxon signed-rank test (with p-value of $0.1$) on the results to identify which results significantly differ from the best obtained result, and which ones do not, and can thus be considered equally good. The comparison is performed between the~best method (the one with the best mean accuracy) and all the other methods, for each experiment separately. The mean accuracy is obtained as an average over 5 runs with the same train/test divisions used by all compared methods. The number of repetitions is relatively small for statistical tests, therefore we set p-value to $0.1$.

The entire code of all experiments (implemented in Matlab) is available at \url{https://github.com/bziiuj/pcodebooks}. For external approaches, we use the publicly available implementations of the original authors. For clustering and bag-of-words encoding, we employ the VLFeat library~\citep{vedaldi08vlfeat}.


\section{Results}
\label{results}

Table~\ref{tab:kernelResults} summarizes the results obtained in our experiments for EXP-A and EXP-B. For~each combination of dataset and approach, we provide the obtained classification accuracy (including the standard deviation) and the processing time needed to construct the representations (excluding the time for classification). Note that for the synthetic dataset and the 3D shape segmentation dataset, results of EXP-A and EXP-B are equal, as no sub-sampling was needed to perform EXP-A.

\begin{sidewaystable}
\begin{center}
\begin{sc}
\begin{tiny}
\footnotetext[1]{2Wd - 2-Wasserstein Distance, MK - multiscale kernel, SWK - sliced Wasserstein kernel, PL - persistence landscape, PI - persistence image, RM - Riemmanian manifold, PBoW - persistence bag of words, wPBoW - weighted persistence bag of words, sPBoW - stable persistence bag of words, PVLAD - persistence VLAD, sPVLAD - stable persistence VLAD, PFV - persistence Fisher vector}
\begin{tabularx}{1.01\textwidth}{c|rr|rr|rr|rr|rr|rr}
\toprule
\multirow{2}{*}{\small \thead{Descr.\\ name\footnotemark[1]}} &  
\multicolumn{2}{c}{\thead{ \textbf{Synthetic} \\\textbf{Data}\\ (5 reps.)}} &  
\multicolumn{2}{c}{\thead{{\bf GeoMat} \\ (subset  $30$ \\smpl./cl.)\\ (5 reps.)}} & 
\multicolumn{2}{c}{\thead{{\bf Reddit-5k} \\(subset  $100$ \\smpl./cl.)\\ (5 reps.)}} &  
\multicolumn{2}{c}{\thead{{\bf Reddit-12k} \\(subset  $50$ \\smpl./cl.)\\ (5 reps.)}} &
\multicolumn{2}{c}{\thead{{\bf PetroSurf3D} \\(subset  $390$ \\smpl./cl.)\\ (4 reps.)}} &
\multicolumn{2}{c}{\thead{{\bf 3D Shape Segm.}\\ (5 reps.) }}\\
{} & Score & Time & Score & Time & Score & Time & Score & Time & Score & Time\\
\hline
2Wd & 
    \boldmath$98.00\pm1.28$ & $133.40s$ & 
    \boldmath$24.74\pm3.14$ & $14740.60s$ &
    $29.60\pm5.55$ & $6198.80s$ &
    $23.27\pm2.37$ & $4978.80s$ &
    \boldmath$79.88\pm6.58$ & $63821.50s$ &
	$71.01\pm0.45$ & $6300.80s$
    \\
MK & 
    $92.34\pm2.82$ & $44.04s$ & 
    $9.12\pm3.08$ & $10883.00s$ & 
    $32.67\pm3.06$ & $7251.67s$ & 
    $26.67\pm4.58$ & $4222.33s$ & 
    \boldmath$80.16\pm3.57$ & $53831.00s$ & 
    $88.48\pm0.47$ & $860.33s$ \\
SWK & 
	\boldmath$97.43\pm1.86$ & $33.31s$ &
	$22.00\pm2.72$ & $1069.61s$ &
	$39.20\pm6.10$ & $373.77s$ &
	$30.91\pm4.98$ & $329.68s$ &
	\boldmath$77.98\pm6.00$ & $3913.04s$ &
	\boldmath$94.22\pm0.48$ & $1995.42s$
    \\
PL & 
    $95.14\pm2.17$ & $81.01s$ & 
    $17.89\pm3.83$ & $1563.67s$ &
    $30.40\pm6.23$ & $790.34s$ &
    $24.73\pm4.19$ & $717.98s$ &
    \boldmath$78.21\pm6.67$ & $7990.66s$ &
	$92.55\pm0.79$ & $2668.32s$
    \\
\hline
PI & 
    \boldmath$98.29\pm1.56$ & $10.06s$ & 
	$11.05\pm2.44$ & $342.27s$ &
	\boldmath$48.40\pm5.73$ & $1250.88s$ &
    \boldmath$35.64\pm4.56$ & $111.78s$ &
    \boldmath$80.00\pm3.67$ & $1876.98s$ &
	\boldmath$94.86\pm0.29$ & $291.22s$
    \\
RM & 
    $92.29\pm2.60$ & $0.88s$ & 
	$19.05\pm2.72$ & $10.34s$ &
	$39.20\pm9.23$ & $7.76s$ &
    $28.73\pm1.99$ & $17.94s$ &
    \boldmath$77.26\pm6.29$ & $128.83s$ &
	$72.29\pm0.42$ & $6.57s$
    \\
\hline
PBoW  & 
    \boldmath$98.00\pm2.17$ & $0.82s$ & 
	\boldmath$26.74\pm2.51$ & $0.27s$ &
	\boldmath$46.80\pm4.82$ & \boldmath$0.53s$ &
    \boldmath$32.73\pm2.23$ & \boldmath$0.32s$ &
    \boldmath$79.88\pm4.64$ & \boldmath$1.78s$ &
	$90.78\pm0.70$ & $5.37s$
    \\
wPBoW &
    \boldmath$97.71\pm0.78$ & \boldmath$0.20s$ & 
    \boldmath$26.42\pm2.45$ & $1.96s$ &
    \boldmath$46.80\pm4.38$ & $1.22s$ &
    \boldmath$32.00\pm4.38$ & $1.33s$ &
    \boldmath$79.64\pm6.33$ & $3.54s$ &
    $90.09\pm0.96$ & $4.25s$
    \\
sPBoW &
    \boldmath$96.86\pm1.56$ & $3.96s$ & 
	$22.21\pm2.59$ & $2.18s$ &
	\boldmath$45.60\pm5.37$ & $0.74s$ &
    \boldmath$31.64\pm2.76$ & $1.29s$ &
    \boldmath$78.81\pm3.68$ & $31.50s$ &
	\boldmath$94.42\pm0.56$ & $14.74s$
    \\
PVLAD & 
    $94.00\pm1.56$ & $0.49s$ & 
	$21.05\pm5.59$ & \boldmath$0.20s$ &
    $38.40\pm8.65$ & $1.23s$ &
    $25.45\pm2.87$ & $1.18s$ &
    \boldmath$78.21\pm3.84$ & $3.94s$ &
	$87.35\pm0.43$ & \boldmath$3.57s$
    \\
sPVLAD &
    $95.71\pm2.47$ & $0.90s$ & 
	\boldmath$22.95\pm3.36$ & $3.80s$ &
    $38.40\pm6.23$ & $1.52s$ &
    $30.91\pm5.30$ & $1.26s$ &
    \boldmath$79.88\pm4.07$ & $6.82s$ &
	$94.04\pm0.41$ & $20.83s$
    \\
PFV & 
    \boldmath$97.14\pm1.01$ & $0.26s$ & 
	\boldmath$26.74\pm1.84$ & $3.54s$ &
    \boldmath$47.20\pm3.90$ & $1.39s$ &
    \boldmath$34.18\pm5.36$ & $0.74s$ &
    \boldmath$80.48\pm5.51$ & $9.23s$ &
	\boldmath$94.19\pm0.73$ & $5.97s$
    \\
\hline
\end{tabularx}
\caption{Results of EXP-A averaged over 5 runs.}
\label{tab:kernelResults}
\vspace{10pt}

\begin{tabularx}{0.855\textwidth}{c|rr|rr|rr|rr|rr}
\toprule
\multirow{2}{*}{\small \thead{Descr.\\ name\footnotemark[1]}} &  
\multicolumn{2}{c}{\thead{ \textbf{Motion} \\\textbf{Capture} \\ (5 reps.)}} &  
\multicolumn{2}{c}{ \thead{{\bf GeoMat}\\ (5 reps.)}} & 
\multicolumn{2}{c}{\thead{{\bf Reddit-5k}\\ (5 reps.)}} &  
\multicolumn{2}{c}{\thead{{\bf Reddit-12k}\\ (5 reps.) }} &
\multicolumn{2}{c}{\thead{{\bf PetroSurf3D}\\ (4 reps.) }}
\\
{} & Score & Time & Score & Time & Score & Time & Score & Time\\
\hline
PI & 
	\boldmath$92.00\pm5.06$ & $166.59s$ &
    $22.45\pm0.00$ & $7243.12s$ &
    \boldmath$49.34\pm2.83$ & $4759.66s$ &
    \boldmath$38.37\pm0.89$ & $11329.96s$ &
    \boldmath$80.40\pm4.92$ & $12137.04s$
    \\
RM & 
	\boldmath$94.00\pm2.79$ & $15.50s$ &
    $9.37\pm0.00$ & $222.59s$ &
    $46.25\pm3.11$ & $108.12s$ &
    $32.55\pm1.08$ & $215.50s$ &
    \boldmath$79.85\pm5.01$ & $1450.95s$
    \\
\hline
PBoW  &
    \boldmath$94.00\pm3.65$ & \boldmath$1.17s$ &
    $28.96\pm0.40$ & \boldmath$5.24s$ &
    \boldmath$49.94\pm3.28$ & \boldmath$1.53s$ &
    \boldmath$38.64\pm0.87$ & \boldmath$4.78s$ &
    \boldmath$80.34\pm5.25$ & \boldmath$28.52s$
    \\
wPBoW &
	\boldmath$94.67\pm2.98$ & $2.48s$ &
    \boldmath$29.79\pm0.32$ & $26 .05s$ &
    \boldmath$48.90\pm2.66$ & $13.57s$ &
    $36.97\pm1.03$ & $20.75s$ &
    \boldmath$80.43\pm5.35$ & $82.96s$
    \\
sPBoW &
	$89.33\pm4.35$ & $7.60s$ &
    $27.79\pm0.75$ & $29.70s$ &
	$46.57\pm3.33$ & $5.17s$ &
    $35.34\pm1.43$ & $29.34s$ &
    \boldmath$79.91\pm5.17$ & $161.76s$
    \\
PVLAD &
	\boldmath$94.67\pm2.98$ & $1.37s$ &
    $23.78\pm0.66$ & $28.24s$ &
    $42.57\pm2.83$ & $14.35s$ &
    $30.90\pm0.31$ & $22.75s$ &
    \boldmath$78.43\pm4.84$ & $80.16s$
    \\
sPVLAD &
	\boldmath$93.33\pm3.33$ & $7.31s$ &
    $27.40\pm0.64$ & $26.81s$ &
    $42.73\pm2.26$ & $31.39s$ &
    $33.47\pm1.01$ & $35.49s$ &
    \boldmath$80.22\pm5.24$ & $122.88s$
    \\
PFV & 
	\boldmath$94.00\pm3.65$ & $3.02s$ &
    \boldmath$29.29\pm0.61$ & $27.91s$ &
    \boldmath$49.10\pm1.91$ & $15.38s$ &
    \boldmath$39.21\pm1.30$ & $21.38s$ &
    \boldmath$80.68\pm5.16$ & $83.00s$
    \\
\hline\hline
\thead{State\\of art} & 
    $89.70$\footnotemark[2] & &%
    $22.32\pm0.76$\footnotemark[3] &  &
    $49.10$\footnotemark[4] & &
    $38.50$\footnotemark[4] & &
    n/a \footnotemark[5] & \\
\hline
\end{tabularx}
\footnotetext[2]{\cite{conf/iccv/AliBS07}}
\footnotetext[3]{Results obtained using original classification algorithm of \cite{degol2016geometry} fed with only surface shape information (i.e. map and histogram of normals)}
\footnotetext[4]{\cite{hofer2017deep}}
\footnotetext[5]{Not applicable, original research\citep{zeppelzauer2017study} use dice similarity index (DSC) instead of accuracy}
\caption{Results of EXP-B averaged over 5 runs.}
\label{tab:vectorResults}
\end{tiny}
\end{sc}
\end{center}
\end{sidewaystable}

Overall, in all experiments codebook representations of persistence diagrams achieve state-of-the-art performance (if available) or above. From EXP-A we further observe that vectorized representations (including the proposed ones), in general, perform better than kernel-based approaches. In case of the PetroSurf3D dataset, it is impossible to unambiguously determine the best method, since all approaches work equally well. For all other datasets, only the 2-Wasserstein distance and the sliced Wasserstein kernel attain accuracy comparable to vectorized approaches. 
Among the compared vectorized representations, PI in most cases outperforms RM and will thus serve as the primary approach for comparison with our approaches in subsequent sections. 
When comparing the stable vs. unstable variants of PBoW and PVLAD, we observe that PBoW in most cases outperforms its stable equivalent, especially for motion capture database in EXP-B, where sPBoW is $\sim$5\% worse than the other methods. Opposite is the case for PVLAD, where sPVLAD in most cases yields a higher performance. Nevertheless, sPVLAD, for almost all datasets, is still significantly worse than any other codebook variant.
The wPBoW works almost as well as non-weighted PBoW. Only in GeoMat and Reddit12K in EXP-B there exist visible differences in performance; however, even these results are very close ($\sim$1\% and $\sim$1.5\%). 
PFV variant seems to be the best approach, since in all experiments we can find it among the~best performing methods. Overall, however, PBoW versus PFV, there is no clear winner.

Large differences exist in the processing times of the different approaches. The highest runtimes are obtained for the kernel-based approaches going up to $63k$ seconds for the~PetroSurf3D dataset. The slowest kernel is 2Wd followed by MK, and approximately one order of magnitude faster PL and SWK. 
Note that computation complexity depends not only on a~number of persistence diagrams, but is also highly affected by the average number of points per diagram.
For the vectorized approaches, PI takes longest to compute. The~runtimes, however, vary strongly, depending on the resolution of the employed PI (note that we have estimated the optimal parameters for each dataset by a grid search over all hyperparameters, see Tables \ref{tab:kernelParameters} and \ref{tab:vectorParameters}). The RM representation is one to two magnitudes faster than PI\footnote{Note that for both representations we use the implementations provided by the original authors}. The proposed approaches outperform almost all state-of-the-art approaches in runtime for all datasets, both for EXP-A and EXP-B. The gain in runtime efficiency ranges from one to up to four orders of magnitude. For the largest dataset in the experiments (PetrSurf3D in EXP-B), the fastest (PBoW) and the slowest (sPBoW) codebook approaches are still 3 and 2 magnitudes faster than PI, while reaching comparable accuracy. Concerning EXP-A, the codebook approaches are comparable in computing time, which is due to the small size of the datasets. EXP-B demonstrates well how the different approaches scale to larger data. It shows that PBoW scales best (is fastest) and still obtains optimal results in all but one case. It thus represents the best tradeoff between time efficiency and classification accuracy. See Section \ref{subsec:resultsAccTime} for further discussion.

From our experiments we conclude that persistence codebooks are significantly faster than related approaches while achieving similar or even better performance level. This shows that the codebooks capture well the essential information contained in the PDs and important for the respective classification tasks. The variablity of runtimes between the~different codebook variants is low compared to the other approaches. Thus, for the~selection of the appropriate codebook approach for a given problem in practice, the runtime plays a~secondary role.

In the following sections, we analyze selected aspects of the novel representations in greater detail, such as runtime, dependency on parameters and the scalability of the approach to large number of input PDs.

\begin{sidewaystable}
\begin{tiny}
\begin{sc}
\centering
\begin{tabularx}{1.04\textwidth}{l|c|c|c|c|c|c|c}
\hline
\toprule
\textbf{Descr.\footnotemark[1]} &  {} &  
    \textbf{Synthetic} &  
    \textbf{GeoMat} & 
    \textbf{Reddit-5k} &  
    \textbf{Reddit-12k} & 
    \textbf{PetroSurf3D} & 
    \textbf{3D Shape Segm.}\\
\midrule
\hline
\thead{MK} & $n$ & 
    \thead{$\{\textbf{0.5}, 1, 2\}$} &
    \thead{$\{0.5, \textbf{1}, 2\}$} &
    \thead{$\{\textbf{0.5}, 1, 2\}$} &
    \thead{$\{0.5, 1, \textbf{2}\}$} &
    \thead{$\{\textbf{0.5}, 1, 2\}$} &
    \thead{$\{\textbf{0.5}, 1, 2\}$} \\
\hline
\thead{SWK} & $n$ &  
    \thead{$\{\textbf{50}, 100, 150, 200, 250\}$} &
    \thead{$\{\textbf{50}, 100, 150, 200, 250\}$} &
    \thead{$\{\textbf{50}, 100, 150, 200, 250\}$} &
    \thead{$\{\textbf{50}, 100, 150, 200, 250\}$} &
    \thead{$\{\textbf{50}, 100, 150, 200, 250\}$} &
    \thead{$\{\textbf{50}, 100, 150, 200, 250\}$} \\
\hline
\multirow{3}{*}{PI} & $r$ & 
        \thead{$\{10,20,...,\textbf{50},...100\}$} &
        \thead{$\{10,20,40,$\\$60,80,\textbf{100}\}$} &
        \thead{$\{10,20,30,40,50,$\\$60,80,100,\textbf{120}\}$} &
        \thead{$\{10,20,30,\textbf{40},50,$\\$60,80,100,120\}$} &
        \thead{$\{10,20,...\textbf{80},...100\}$} &
        \thead{$\{10,20,...\textbf{70},...100\}$}
        \\
    {}  & $\sigma$ & 
        \thead{$\{\textbf{0.1}, 0.5, 1, 2\}$} & 
        \thead{$\{\textbf{0.5}, 1, 2\}$} & 
        \thead{$\{\textbf{0.5}, 1, 2, 3\}$} & 
        \thead{$\{\textbf{0.5}, 1, 2, 3\}$} & 
        \thead{$\{0.5, \textbf{1}, 2\}$} & 
        \thead{$\{\textbf{0.5}, 1, 2, 3\}$} \\
    {}  & $\sfw$ & 
        $\{\textbf{\sfw}, \sfnw\}$ & 
        $\{\sfw, \textbf{\sfnw}\}$ &
        $\{\sfw, \textbf{\sfnw}\}$ & 
        $\{\sfw, \textbf{\sfnw}\}$ &
        $\{\textbf{\sfw}, \sfnw\}$ & 
        $\{\textbf{\sfw}, \sfnw\}$\\  
\hline
\multirow{3}{*}{\thead{RM}} & $r$ &
        $\thead{\{\textbf{10}, 20, 40\}}$ & 
        $\thead{\{\textbf{10}, 20, 40, 60\}}$ & 
        $\thead{\{\textbf{10}, 20, 40, 60\}}$ & 
        $\thead{\{10, 20, \textbf{40}, 60\}}$ & 
        $\thead{\{10, 20, 40, \textbf{60}\}}$ & 
        $\thead{\{\textbf{10}, 20, 40, 60\}}$ \\
    {}  & $\sigma$ &
        $\{0.1, \textbf{0.2}, 0.3\}$ &
        $\{\textbf{0.1}, 0.2, 0.3\}$ &
        $\{\textbf{0.1}, 0.2, 0.3\}$ &
        $\{\textbf{0.1}, 0.2, 0.3\}$ &
        $\{\textbf{0.1}, 0.2, 0.3\}$ &
        $\{\textbf{0.1}, 0.2, 0.3\}$ \\
    {}  & $d$ & 
        $\thead{\{\textbf{50}, 75, 100\}}$ &
        $\thead{\{50, \textbf{75}, 100\}}$ &
        $\thead{\{50, \textbf{75}, 100\}}$ &
        $\thead{\{50, 75, \textbf{100}\}}$ &
        $\thead{\{50, \textbf{75}, 100\}}$ &
        $\thead{\{\textbf{50}, 75, 100\}}$ \\
\hline
\multirow{2}{*}{PBOW}   & $N$ & 
        \thead{$\{10,20,30,...\textbf{200}\}$} & 
        \thead{$\{10,20,\textbf{30},40,60,$\\$80,...200\}$} & 
        \thead{$\{10,20,\textbf{30},...,60,$\\$80,100,...200\}$} & 
        \thead{$\{10,20,30,40,\textbf{50},$\\$60,80,100,120\}$} &
        \thead{$\{10,20,30,40,50,$\\$60,80,...\textbf{160},...200\}$} &
        \thead{$\{10,20,...,\textbf{80},...100\}$}
        \\
    {}  & $\sfw$ & 
        $\{\textbf{\sfw}, \sfnw\}$ & 
        $\{\sfw, \textbf{\sfnw}\}$ &
        $\{\sfw, \textbf{\sfnw}\}$ & 
        $\{\sfw, \textbf{\sfnw}\}$ &
        $\{\sfw, \textbf{\sfnw}\}$ & 
        $\{\textbf{\sfw}, \sfnw\}$\\ 
    {} & $S$ &
        \thead{$\{1000, 5000, \textbf{10000}\}$} &
        \thead{$\{\textbf{2000}, 10000, 50000\}$} &
        \thead{$\{\textbf{2000}, 10000, 50000\}$} &
        \thead{$\{5000, \textbf{10000}, 50000\}$} &
        \thead{$\{\textbf{5000}, 10000, 50000\}$} &
        \thead{$\{5000, \textbf{10000}, 20000\}$} \\
\hline
\multirow{2}{*}{wPBOW}   & $N$ & 
        \thead{$\{10,20,...\textbf{110},...200\}$} & 
        \thead{$\{10,20,30,40,60,$\\$...,\textbf{160},...200\}$} & 
        \thead{$\{10,\textbf{20},...,60,$\\$80,100,...200\}$} & 
        \thead{$\{10,20,30,40,50,$\\$60,\textbf{80},100,120\}$} &
        \thead{$\{10,20,30,40,\textbf{50},$\\$60,80,...200\}$} &
        \thead{$\{10,20,...\textbf{60},...00\}$}
        \\
    {}  & $\sfw$ &
        $\{\sfw, \textbf{\sfnw}\}$ & 
        $\{\textbf{\sfw}, \sfnw\}$ &
        $\{\sfw, \textbf{\sfnw}\}$ & 
        $\{\sfw, \textbf{\sfnw}\}$ &
        $\{\sfw, \textbf{\sfnw}\}$ & 
        $\{\textbf{\sfw}, \sfnw\}$\\     
    {} & $S$ &
        \thead{$\{\textbf{1000}, 5000, 10000\}$} &
        \thead{$\{2000, \textbf{10000}, 50000\}$} &
        \thead{$\{\textbf{2000}, 10000, 50000\}$} &
        \thead{$\{\textbf{5000}, 10000, 50000\}$} &
        \thead{$\{\textbf{5000}, 10000, 50000\}$} &
        \thead{$\{5000, \textbf{10000}, 20000\}$} \\
\hline
\multirow{2}{*}{sPBOW}   & $N$ & 
        \thead{$\{10,20,...\textbf{140},...200\}$} & 
        \thead{$\{10,20,\textbf{30},40,60,$\\$80,...200\}$} & 
        \thead{$\{10,\textbf{20},...,60,$\\$80,...200\}$} & 
        \thead{$\{\textbf{10},20,30,40,50,$\\$60,80,100,120\}$} &
        \thead{$\{10,20,30,40,50,$\\$60,80,\textbf{100}...200\}$} &
        \thead{$\{10,20,...\textbf{70},...100\}$}
        \\
    {}  & $\sfw$ &
        $\{\textbf{\sfw}, \sfnw\}$ & 
        $\{\textbf{\sfw}, \sfnw\}$ &
        $\{\sfw, \textbf{\sfnw}\}$ & 
        $\{\textbf{\sfw}, \sfnw\}$ &
        $\{\textbf{\sfw}, \sfnw\}$ & 
        $\{\textbf{\sfw}, \sfnw\}$\\     
    {} & $S$ &
        \thead{$\{1000, \textbf{5000}, 10000\}$} &
        \thead{$\{\textbf{2000}, 10000, 50000\}$} &
        \thead{$\{2000, \textbf{10000}, 50000\}$} &
        \thead{$\{\textbf{5000}, 10000, 50000\}$} &
        \thead{$\{5000, 10000, \textbf{50000}\}$} &
        \thead{$\{5000, 10000, \textbf{20000}\}$} \\
\hline
\multirow{2}{*}{PVLAD}   & $N$ & 
        \thead{$\{\textbf{10},20,...200\}$} & 
        \thead{$\{\textbf{10},20,30,40,60,$\\$80,...200\}$} & 
        \thead{$\{\textbf{10},20,...,60,$\\$80,100,...200\}$} & 
        \thead{$\{\textbf{10},20,30,40,50,$\\$60,80,100,120\}$} &
        \thead{$\{10,20,\textbf{30},40,50,$\\$60,80,...200\}$} &
        \thead{$\{10,\textbf{20},...,100\}$}
        \\
    {}  & $\sfw$ &
        $\{\textbf{\sfw}, \sfnw\}$ & 
        $\{\sfw, \textbf{\sfnw}\}$ &
        $\{\textbf{\sfw}, \sfnw\}$ & 
        $\{\textbf{\sfw}, \sfnw\}$ &
        $\{\textbf{\sfw}, \sfnw\}$ & 
        $\{\sfw, \textbf{\sfnw}\}$\\ 
    {} & $S$ &
        \thead{$\{1000, 5000, \textbf{10000}\}$} &
        \thead{$\{\textbf{2000}, 10000, 50000\}$} &
        \thead{$\{2000, \textbf{10000}, 50000\}$} &
        \thead{$\{5000, \textbf{10000}, 50000\}$} &
        \thead{$\{5000, \textbf{10000}, 50000\}$} &
        \thead{$\{5000, 10000, \textbf{20000}\}$} \\
\hline
\multirow{2}{*}{sPVLAD}   & $N$ & 
        \thead{$\{10,20,\textbf{30},...200\}$} & 
        \thead{$\{10,\textbf{20},30,40,60,$\\$80,...200\}$} & 
        \thead{$\{10,20,\textbf{30},...,60,$\\$80,100,...200\}$} & 
        \thead{$\{\textbf{10},20,30,40,50,$\\$60,80,100,120\}$} &
        \thead{$\{10,20,30,40,\textbf{50},$\\$60,80,...200\}$} &
        \thead{$\{10,20,...,\textbf{100}\}$}
        \\
    {}  & $\sfw$ &
        $\{\sfw, \textbf{\sfnw}\}$ & 
        $\{\sfw, \textbf{\sfnw}\}$ &
        $\{\textbf{\sfw}, \sfnw\}$ & 
        $\{\textbf{\sfw}, \sfnw\}$ &
        $\{\textbf{\sfw}, \sfnw\}$ & 
        $\{\textbf{\sfw}, \sfnw\}$\\ 
    {} & $S$ &
        \thead{$\{1000, \textbf{5000}, 10000\}$} &
        \thead{$\{2000, 10000, \textbf{50000}\}$} &
        \thead{$\{\textbf{2000}, 10000, 50000\}$} &
        \thead{$\{5000, \textbf{10000}, 50000\}$} &
        \thead{$\{5000, \textbf{10000}, 50000\}$} &
        \thead{$\{5000, 10000, \textbf{20000}\}$} \\
\hline
\multirow{2}{*}{PFV}   & $N$ & 
        \thead{$\{10,\textbf{20},...200\}$} & 
        \thead{$\{10,\textbf{20},30,40,60,$\\$80,...200\}$} & 
        \thead{$\{\textbf{10},20,...,60,$\\$80,100,...200\}$} & 
        \thead{$\{10,20,30,\textbf{40},50,$\\$60,80,100,120\}$} &
        \thead{$\{10,20,\textbf{30},40,50,$\\$60,80,...200\}$} &
        \thead{$\{10,\textbf{20},...,100\}$}
        \\
    {}  & $\sfw$ &
        $\{\textbf{\sfw}, \sfnw\}$ & 
        $\{\sfw, \textbf{\sfnw}\}$ &
        $\{\textbf{\sfw}, \sfnw\}$ & 
        $\{\sfw, \textbf{\sfnw}\}$ &
        $\{\sfw, \textbf{\sfnw}\}$ & 
        $\{\sfw, \textbf{\sfnw}\}$\\ 
    {} & $S$ &
        \thead{$\{\textbf{1000}, 5000, 10000\}$} &
        \thead{$\{2000, 10000, \textbf{50000}\}$} &
        \thead{$\{\textbf{2000}, 10000, 50000\}$} &
        \thead{$\{5000, \textbf{10000}, 50000\}$} &
        \thead{$\{5000, 10000, \textbf{50000}\}$} &
        \thead{$\{5000, 10000, \textbf{20000}\}$} \\
\hline
\end{tabularx}

\caption{Parameters tested for EXP-A. The optimal parameters are in bold. Abbreviations: $n$ is the number of lines slicing the plane in SWK, $r$ is resolution of PI or density map in RM, $\sigma$ is sigma of the Gaussians employed, $\sfw$ is weighting (either no weighting (\sfnw) or with weighting (\sfw), $d$ is dimension of Principal Geodesic Analysis on the hypersphere, $N$ is the number of codewords of persistence codebooks.}
\label{tab:kernelParameters}

\end{sc}
\end{tiny}
\end{sidewaystable}

\begin{sidewaystable}
\begin{tiny}
\begin{sc}
\centering
\begin{tabularx}{0.845\textwidth}{l|c|c|c|c|c|c}
\hline
\toprule
\textbf{Descr.\footnotemark[1]} &  {} &  
    \textbf{Motion Capture} &  
    \textbf{GeoMat} & 
    \textbf{Reddit-5k} &  
    \textbf{Reddit-12k} & 
    \textbf{PetroSurf3D}\\
\midrule
\hline
\multirow{3}{*}{PI} & $r$ & 
        \thead{$\{\textbf{10},20,30,40,50,60\}$} &
        \thead{$\{10,20,30,40,50,\textbf{60}\}$} &
        \thead{$\{10,20,30,40,50,\textbf{60}\}$} &
        \thead{$\{10,20,30,40,\textbf{50},60\}$} &
        \thead{$\{10,20,30,40,\textbf{50},60\}$}
        \\
    {}  & $\sigma$ & 
        \thead{$\{0.5, 1, \textbf{2}\}$} & 
        \thead{$\{\textbf{0.5}, 1, 2, 3\}$} & 
        \thead{$\{\textbf{0.5}, 1, 2\}$} & 
        \thead{$\{\textbf{0.5}, 1, 2\}$} & 
        \thead{$\{0.5, 1, \textbf{2}\}$} \\
    {}  & $\sfw$ & 
        $\{\sfw, \textbf{\sfnw}\}$ &
        $\{\sfw, \textbf{\sfnw}\}$ &
        $\{\sfw, \textbf{\sfnw}\}$ &
        $\{\sfw, \textbf{\sfnw}\}$ &
        $\{\textbf{\sfw}, \sfnw\}$ \\  
\hline
\multirow{3}{*}{\thead{RM}} & $r$ &
        $\thead{\{\textbf{20}, 40, 60\}}$ &
        $\thead{\{\textbf{10}, 20, 40, 60\}}$ &
        $\thead{\{\textbf{20}, 40, 60\}}$ &
        $\thead{\{\textbf{20}, 40, 60\}}$ &
        $\thead{\{20, \textbf{40}, 60\}}$  \\
    {}  & $\sigma$ &
        $\{\textbf{0.1}, 0.2, 0.3\}$ &
        $\{\textbf{0.1}, 0.2, 0.3\}$ &
        $\{\textbf{0.1}, 0.2, 0.3\}$ &
        $\{\textbf{0.1}, 0.2, 0.3\}$ &
        $\{0.1, 0.2, \textbf{0.3}\}$ \\
    {}  & $d$ & 
        $\thead{\{25, 50, 75, \textbf{100}\}}$ &
        $\thead{\{\textbf{50}, 75, 100\}}$ &
        $\thead{\{\textbf{25}, 50, 75, 100\}}$ &
        $\thead{\{\textbf{25}, 50, 75, 100\}}$ &
        $\thead{\{25, 50, \textbf{75}, 100\}}$ \\
\hline
\multirow{2}{*}{PBOW}   & $N$ & 
        \thead{$\{10,20,...50,$\\$60, \textbf{80},100\}$} & 
        \thead{$\{10,20,40,60,$\\$80,\textbf{100},...200\}$} & 
        \thead{$\{10,20,...60,$\\$80,\textbf{100}...200\}$} & 
        \thead{$\{10,20,...50,$\\$60,80,\textbf{100}\}$} & 
        \thead{$\{10,20,...\textbf{50},$\\$60,80,100\}$}
        \\
    {}  & $\sfw$ & 
        $\{\sfw, \textbf{\sfnw}\}$ &
        $\{\sfw, \textbf{\sfnw}\}$ &
        $\{\sfw, \textbf{\sfnw}\}$ &
        $\{\sfw, \textbf{\sfnw}\}$ &
        $\{\sfw, \textbf{\sfnw}\}$ \\ 
    {} & $S$ &
        \thead{$\{\textbf{2000}, 5000, 10000\}$} &
        \thead{$\{2000, 10000, \textbf{50000}\}$} &
        \thead{$\{\textbf{2000}, 10000, 50000\}$} &
        \thead{$\{5000, 10000, \textbf{50000}\}$} &
        \thead{$\{\textbf{5000}, 10000, 50000\}$} \\
\hline
\multirow{2}{*}{wPBOW}   & $N$ & 
        \thead{$\{10,20,...\textbf{50},$\\$60, 80,100\}$} & 
        \thead{$\{10,20,40,60,$\\$\textbf{80},...200\}$} & 
        \thead{$\{10,20,...\textbf{50},60,$\\$80,...200\}$} & 
        \thead{$\{10,20,...50,$\\$60,80,\textbf{100}\}$} & 
        \thead{$\{10,20,...50,$\\$60,80,\textbf{100}\}$}
        \\
    {}  & $\sfw$ & 
        $\{\sfw, \textbf{\sfnw}\}$ &
        $\{\sfw, \textbf{\sfnw}\}$ &
        $\{\sfw, \textbf{\sfnw}\}$ &
        $\{\sfw, \textbf{\sfnw}\}$ &
        $\{\textbf{\sfw}, \sfnw\}$ \\
    {} & $S$ &
        \thead{$\{\textbf{2000}, 5000, 10000\}$} &
        \thead{$\{\textbf{2000}, 10000, 50000\}$} &
        \thead{$\{\textbf{2000}, 10000, 50000\}$} &
        \thead{$\{\textbf{5000}, 10000, 50000\}$} &
        \thead{$\{\textbf{5000}, 10000, 50000\}$} \\
\hline
\multirow{2}{*}{sPBOW}   & $N$ & 
        \thead{$\{\textbf{10},20,...50,$\\$60, 80,100\}$} & 
        \thead{$\{10,\textbf{20},40,60, ...200\}$} & 
        \thead{$\{10,20,\textbf{30}...60,$\\$80,...200\}$} & 
        \thead{$\{10,20,...50,$\\$60,80,\textbf{100}\}$} & 
        \thead{$\{10,20,...50,$\\$60,80,\textbf{100}\}$}
        \\
    {}  & $\sfw$ & 
        $\{\textbf{\sfw}, \sfnw\}$ &
        $\{\textbf{\sfw}, \sfnw\}$ &
        $\{\sfw, \textbf{\sfnw}\}$ &
        $\{\sfw, \textbf{\sfnw}\}$ &
        $\{\textbf{\sfw}, \sfnw\}$ \\
    {} & $S$ &
        \thead{$\{2000, 5000, \textbf{10000}\}$} &
        \thead{$\{\textbf{2000}, 10000, 50000\}$} &
        \thead{$\{2000, \textbf{10000}, 50000\}$} &
        \thead{$\{5000, \textbf{10000}, 50000\}$} &
        \thead{$\{5000, \textbf{10000}, 50000\}$} \\
\hline
\multirow{2}{*}{PVLAD}   & $N$ & 
        \thead{$\{10,20,...50,$\\$\textbf{60}, 80,100\}$} & 
        \thead{$\{10,20,\textbf{40},60, ...200\}$} & 
        \thead{$\{10,20,...60,$\\$80,\textbf{100}...200\}$} & 
        \thead{$\{10,20,...50,$\\$60,80,\textbf{100}\}$} & 
        \thead{$\{10,20,\textbf{30},40,50,$\\$60,80,100\}$}
        \\
    {}  & $\sfw$ & 
        $\{\sfw, \textbf{\sfnw}\}$ &
        $\{\textbf{\sfw}, \sfnw\}$ &
        $\{\textbf{\sfw}, \sfnw\}$ &
        $\{\textbf{\sfw}, \sfnw\}$ &
        $\{\textbf{\sfw}, \sfnw\}$ \\
    {} & $S$ &
        \thead{$\{\textbf{2000}, 5000, 10000\}$} &
        \thead{$\{2000, \textbf{10000}, 50000\}$} &
        \thead{$\{2000, 10000, \textbf{50000}\}$} &
        \thead{$\{5000, \textbf{10000}, 50000\}$} &
        \thead{$\{\textbf{5000}, 10000, 50000\}$} \\
\hline
\multirow{2}{*}{sPVLAD}   & $N$ & 
        \thead{$\{10,20,30,...50,$\\$60, 80,\textbf{100}\}$} & 
        \thead{$\{\textbf{10},20,40,60, ...200\}$} & 
        \thead{$\{10,20,...60,$\\$80,...\textbf{160},...200\}$} & 
        \thead{$\{10,20,...50,$\\$60,\textbf{80},100\}$} & 
        \thead{$\{10,20,...50,$\\$60,\textbf{80},100\}$}
        \\
    {}  & $\sfw$ & 
        $\{\sfw, \textbf{\sfnw}\}$ &
        $\{\textbf{\sfw}, \sfnw\}$ &
        $\{\textbf{\sfw}, \sfnw\}$ &
        $\{\textbf{\sfw}, \sfnw\}$ &
        $\{\textbf{\sfw}, \sfnw\}$ \\
    {} & $S$ &
        \thead{$\{\textbf{2000}, 5000, 10000\}$} &
        \thead{$\{2000, \textbf{10000}, 50000\}$} &
        \thead{$\{2000, 10000, \textbf{50000}\}$} &
        \thead{$\{5000, 10000, \textbf{50000}\}$} &
        \thead{$\{5000, 10000, \textbf{50000}\}$} \\
\hline
\multirow{2}{*}{PFV}   & $N$ & 
        \thead{$\{10,20,...50,$\\$60, 80,\textbf{100}\}$} & 
        \thead{$\{\textbf{10},20,40,60, ...200\}$} & 
        \thead{$\{10,20,\textbf{30},40,50,$\\$60,80,...200\}$} & 
        \thead{$\{\textbf{10},20,...50,$\\$60,80,100\}$} & 
        \thead{$\{10,20,30,\textbf{40},50,$\\$60,80,100\}$}
        \\
    {}  & $\sfw$ & 
        $\{\sfw, \textbf{\sfnw}\}$ &
        $\{\textbf{\sfw}, \sfnw\}$ &
        $\{\textbf{\sfw}, \sfnw\}$ &
        $\{\textbf{\sfw}, \sfnw\}$ &
        $\{\textbf{\sfw}, \sfnw\}$ \\
    {} & $S$ &
        \thead{$\{\textbf{2000}, 5000, 10000\}$} &
        \thead{$\{2000, \textbf{10000}, 50000\}$} &
        \thead{$\{2000, 10000, \textbf{50000}\}$} &
        \thead{$\{5000, 10000, \textbf{50000}\}$} &
        \thead{$\{5000, 10000, \textbf{50000}\}$} \\
\hline
\end{tabularx}
\caption{Parameters tested for EXP-B.}

\label{tab:vectorParameters}
\end{sc}
\end{tiny}
\end{sidewaystable}

\subsection{Accuracy vs. Codebook Size and weighted sub-sampling}
\label{subsec:resultAccuracyvsCodebookSize}

\begin{figure}
\begin{tabular}{c c}
\includegraphics[width=0.45\linewidth]{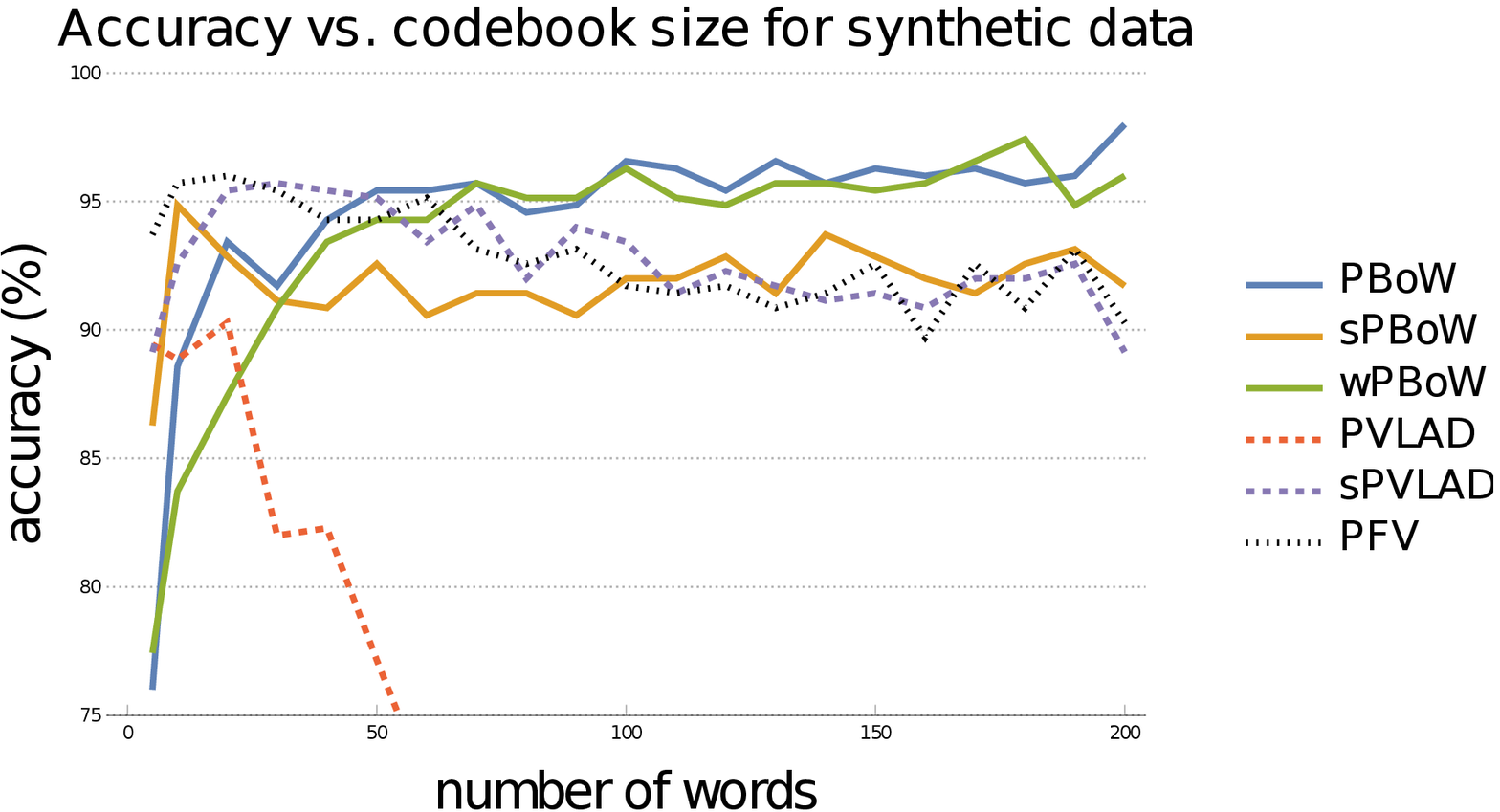} &
\includegraphics[width=0.45\linewidth]{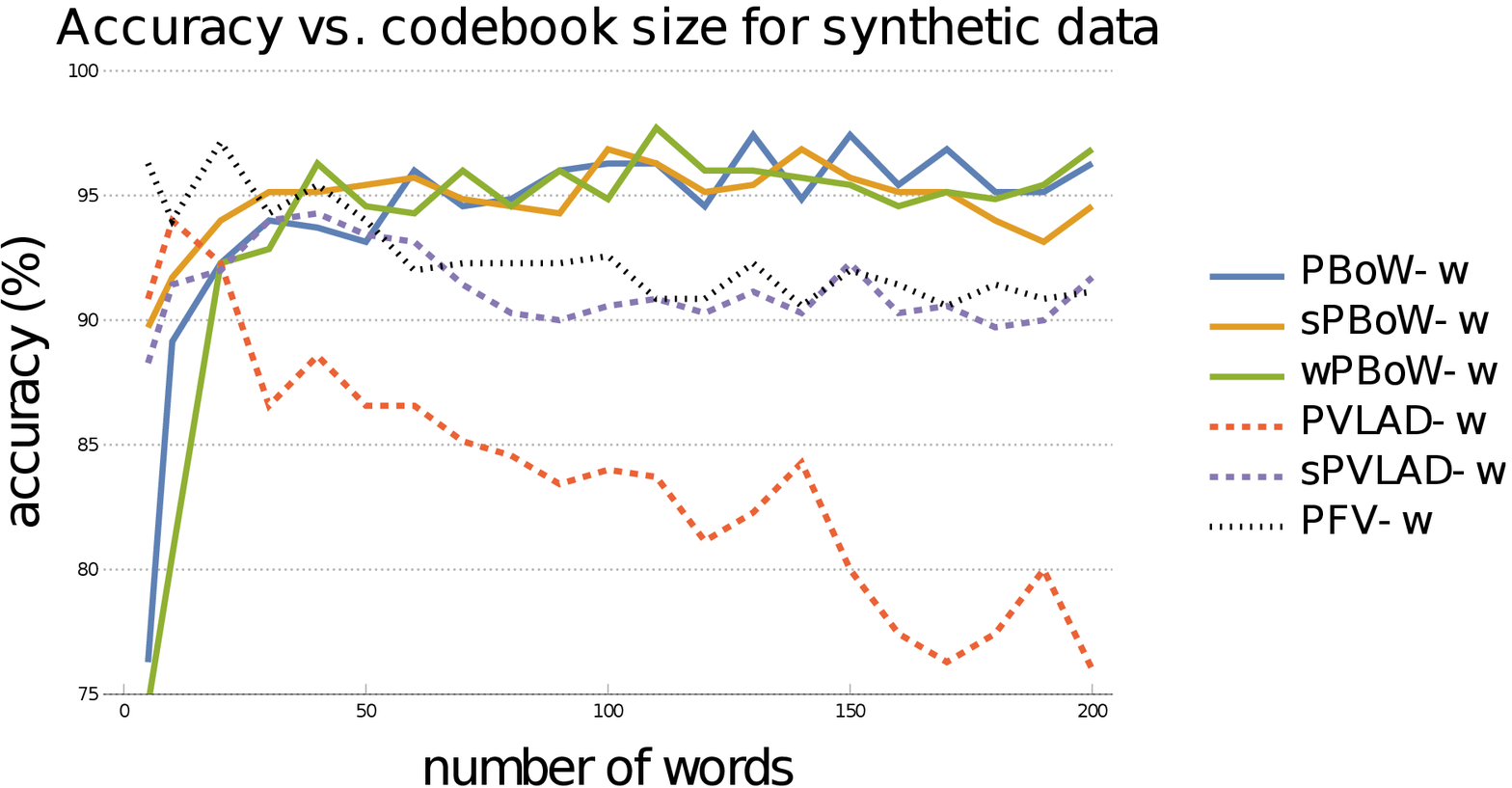} \\
\includegraphics[width=0.45\linewidth]{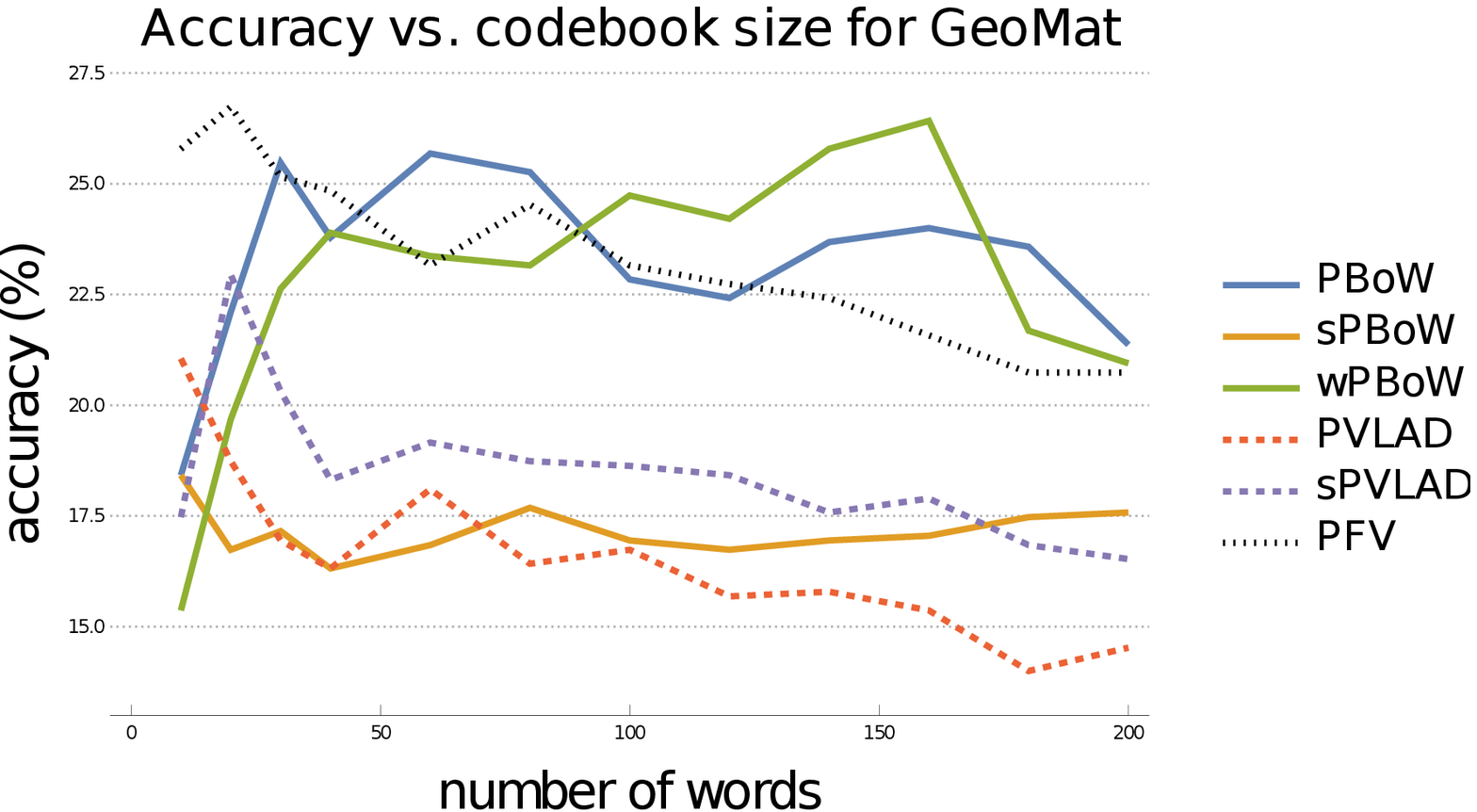} &
\includegraphics[width=0.45\linewidth]{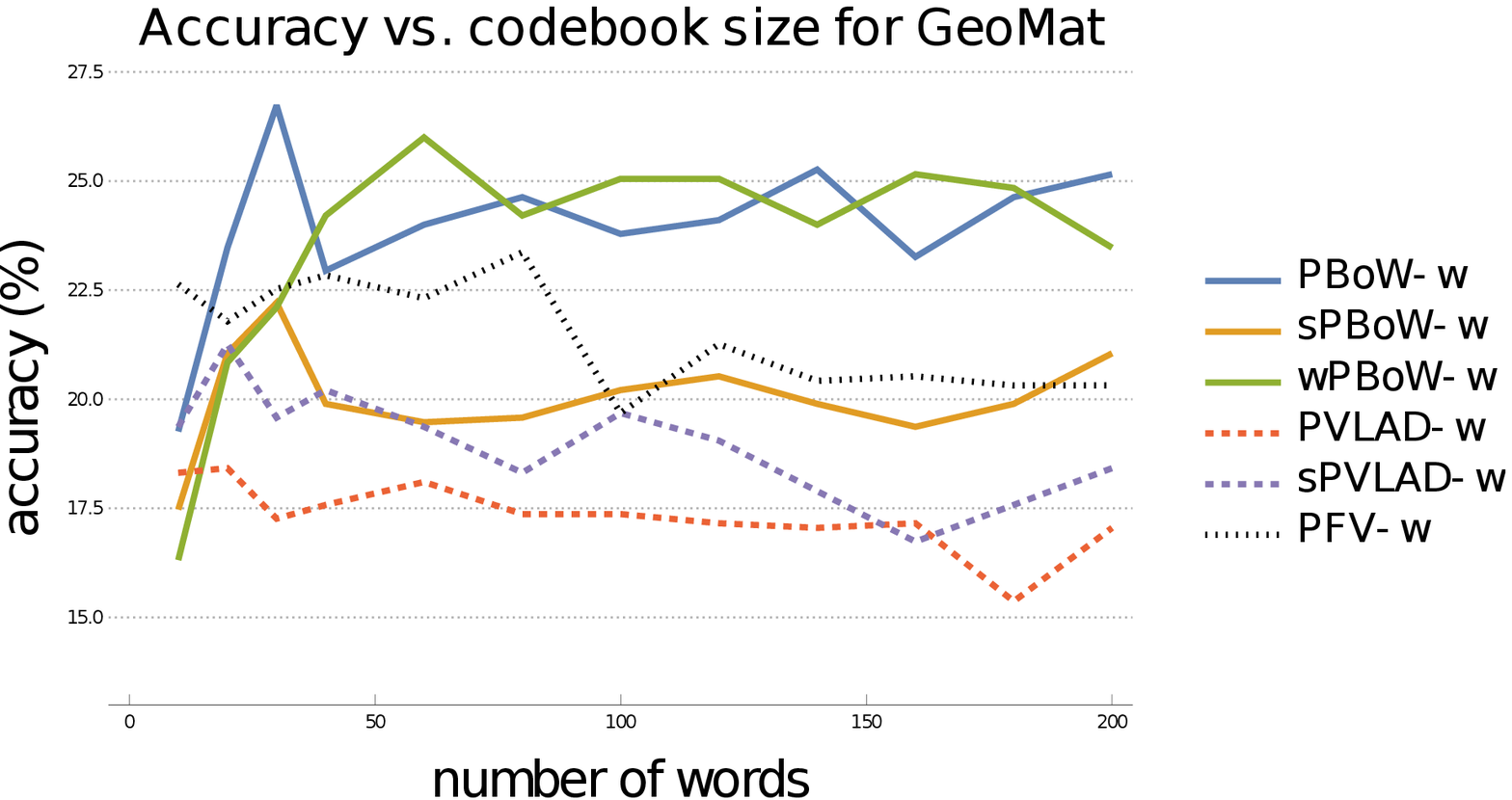} \\
\includegraphics[width=0.45\linewidth]{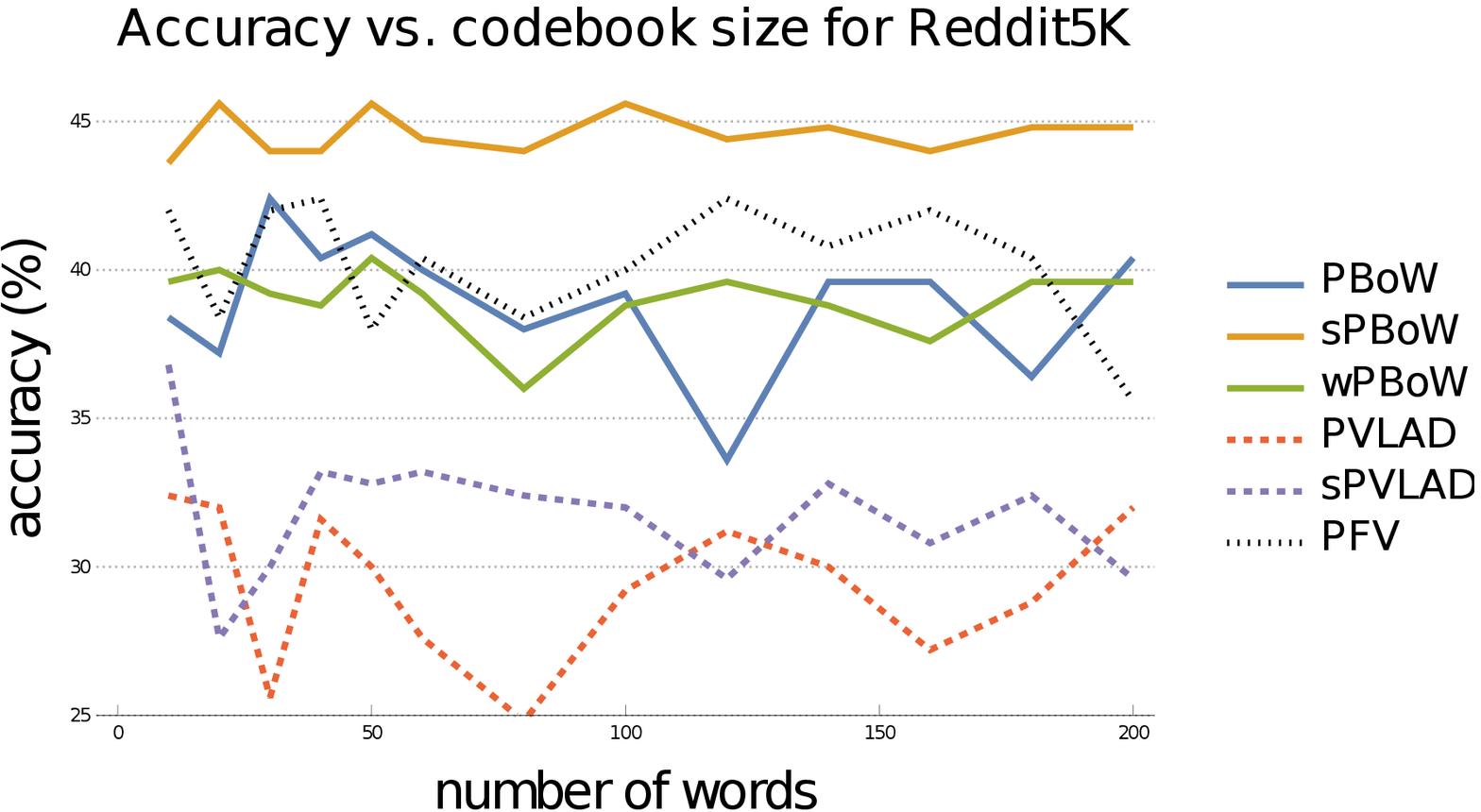} &
\includegraphics[width=0.45\linewidth]{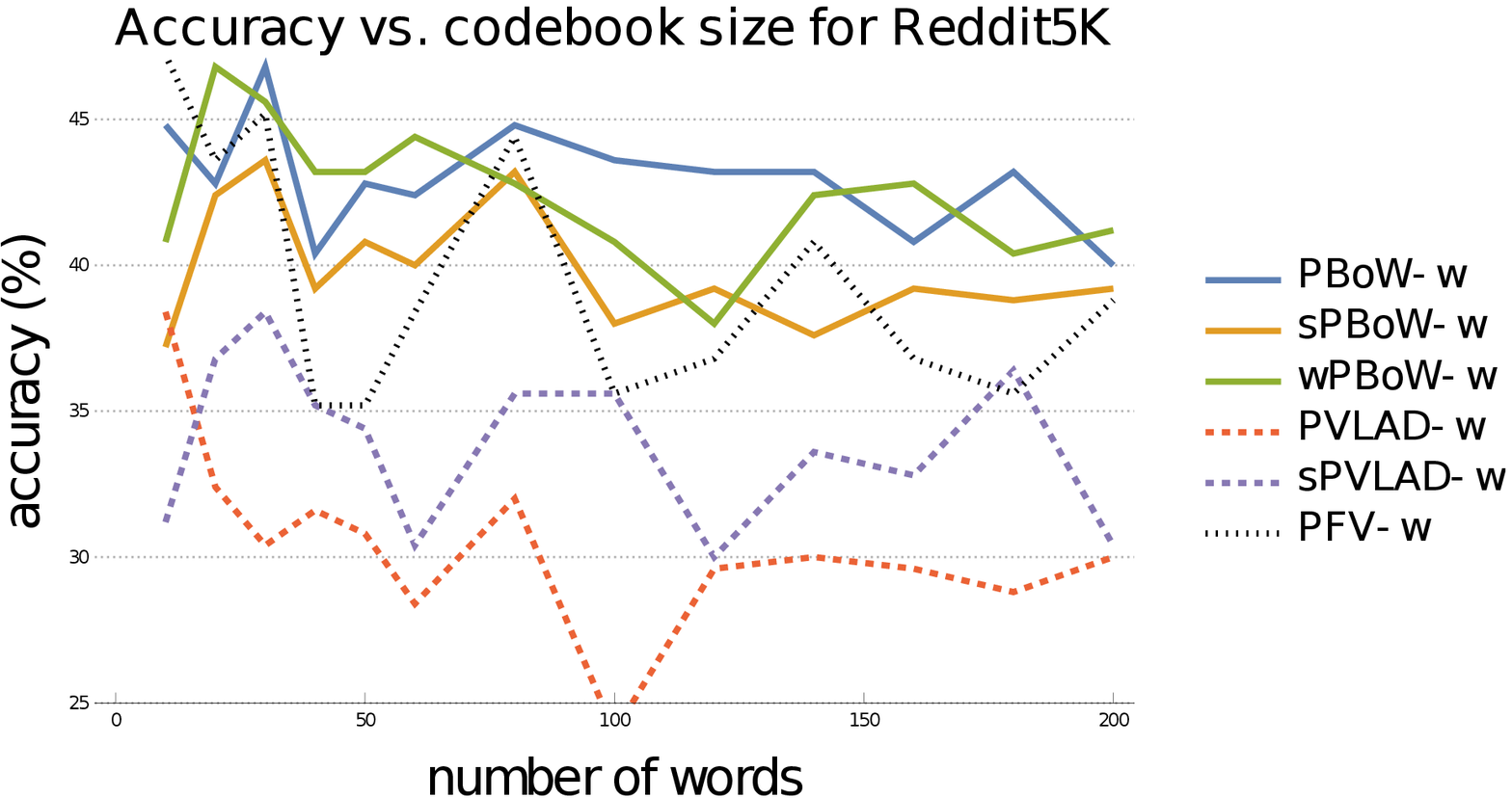} \\
\includegraphics[width=0.45\linewidth]{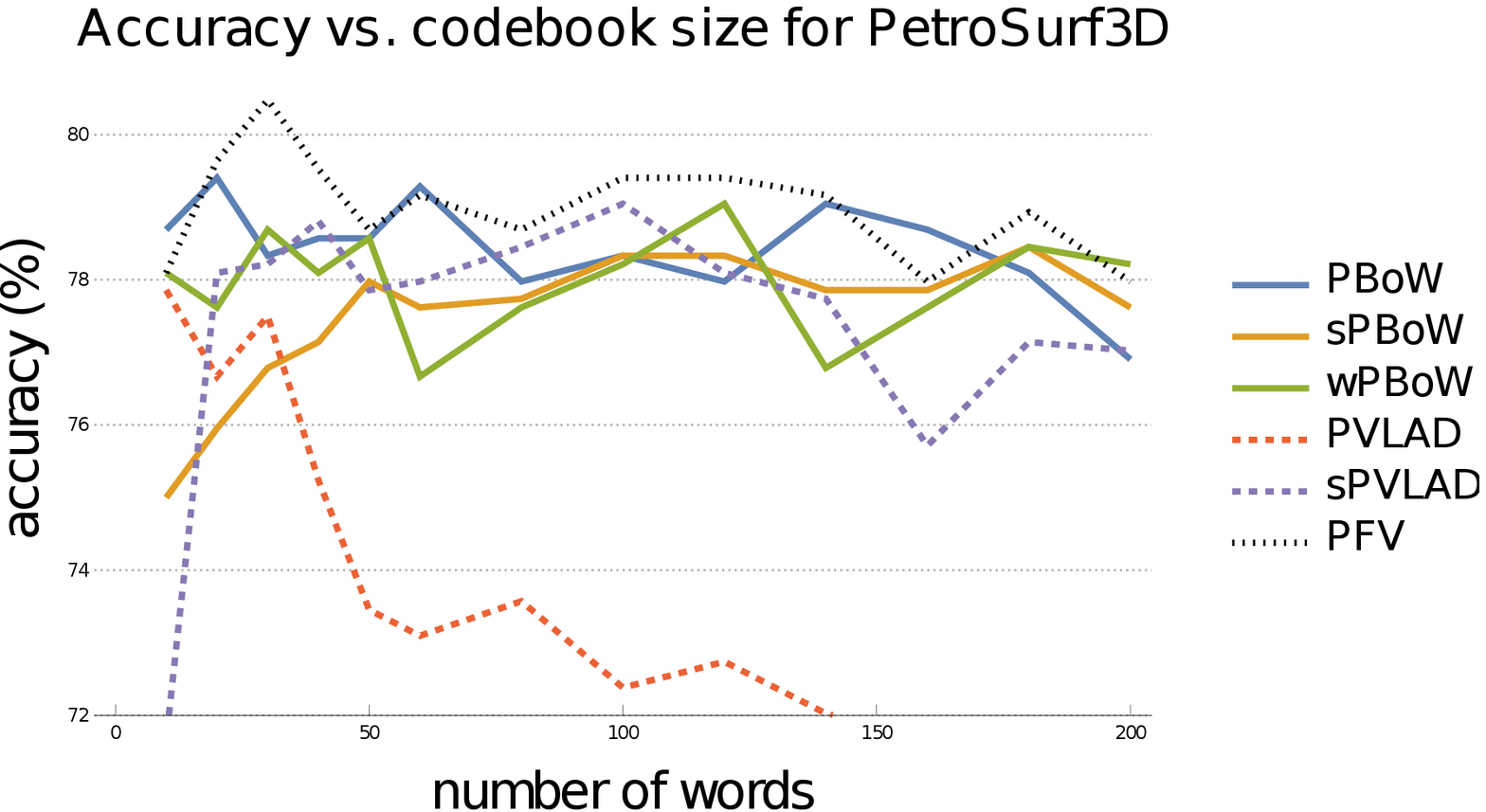} &
\includegraphics[width=0.45\linewidth]{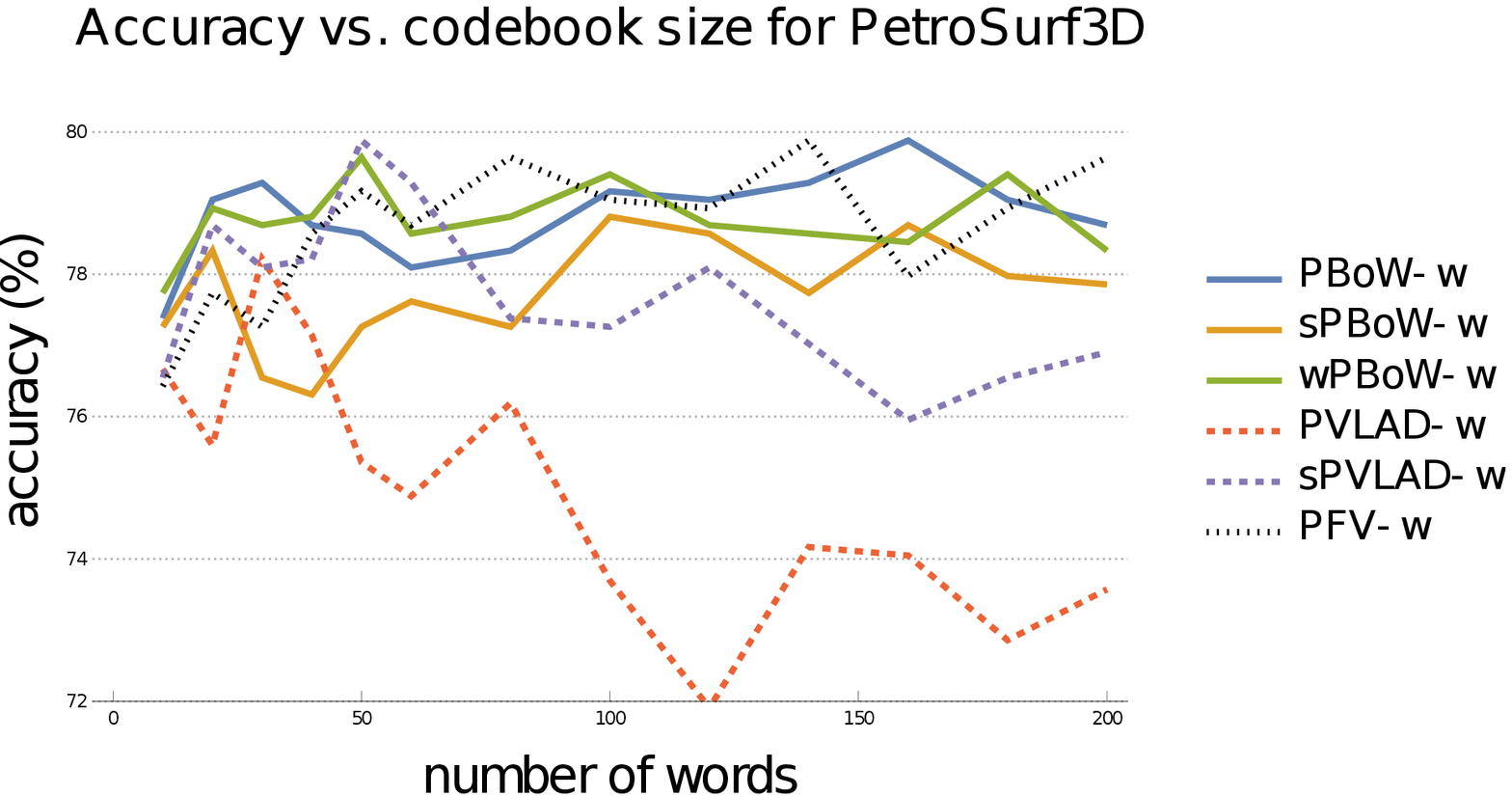} \\
\includegraphics[width=0.45\linewidth]{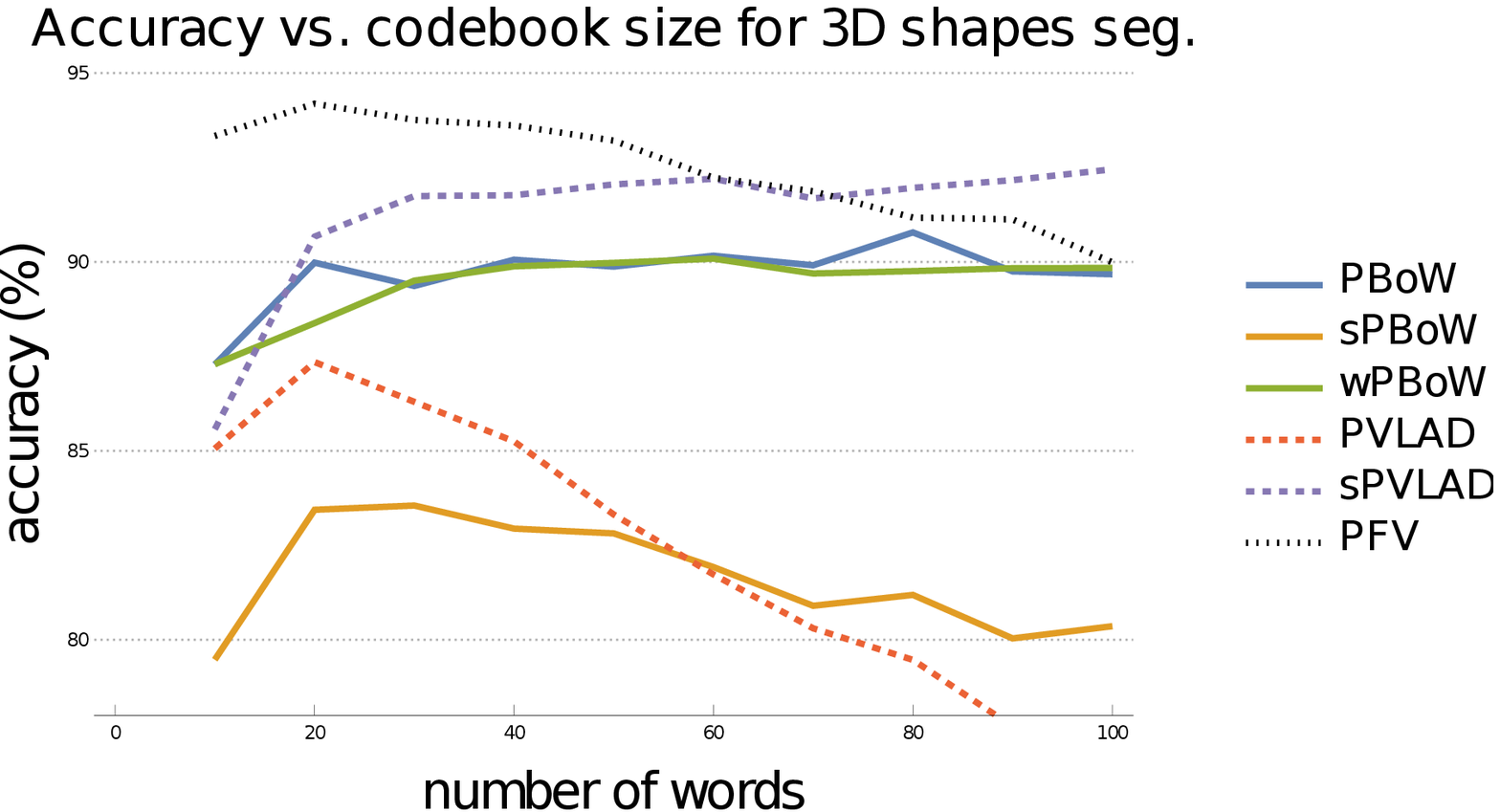} &
\includegraphics[width=0.45\linewidth]{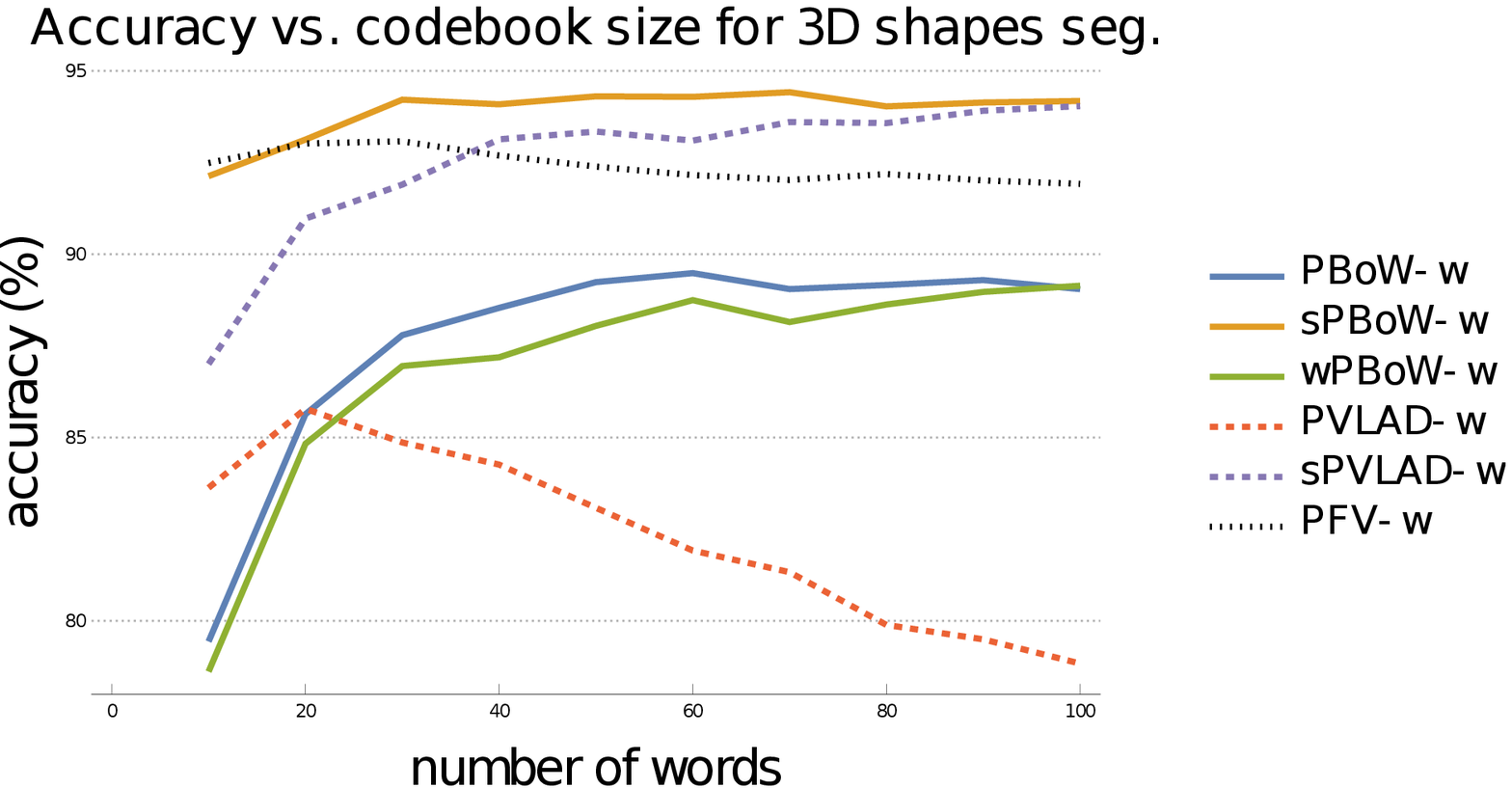}
\end{tabular}
\caption{Accuracy vs. size of a codebook for datasets from EXP-A without (left column) and with codebook weighting (right column) in codebook generation.}
\label{fig:acc_words}
\end{figure}

The most important parameter for codebook-based representations is the codebook size $N$, i.e. the number of clusters. There is no commonly agreed analytic method to estimate the~optimal codebook size; thus, the estimation is usually performed empirically. To investigate the sensitivity of codebook approaches and their performance on the codebook size, each approach was evaluated for a sequence of $N$ values (see Tables~\ref{tab:kernelParameters} and~\ref{tab:vectorParameters}). The results are presented in Fig.~\ref{fig:acc_words}, both without (left column) and with weighted sub-sampling (right column) of the consolidated PD.

We can observe that all three variants of PBoW (PBoW, wPBoW and sPBoW) reach optimal performance at the level of about 50 words in a codebook (or earlier), a further increase of  codebook size does not necessarily further improve its efficiency. In some cases, there is a slight improvement (synthetic data), in other cases performance goes down slightly (GeoMat) or remains a the same level. This shows that the  codebook size is not rather insensitive parameter, once a certain minimum size is surpassed.

The remaining approaches (PVLAD, sPVLAD and PFV) show a general tendency to achieve their best performance early, with codebooks containing less than 40 words, and after that the accuracy drops substantially. The most prominent example of this behavior is depicted by the PVLAD method. It is caused by the fact that in cases where there is just a few clusters, it is simpler to capture well both the zeroth and the first moments, because clusters occupy large regions. However, once the number of clusters gets larger, cluster size shrinks and the assignment gets unstable.

 Fig.~\ref{fig:acc_words} further shows the effect of weighting during sub-sampling for codebook generation. This can be best observed from the performance  curves of sPBoW, where the effect is largest. For 3D shapes, weighting leads to a dramatic improvement in accuracy. For for synthetic data, GeoMat and PetroSurf3D, there is also a moderate improvement in performance. Only for Reddit5K, weighted subsampling degrades performance. For other methods, however, the introduction of weighted subsampling does yield an improvement on the  Reddit5K experiment (see e.g. PBoW, wPBoW and sPVLAD). In the majority of cases, weighted subsampling has a positive impact on performance. For PVLAD, weighted subsampling even seems to compensate for weaknesses of the representation in situations where codebook sizes are large.

Overall, we conclude, that optimal and universal choice for codebook size is about 50 in case of PBoW, wPBoW and sPBoW; while for the remaining methods, 20 words seems to be sufficient. These values are thus good starting points for hyperparameter optimization on other datasets. The choice of weighted vs. non-weighted subsampling seems to be dataset dependent. For 3D shapes, for example, strong performance gains are achieved. For~the~other datasets the trend is not so clear.

\subsection{Accuracy vs. Time}
\label{subsec:resultsAccTime}

Tables~\ref{tab:kernelResults}~and~\ref{tab:vectorResults} show that our approaches achieve state-of-the-art performance or even better performance on almost all of the evaluated datasets. Furthermore, they outperform all methods in speed. While above tables show only results for the optimal parameters (from the  point of view of classification accuracy), here, we analyze the relation between accuracy and computation time in more detail. For this purpose we use  PI as a reference for comparison, as it represents the strongest competitor (in the sense of accuracy) of the~proposed representations. 

\begin{figure}
\centering
\begin{tabular}{c c}
\includegraphics[width=0.47\linewidth]{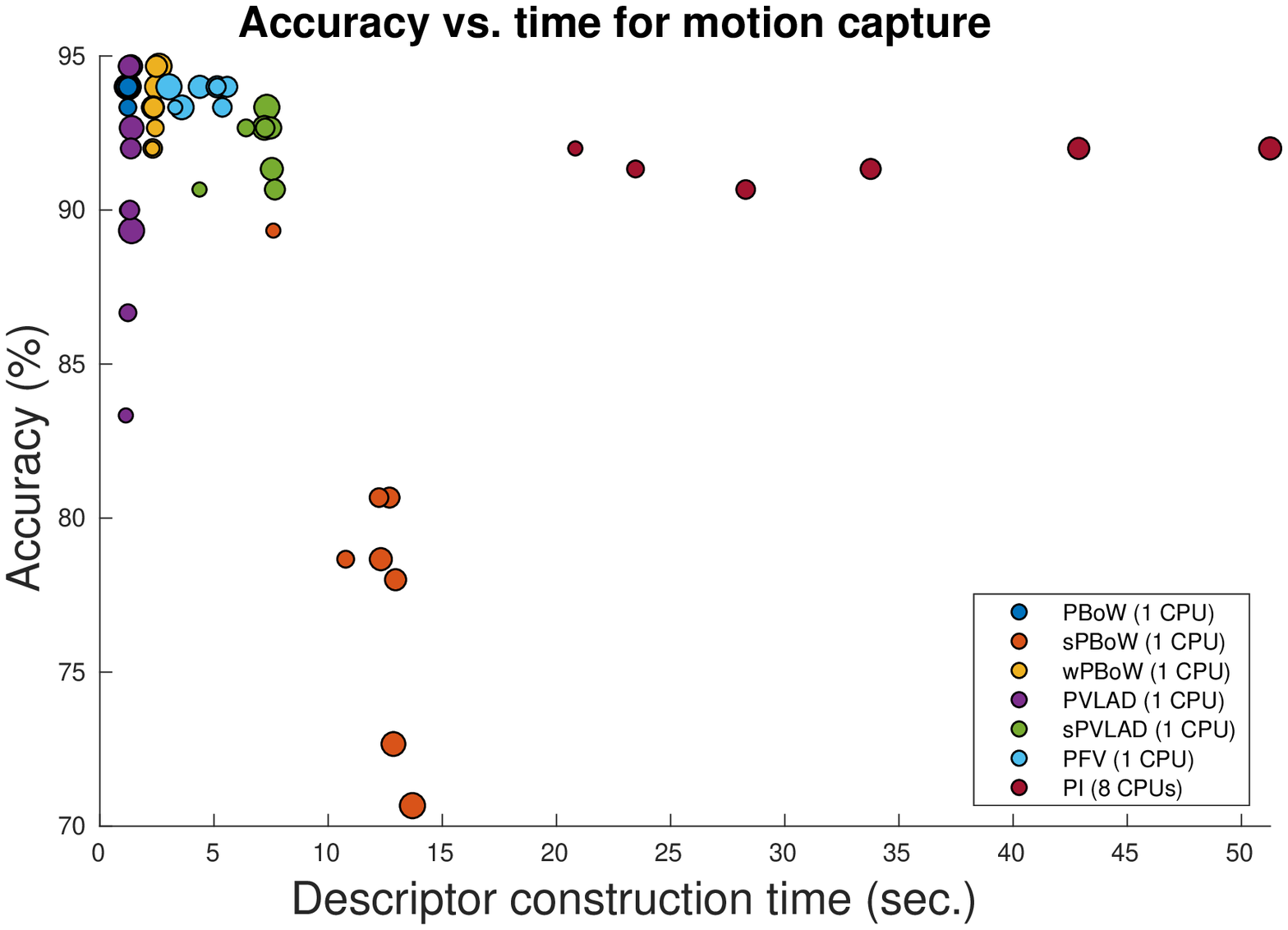} &
\includegraphics[width=0.47\linewidth]{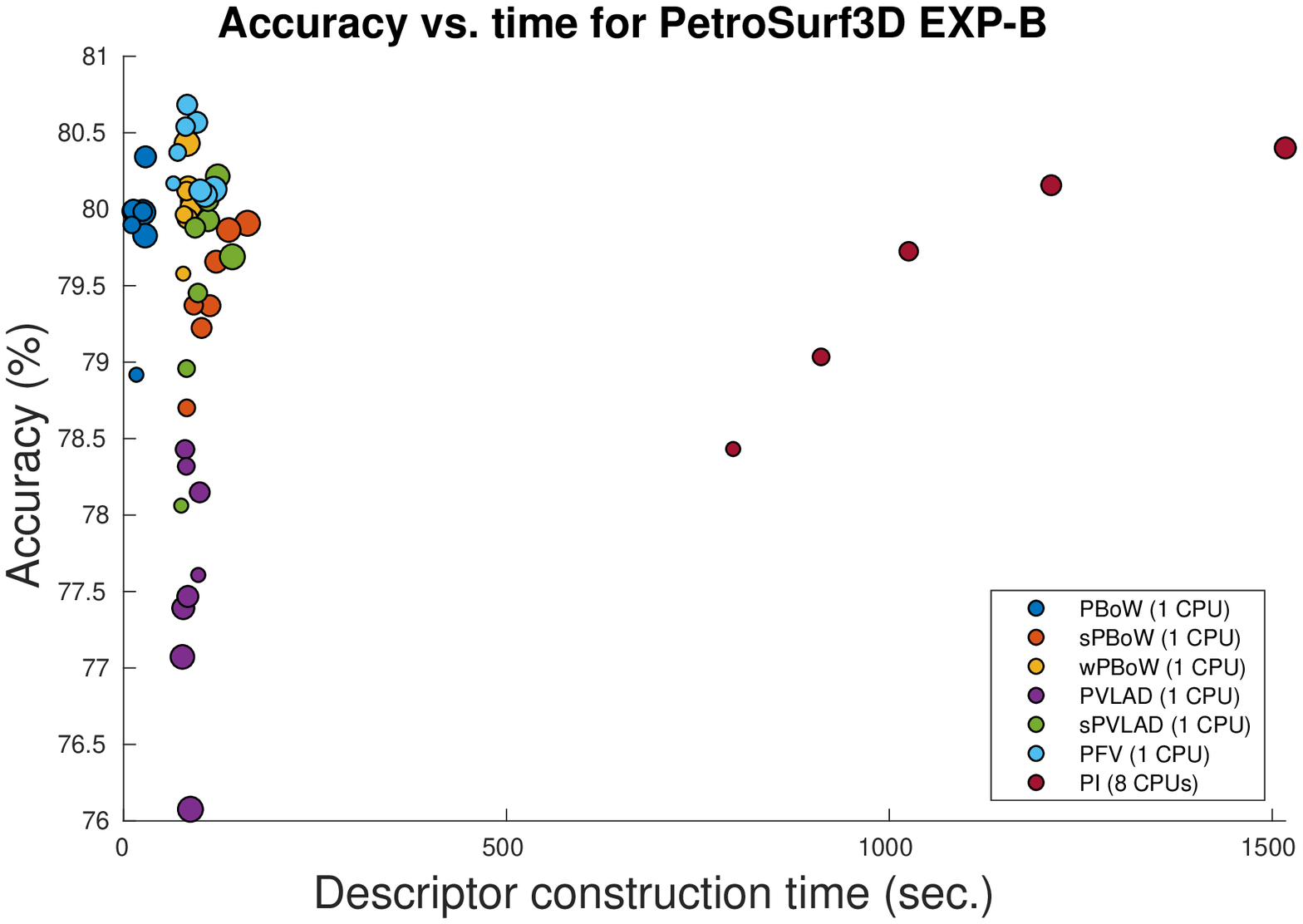} \\
\includegraphics[width=0.47\linewidth]{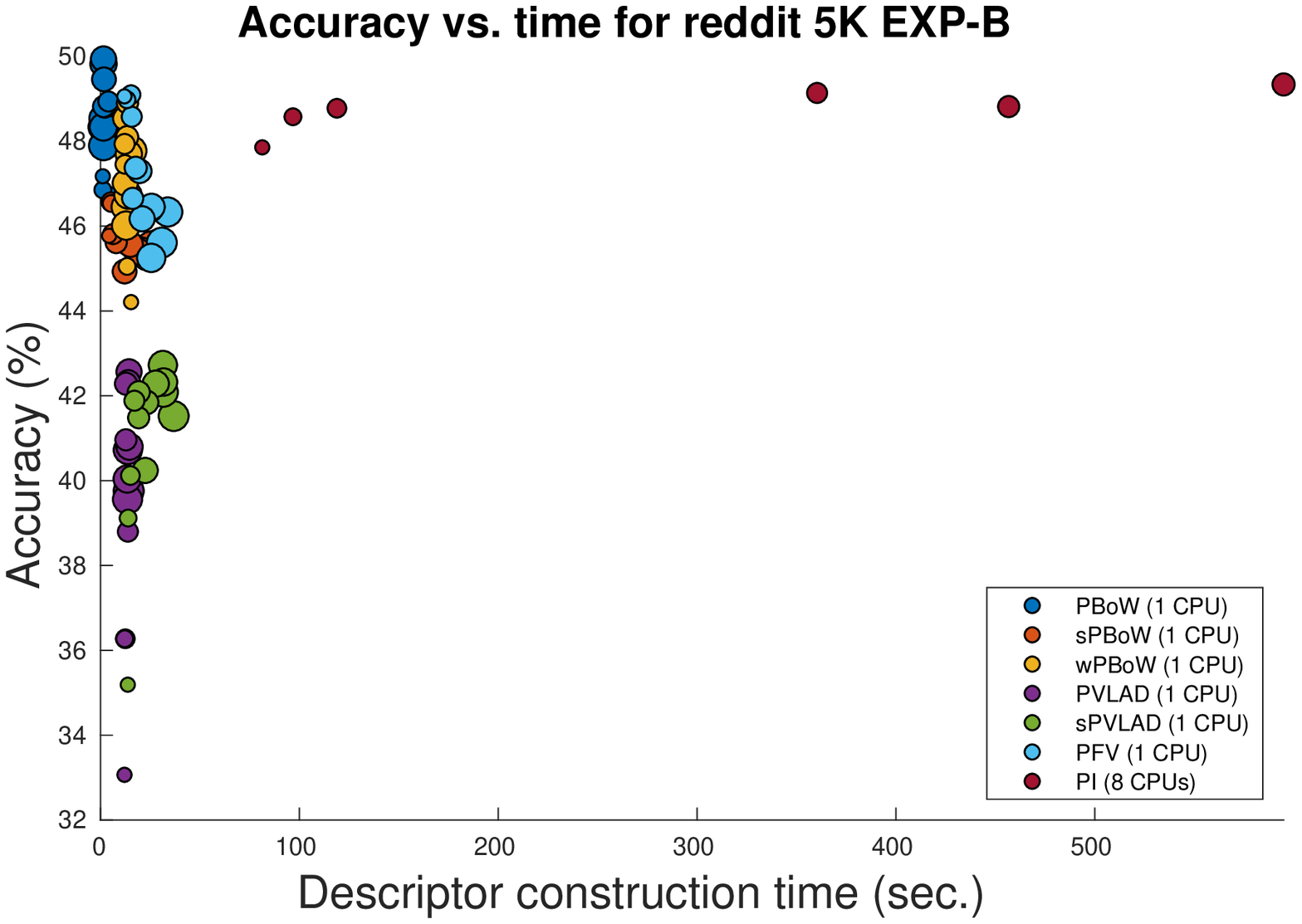} &
\includegraphics[width=0.47\linewidth]{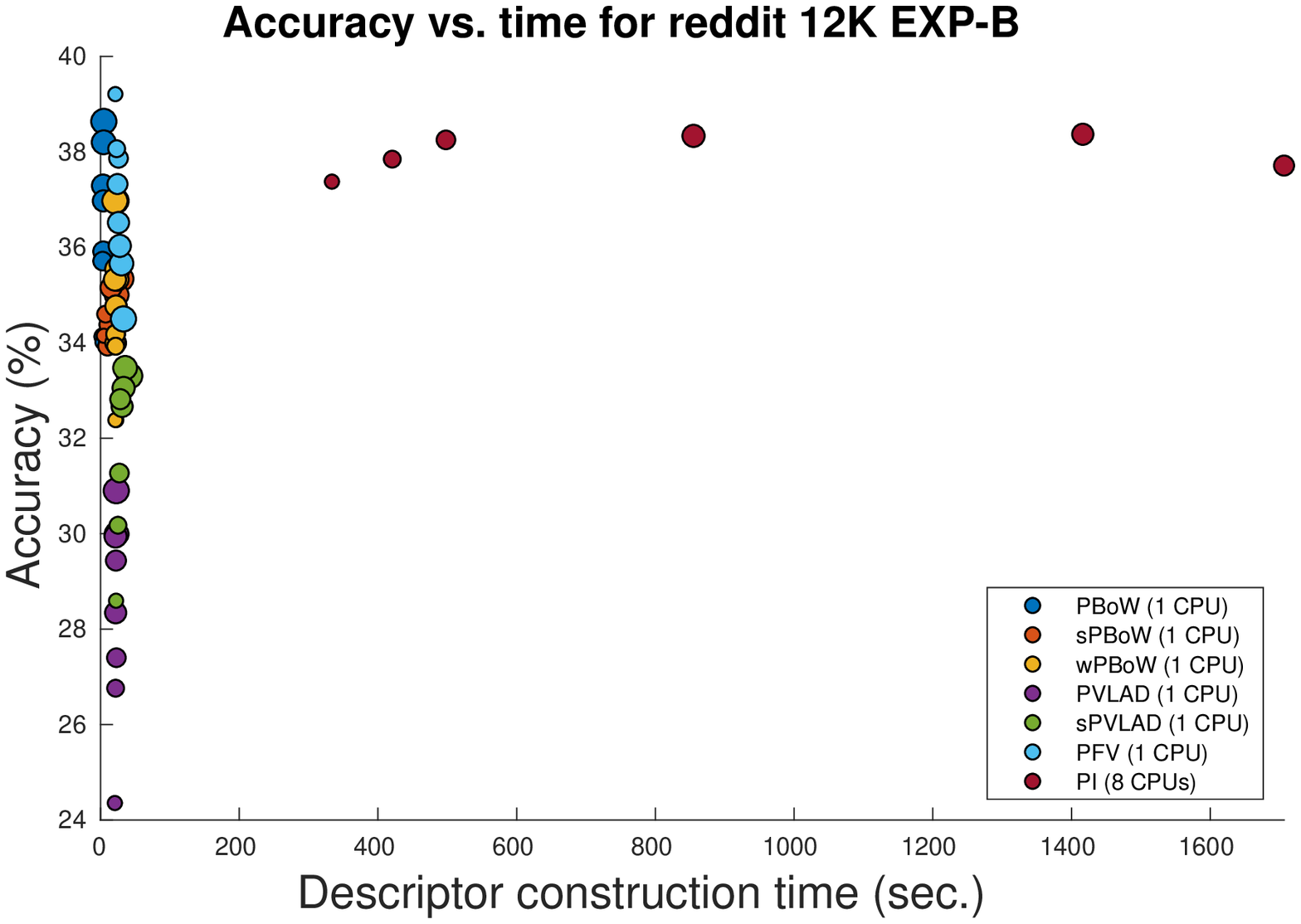} \\
\includegraphics[width=0.47\linewidth]{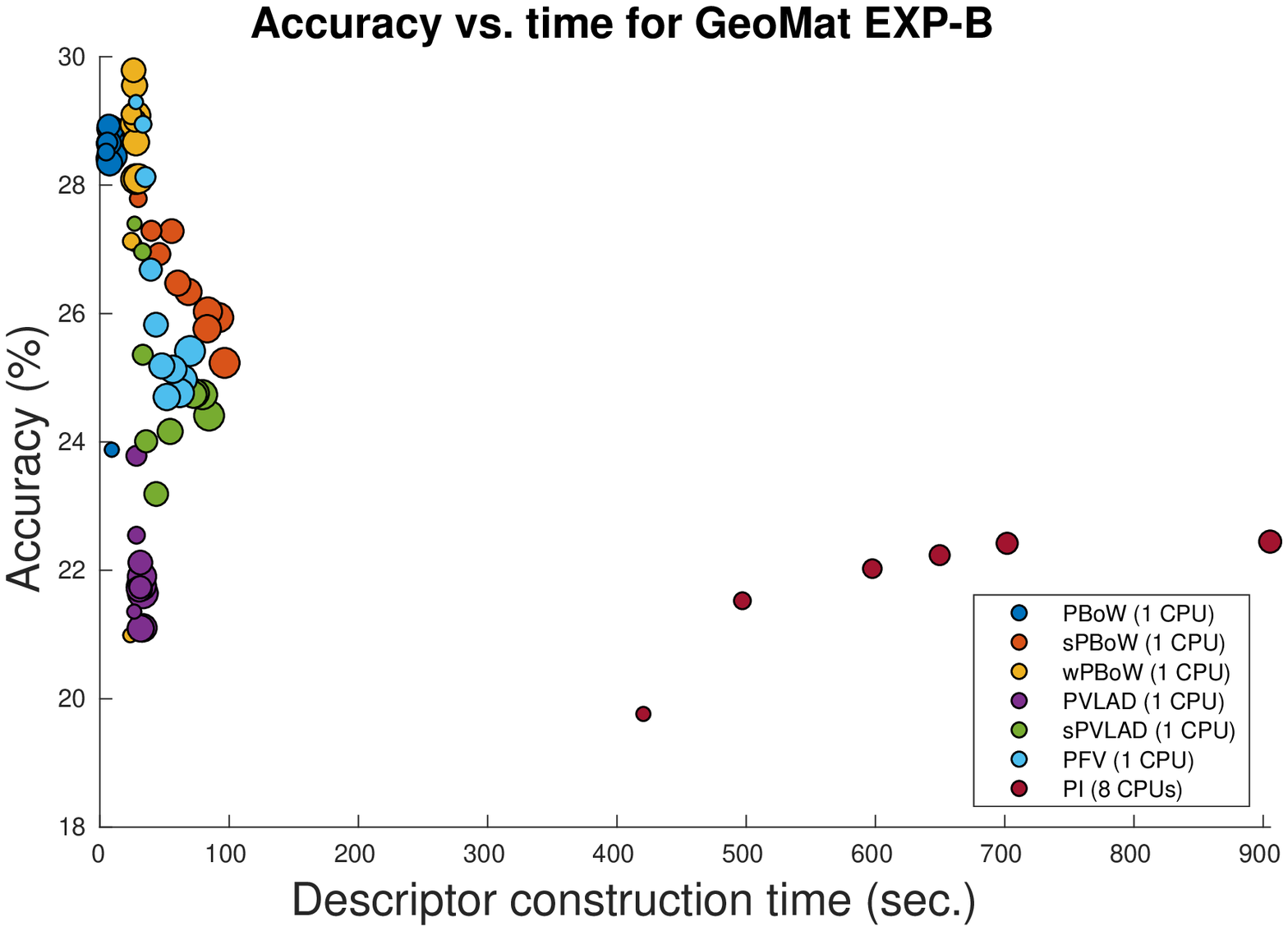} & 
\includegraphics[width=0.47\linewidth]{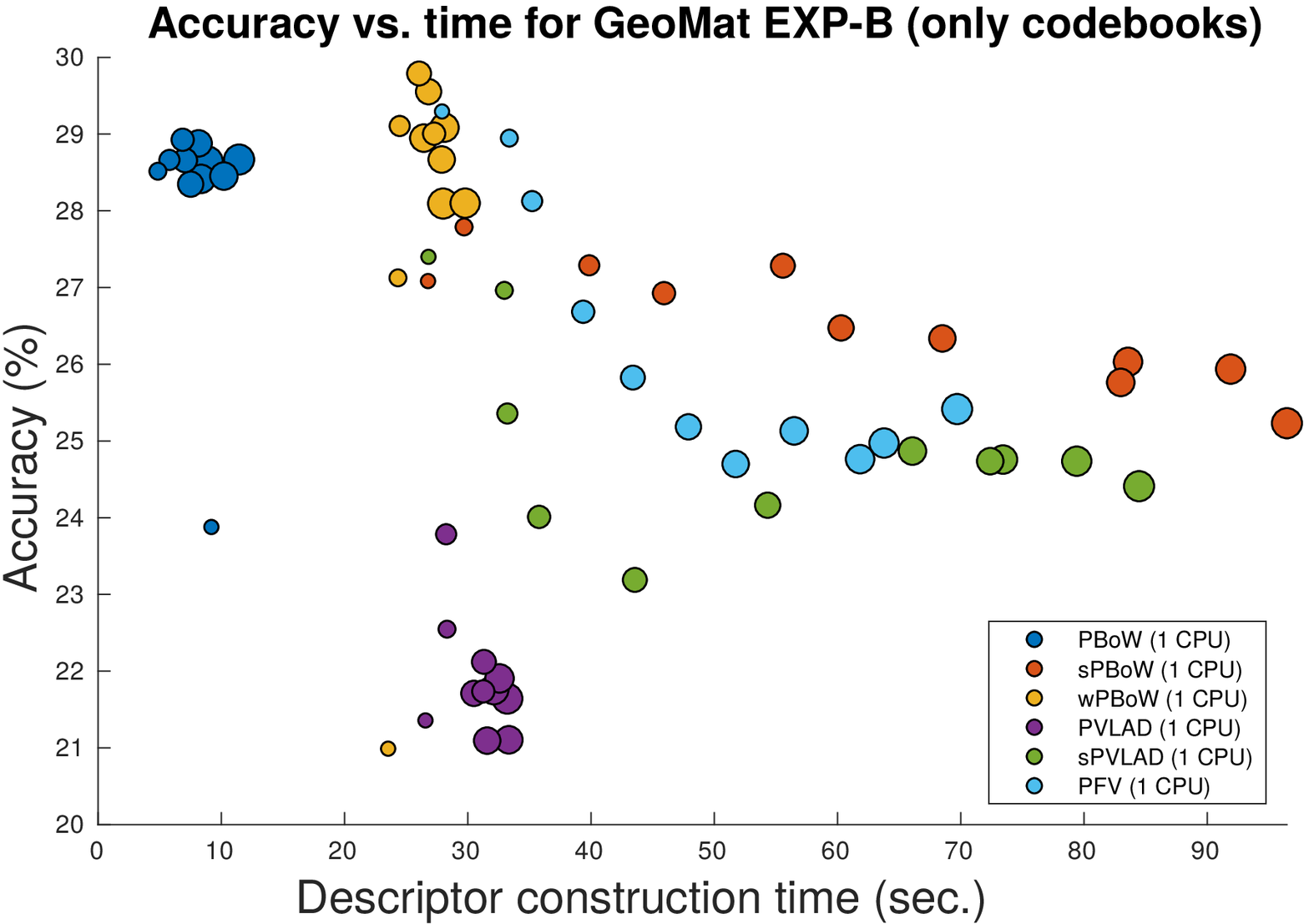}
\end{tabular}
\caption{Accuracy vs. time for codebook approaches compared with PI (the strongest related representation) applied to all datasets from EXP-B. The size of colored points represents the size $N$ of codebooks or the resolution $r$ of PI (evaluated values for $N$ and $r$ are those listed in Table~\ref{tab:vectorParameters}). Note that the actual times of computation for the construction of the representations are presented. Codebooks were computed on 1 CPU, while PI was constructed by using 8 CPUs in parallel. Moreover, SVM training and prediction time was not taken into consideration. This would further increase computational times, especially for PI due to their larger dimension. The bottom-right plot shows results only for codebook approaches in case of GeoMat dataset (for better visibility).}
\label{fig:time_score}
\end{figure}

In Fig.~\ref{fig:time_score}, we plot accuracy vs. time for the proposed approaches and PI for all datasets from EXP-B. We decided to focus on EXP-B here, because it operates on larger datasets than EXP-A and is thus better suited to study runtime efficienty. We vary the parameters with most influence placed on runtime (codebook size for persistent codebooks as well as the resolution of PI) according to the values provided in Table~\ref{tab:vectorParameters}. This directly influences the output dimension of the representation and is reflected by the area of the circles in Fig.~\ref{fig:time_score}, i.e. larger diameter means higher dimension. Note that experiments with codebooks were performed on 1 CPU, while experiments on PI were performed in parallel on 8 CPUs. Therefore, the total runtime differences are in fact even larger than depicted. For more compact visualization (and avoiding a logarithmic scale which would compress too much), we decided not to take the number of CPUs into account for plotting. We can see clearly that the runtime of PI is always significally larger than any codebook representation. The~accuracy obtained varies. For all datasets the performance level of PI is reached (or even superseeded) much quicker. In the case of GeoMat dataset, codebooks clearly outperform PI (while consuming much less time); and in case of the other experiments, they quickly achieve a similar performance level. The computational cost of achieving a higher performance with PI is over-proportionally high, while the performance gain is actually rather limited (approx. $+1\%$). 

In order to study runtime difference only between codebook approaches, we again plotted results for GeoMat dataset without PI (bottom-right, Fig.~\ref{fig:time_score}). As expected, ordinary PBoW is fastest. We can observe that runtimes of PBoW, wPBoW and PVLAD are almost not affected by codebook size. The other approaches, i.e. sPBoW, sPVLAD and PFV, that involve Gaussians computation, clearly require much more time when codebook size is increased. However, as observed before, larger codebook sizes are not necessarily required to obtain good accuracy, which mitigates the situation. 

\subsection{Time vs. Dataset Size}

To investigate the runtime behavior of the proposed approaches in more detail, we evaluate how they scale to increasing dataset sizes (i.e. increasing numbers of input PDs). To this end, we employ the largest dataset in our experiments (PetroSurf3D) and randomly sample different numbers of PDs, starting from $1000$ to $10000$ in steps of $1000$. To get a detailed breakdown of computation time, we separately measure the time needed for codebook generation, histogram assignment, and classification. The computation of the~PDs is the same for all approaches, and thus is not included in this breakdown. 

\begin{figure}
\centering
\includegraphics[width=0.8\linewidth]{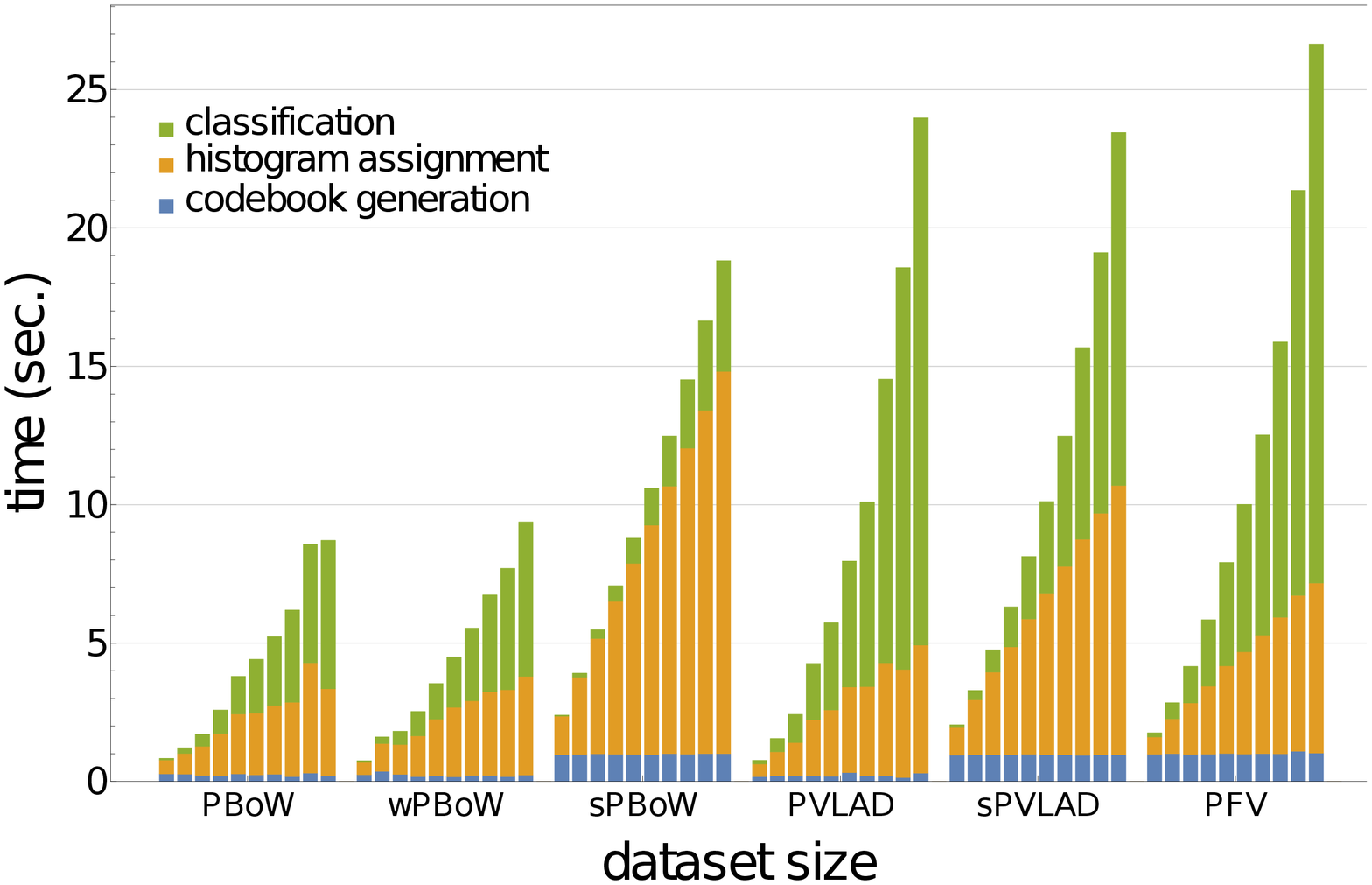}

\vspace{20pt}
\includegraphics[width=0.8\linewidth]{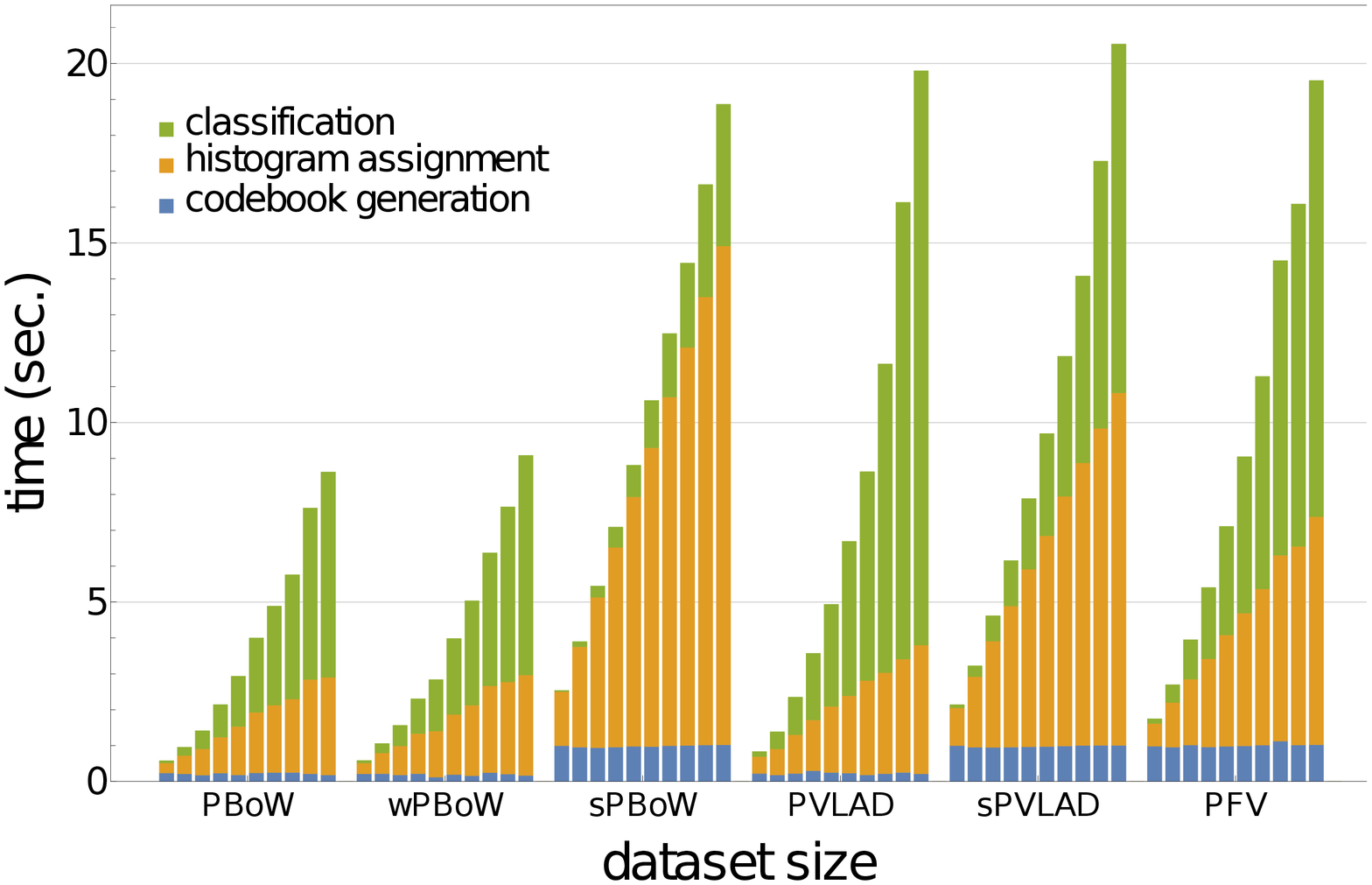}
\caption{Time vs. dataset size for all proposed persistent bag-of-words approaches (all with codebook size $N=50$). We measure the time needed for codebook generation (blue, bottom), histogram assignment (orange, middle), and classification (green, top), separately. Every consecutive bar represents an increasingly growing number of samples, from $1000, 2000, ..., 10000$. Upper figure shows results for unweighted methods, while those for the weighted versions are presented in the bottom.}
\label{fig:time_datasize}
\end{figure}

From the results presented in Fig.~\ref{fig:time_datasize}, we conclude that runtime grows almost linearly with dataset size. For the approaches \emph{without} weighted subsampling (upper part in Fig.~\ref{fig:time_datasize}), most of the computation time is spent on histogram assignment and classification. Histogram assignment takes more time for more complex encoding methods, such as PVLAD and PFV. In case of PBoW, histogram assignment is particularly fast because of k-d trees being used~\citep{bentley1975multidimensional}. For sPBoW, Gaussian likelihood has to be computed, which slows down the computation. Assignment time, however, grows linearly with dataset size. Classification time takes the major part for PVLAD and PFV. This is due to the fact that the computational complexity of both, primal and dual SVM optimization, depends on dimensionality~\citep{chapelle2007training}, which is higher in case of PVLAD and PFV. The~distribution of computation times is similar for the persistence codebook approaches \emph{with} weighted subsampling (lower part in Fig.~\ref{fig:time_datasize}).

\subsection{Qualitative Analysis}
\label{subsec:resultsPBoWforSyntheticDataset}

In this section, we investigate PBoW with a special focus on its discriminative abilities. For~this purpose, we employ a systhetic dataset as a proof-of-concept and GeoMat (for which we outperform other representations by a large margin) to investigate how this performance increase is achieved by PBoW compared to related approaches.

\subsubsection{Synthetic dataset}

\begin{figure}[ht]
\begin{center}
\includegraphics[width=\linewidth]{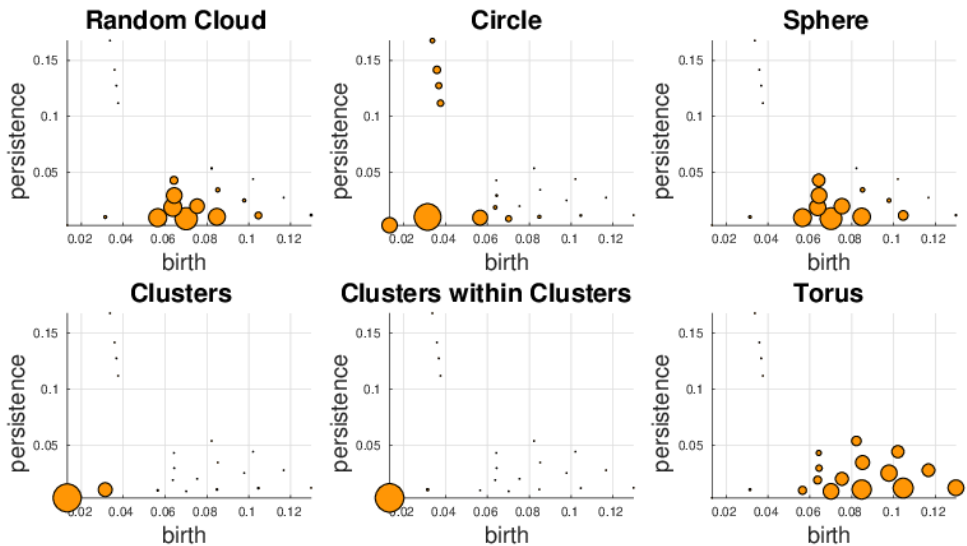}
\caption{Average codebook histograms computed for each of the six shape classes of the~synthetic dataset. The cluster center of each codeword is presented as a circle in the birth-persistence domain. The area of the circles reflects the histogram values of the specific class. For all classes, the same codebook (same clustering) is employed; thus, dot locations are the same on all plots. The differences between the circles reflect the class differences.}
\label{fig:bow_assignment}
\end{center}
\end{figure}

We compute PBoW with $N=20$ clusters for the synthetic dataset and visually analyze the codeword histograms obtained by (hard) assignment. To this end, for each of the six shape classes, we compute the average codebook histogram (over all samples of each class) to obtain one representative PBoW vector per class. The averaged PBoW histograms for all classes are presented in Fig.~\ref{fig:bow_assignment}. Instead of only providing the histograms themselves, for each codeword of the histogram we plot the corresponding cluster center as a circle in the original birth-persistence domain and encode the number of assigned codeworks (the actual values of the histograms) in the area of the circles, i.e. the larger the count for a cluster, the larger the circle. The advantage of this representation is that the spatial distribution of the codewords in the PD is preserved. 

From Fig.~\ref{fig:bow_assignment} we can see that, except for the classes ``random cloud" and ``sphere" (which are difficult to differentiate), all classes generate strongly different cluster distributions. Class ``circle", for example, uniquely activates four clusters with strong persistence (top-left corner) and the ``torus" class distributes its corresponding code words across a large number of clusters representing less persistent components. 

Fig.~\ref{fig:bow_assignment} further illustrates an important property of persistence bag-of-words, namely its sparse nature. More specifically, areas with no points in the consolidated persistence diagram will contain no codewords (clusters). In Fig.~\ref{fig:bow_assignment}, for example, no codeword is obtained in the upper-right quadrant of the diagram, since no components are located there for the~underlying data. Therefore, these unimportant areas are neglected and not encoded into the~final representation. This not only reduces the dimension of the final representation, but further makes the representation adaptive to the underlying data. This in turn increases the information density in the obtained representation.

\subsubsection{GeoMat dataset}
\label{subsec:resultsPBoWforGeomat}

\begin{figure}
\centering
\includegraphics[width=0.4\columnwidth]{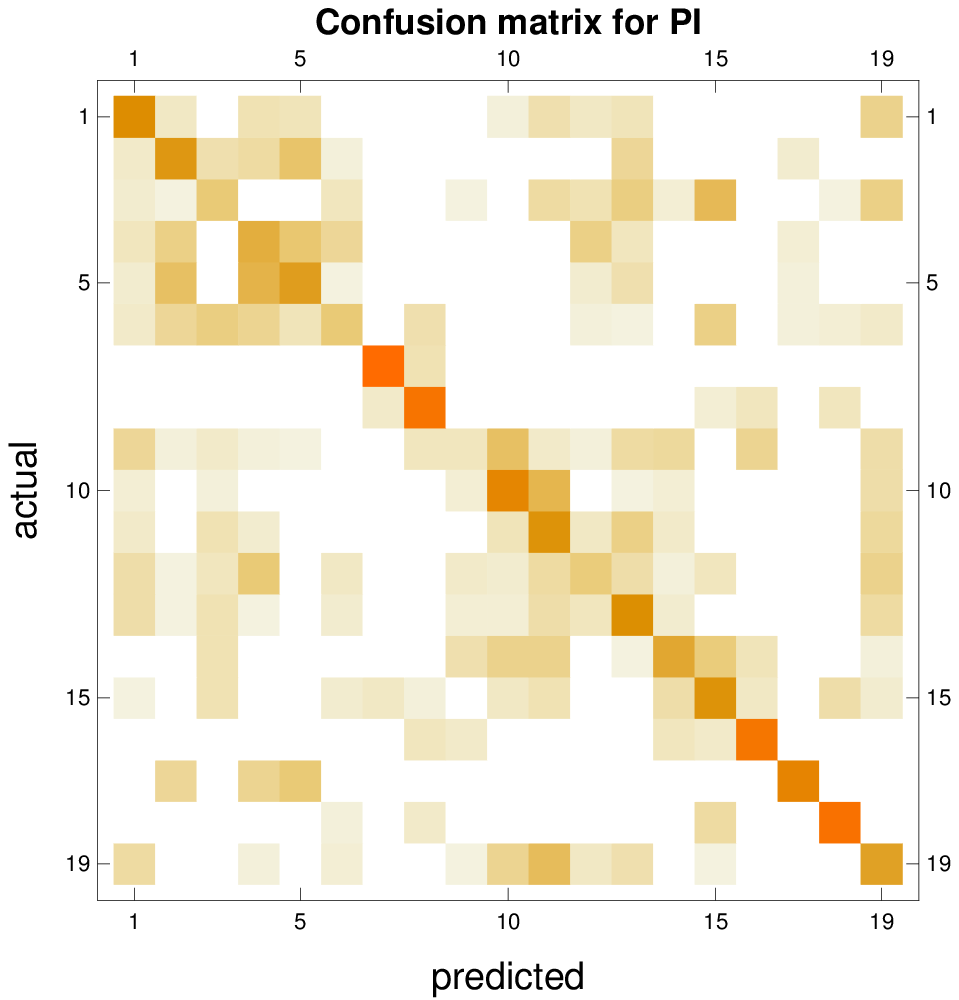}
\includegraphics[width=0.4\columnwidth]{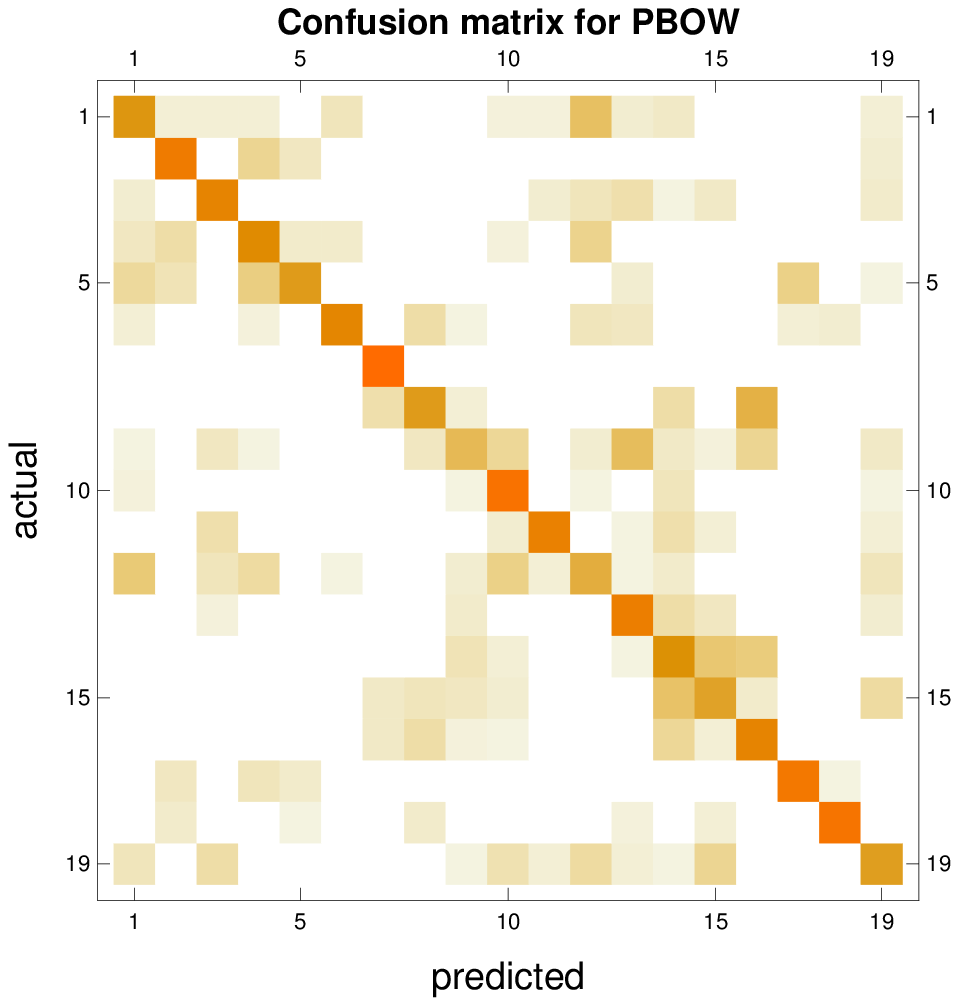}
\caption{Confusion matrix for PI (left) and PBoW (right) on the GeoMat dataset from EXP-A. From the diagonal of the matrices we can see that PBoW outperforms PI for many classes (e.g. classes 2-5, 9 and 12). Furthermore, there are less confusions (off-diagonal values) for PBoW.}
\label{fig:conf_mat_exp3a}
\end{figure}

\begin{figure}
\begin{center}
\includegraphics[width=1.\linewidth]{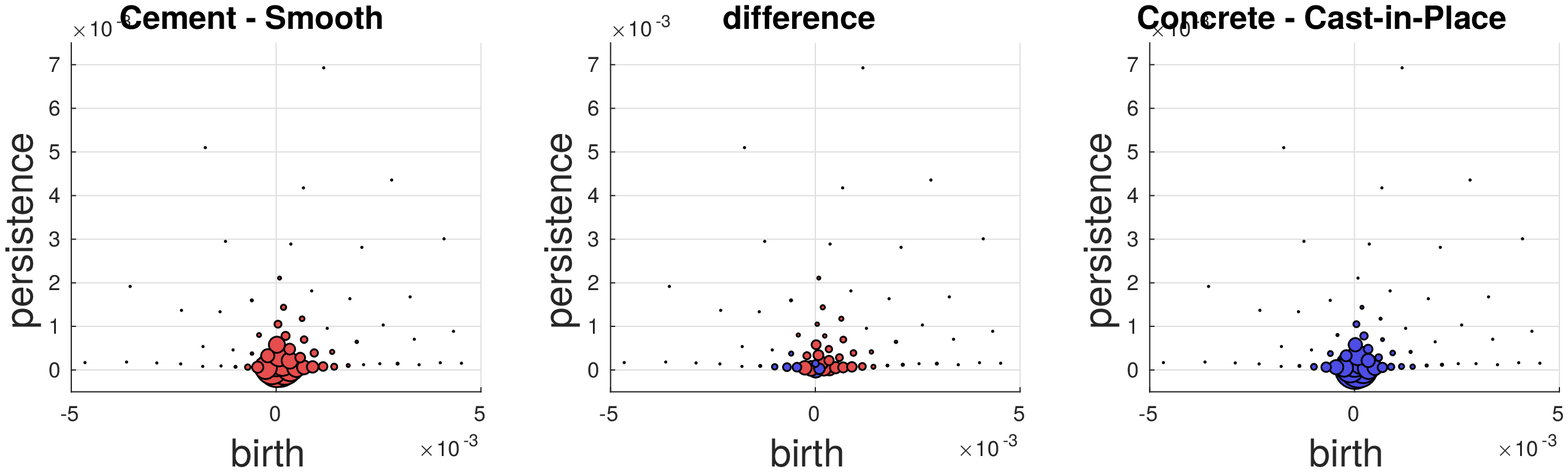}
\includegraphics[width=1.\linewidth]{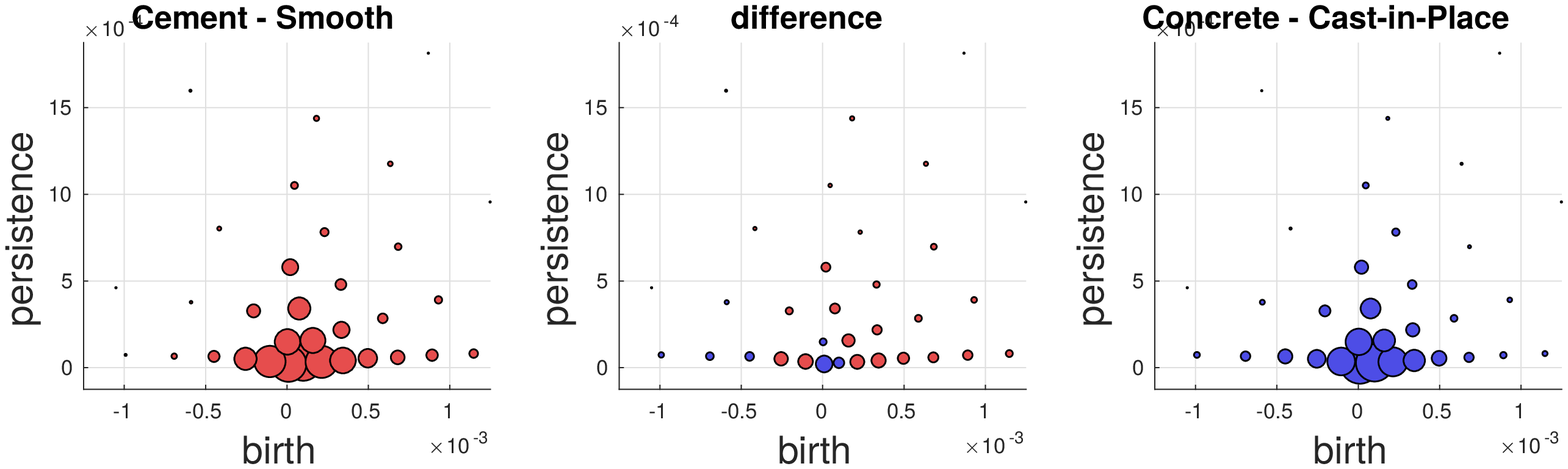}
\caption{Comparison of averaged PBoW histograms for class ``cement smooth'' (left, red) and ``concrete cast-in-place'' (right, blue) from GeoMat dataset (top row: total view; bottom row is zoomed in). The plot in the center shows the difference between the classes, where red color means that the left class has stronger support for this cluster and blue means that the right class has stronger support. The classes differ by fine-grained spatial differences, which are not distinguishable in other vectorized representations.}
\label{fig:bow_assignment_diff_exp3a_4vs5}
\end{center}
\end{figure}

\begin{figure}
\begin{center}
\includegraphics[width=1.\linewidth]{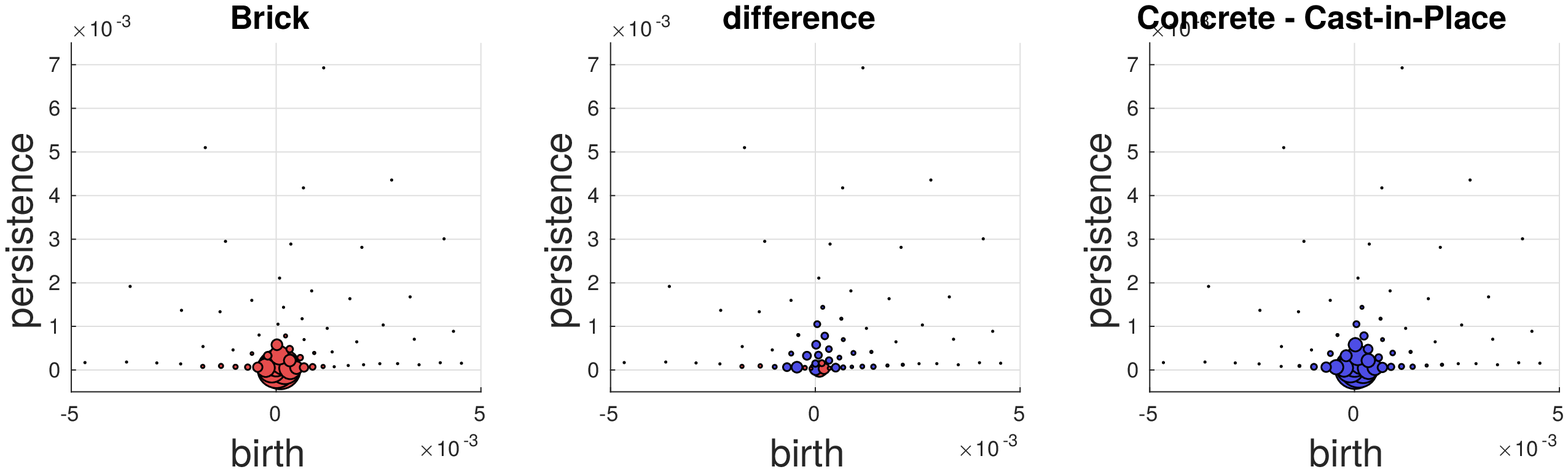}
\includegraphics[width=1.\linewidth]{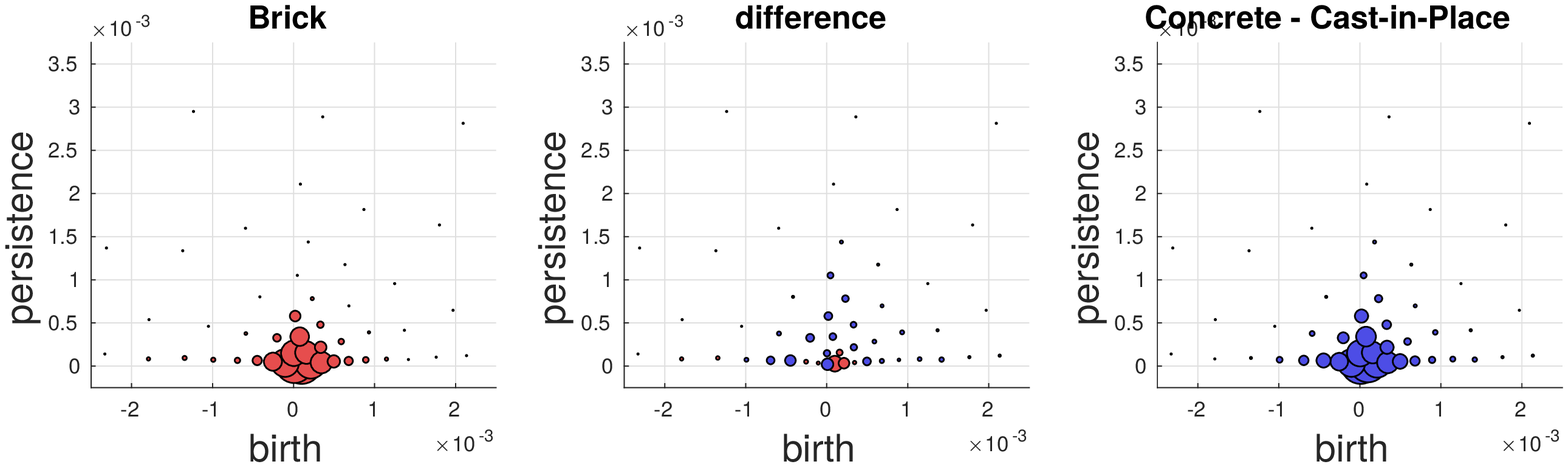}
\includegraphics[width=1.\linewidth]{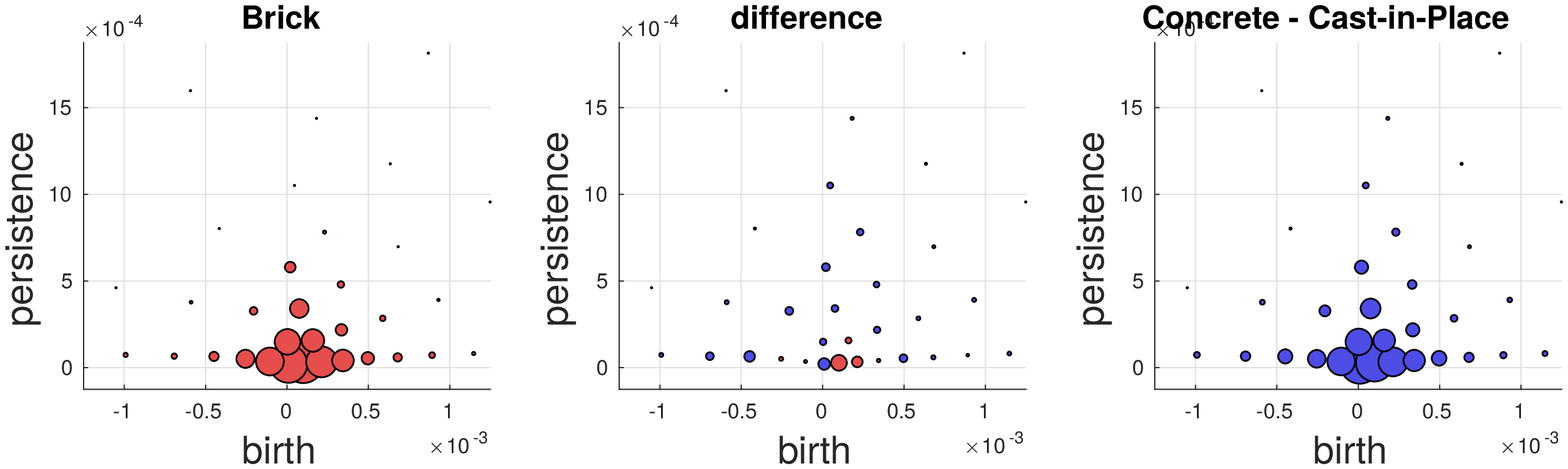}
\caption{Comparison of averaged PBoW histograms for classes ``brick'' (left, red) and ``concrete cast-in-place'' (right, blue) from GeoMat dataset (top row: total view; 2nd row: zoomed in view; 3rd row: even further zoomed in view). The plot in the center shows the difference between the classes, where red color means that the left class has stronger support for this cluster and blue means that the right class has stronger support. The classes differ by fine-grained spatial differences, which are not distinguishable in other vectorized representations.}
\label{fig:bow_assignment_diff_exp3a_2vs5}
\end{center}
\end{figure}

We further investigate the performance on the GeoMat dataset to explain why (s)PBoW outperforms PI and RM by such a large margin (see Table~\ref{tab:kernelResults}). To this end, we generate confusion matrices for PI and PBoW (see Fig.~\ref{fig:conf_mat_exp3a}) to investigate their discriminative abilities. The matrices show that PBoW, for example, achieves better discrimination between classes ``cement smooth'' and ``concrete cast-in-place'' (i.e. classes $4$ and $5$). Average PBoW histograms for those classes are shown in Fig.~\ref{fig:bow_assignment_diff_exp3a_4vs5}. The histograms are on the first sight similar (upper row in Fig.~\ref{fig:bow_assignment_diff_exp3a_4vs5}). However, by zooming-in towards the birth-persistence plane in Fig.~\ref{fig:bow_assignment_diff_exp3a_4vs5} (bottom row), differences become better visible. The plots in the center illustrate the difference between the class distributions (red color means left class is stronger, blue means right class is stronger for this cluster). The classes differ by fine-grained spatial differences. The set of three blue points around birth time of 0 (which are characteristic for class ``concrete cast-in-place'') surrounded by red points (which are characteristic for class ``cement smooth") illustrates this well (see lower central plot). For the discrimination of these two classes, a particularly fine-grained codebook with many clusters is needed. The PI has problems with such fine-grained structures, because due to its limited resolution, all topological components in the most discriminative area would most likely fall into one PI-pixel. Therefore, an extraordinary high resolution would be necessary to capture the discriminative patterns between those two classes. The bag-of-words model makes our approaches independent of the resolution and enables to capture even fine differences adaptively and in an unsupervised way. In Fig.~\ref{fig:bow_assignment_diff_exp3a_2vs5} we show a similar comparison for classes ``brick'' and ``concrete cast-in-place'' (i.e. classes $2$ and $5$).

\section{Conclusion}
\label{sec:conclusion}

We have introduced the concept of persistence codebooks, a novel fixed-length vectorial representation for persistence diagrams. Persistence codebooks employ bag-of-words encodings to quantize the persistence diagram into a vectorized representation. We propose different types of encodings (based on traditional bag-of-words, VLAD and Fisher Vectors), investigate their theoretic properties, such as their stability with respect to 1-Wasserstein, and introduce robust variants of the representations. Experiments on seven heterogeneous datasets show that the novel representations achieve state-of-the-art, and partly even better, performance while requiring significantly less computation time. The novel representations have both attractive theoretic properties as well as practical properties, i.e compactness, expressiveness, as well as the ability to adapt to the inherent sparsity of persistence diagrams. They can be constructed in a completely unsupervised fashion and achieve a high discriminativity compared to related approaches. The high computational efficiency of persistence codebooks could in future facilitate the application of TDA to larger datasets than possible today and enable real-time applications.

\section*{Acknowledgements}

The work presented in this article was supported by the National Science Centre, Poland under grant no. 2015/19/D/ST6/01215, by the Austrian Research Promotion Agency (FFG) under project no. 856333, and by EPSRC grant no. EP/R018472/1.

\section*{Appendix: Background on Persistent Homology}
\label{persistence}

In this section we present basic introduction to persistent homology. Please consult ~\citep{edelsbrunner2010computational,EdLeZo2002,zomorodian2005computing} for more information. 

Topological spaces are typically infinite objects and, for the sake of data analysis, they have to be finitely represented by simplified objects called \emph{cell complexes}. Cell complexes are build from \emph{cells}: topologically simple objects having the property that an intersection of every pair of cells is either empty, or contains yet another cell in the cell complex.

A \emph{simplicial complex} is a particular instance of a general cell complex. It is a natural tool in the study of multi-dimensional point cloud data. Cells of simplicial complex are called \emph{simplices} and, in this particular case, are formed with convex hulls of collections of nearby points in the point cloud. Simplices are uniquely characterized by a collection of points involved in their convex hulls. A simplicial complex $\mathcal{X}$ needs to satisfy the following property: for every pair of simplices $\sigma, \tau \in \mathcal{X}$, $\sigma \cap \tau$ is either empty or a simplex in $\mathcal{X}$. 

Given a point cloud $X$ with a distance or a similarity measure $d$ and a parameter $r > 0$, one can define a {\em Vietoris-Rips complex} $VR(X,r)$. It is a simplicial complex whose every simpliex $\sigma = \{ v_0, v_1, \ldots, v_k \}$ satisfies $d(v_i, v_j) \le r$ for every $i,j \in \{ 0,\ldots,k \}$. 
For every simplex $\sigma \in VR(X,r)$, one can define a diameter of $\sigma$ being the largest distance between the points in $\sigma$. This gives a natural ordering of simplices in $VR(X,r)$: primarily by diameter of simplices and secondarily (when diameters of two simplices are the same) by inverse of the number of points in simplices\footnote{A number of points involved in the simplex minus one is a \emph{dimension} of the simplex.}. It is easy to see that every prefix of such an ordering forms a simplicial complex, and therefore any increasing sequence of numbers $0 < r_1 < r_2 < \ldots < r_n$ yields a nested sequence of simplicial complexes:
\begin{align*}
\emptyset \subset X=VR(X,0)\subset VR(X,r_1) \subset \\     VR(X,r_2) \subset \cdots\subset VR(X,r_n)
\end{align*}

Another typical scenario when such a nested sequence of cell complexes arises is the~case of values of a function $f$ discretized on a grid $G$. The function $f : G \rightarrow \mathbb{R}$ is typically an output of some numerical method. The grid $G$ naturally corresponds to cubical complex $\mathcal{G}$, and the function $f$ provides an ordering of maximal cubes in the complex. This ordering induces a nested sequence of cubical complex, very much like a nested sequence of Vietoris-Rips complexes discussed above. 

To cover those and other possible cases, later in this section we will focus on a general case of filtered cell complex:
\[
\emptyset = \mathcal{C}_0 \subset \mathcal{C}_1 \subset \cdots \subset \mathcal{C}_n = \mathcal{C},
\]
keeping in mind that most typically it will come from a point cloud, or numerical simulations on a grid. 

Having a complex $\mathcal{C}_i$ in the filtration, one can define its homology, $H(\mathcal{C}_i)$. Rather than providing a formal definition, which can be found in~\citep{edelsbrunner2010computational}, we will focus on the intuitive understanding of the concept. Homology in dimension $0$ measures number of connected components. In dimension $1$ it measures the cycles, which do not bound to a (deformed) surface. In dimension $2$ it corresponds to voids, i.e. regions of space totally bounded by a collection of triangles (very much like a ball bounds the void inside it). The idea of a cycle bounding a hole in the complex can be formalized using homology theory for arbitrary dimension. 

\begin{figure}[ht]
\centering
\includegraphics[width=0.8\linewidth]{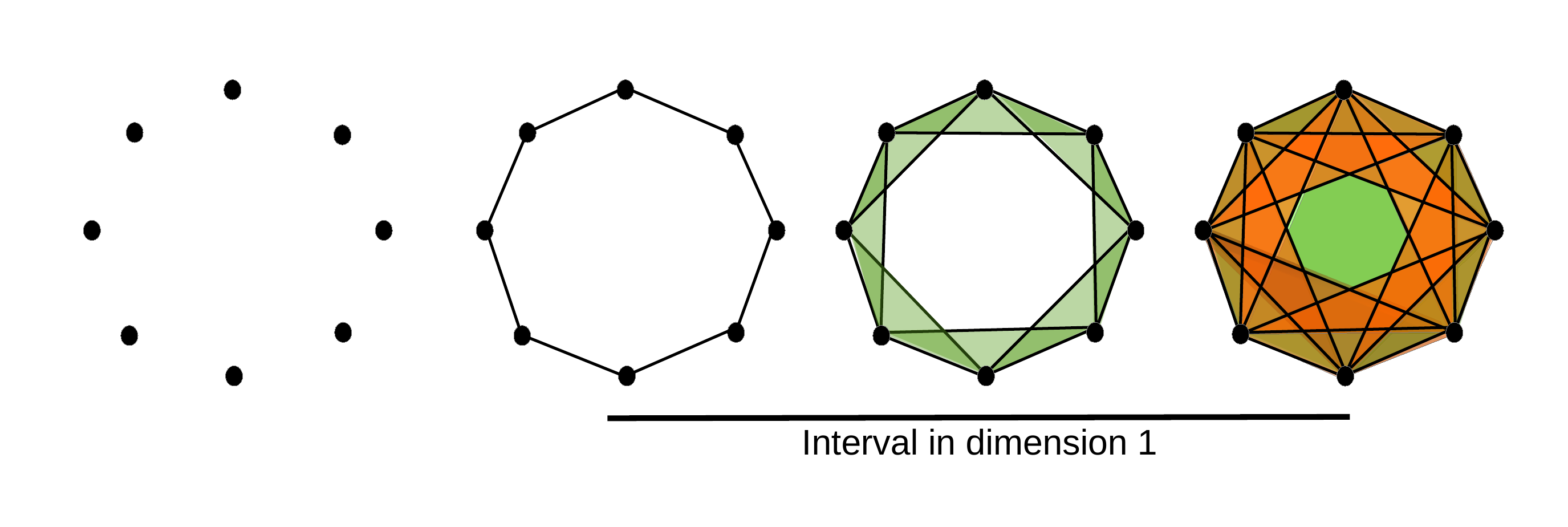}
\caption{Various stages of a construction of a Vietoris-Rips complex for eight points sampled from a circle. Initially, for sufficiently small radius, only vertices are present in the complex. Gradually, more and more edges along with higher dimensional simplices of an increasing diameter are added. In all but the initial and final stage of the construction the topology of a circle is visible, and therefore will be recovered by PH in dimension one (depicted by the long bar below the picture).}
\label{fig:persistence_llustration}
\end{figure}

Persistent homology measures the evolution of homology for the constitutive complexes in filtration. Once more and more cells are being added to a complex $\mathcal{C}_i$, new connected components or cycles may appear, old ones may become trivial or become identical (homologous) to others created earlier. For every connected component or a cycle, there are two important characteristics we will store: the first moment $b$, referred to as a \emph{birth time}, when it appears in the filtration, and the last moment $d$, referred to as \emph{death time}, when it either becomes trivial or becomes identical to other cycle created earlier. In this paper, instead of a standard birth-death summaries of persistent homology, we use birth-persistence coordinates, which can be obtained by the $[b,d] \rightarrow [b,d-b]$ transformation. The basic geometrical idea behind PH is presented in Fig.~\ref{fig:persistence_llustration}.

A couple of assumptions about PDs are made. Firstly, as our aim is to perform computations, we assume that persistence diagrams consist of finitely many points of nonzero persistence. Secondly, PDs may also contain infinite intervals that correspond to the so-called \emph{essential classes}, i.e. the cycles that are born but never die. Those infinite intervals need to be processed prior to the computations. There are at least three strategies one can apply:
\begin{enumerate}
\item to ignore infinite intervals and use only the finite ones for consideration;
\item to substitute $+\infty$ in the death coordinates of the essential classes with a number $N$ chosen by the user (a logical choice would be a number which is larger than a filtration value of any cell in the considered complex);
\item to build a pair of descriptors: one for finite, and one for infinite intervals and use them together as a final descriptor. 
\end{enumerate}
Given the available options, in the numerical experiments presented in this paper, we have chosen the simplest one, i.e. to ignore the infinite intervals. 
There are various classical metrics used to compare persistence diagrams~\citep{edelsbrunner2010computational}. We will review them here, as they are essential in the study of stability of the presented representations. Note that the presentation is a bit non standard, as we are working on birth-peristence coordinates. Given two diagrams $B$ and $B'$, we construct a matching $\eta : B \rightarrow B'$ assuming that points can also be matched to $y=0$ axis. Putting $B$ and $B'$ in the same diagram, one can visualize a matching $\eta$ by drawing a line segment between $x \in B$ and $\eta(x)$ (note that $\eta(x)$ is either in $B'$, or is a projection of $x$ to its first coordinate). Given all the line segments, for each matching we can store the longest one, or a sum of lengths of all of them (raised to power $q$). Taking the minimum over all possible matchings of the obtained numbers will yield the \emph{bottleneck} distance in the first case, and the \emph{Wasserstein} distance (raised to power $\frac{1}{q}$) in the second case. More formally:

\medskip
\noindent
{\bf Definition} {\it q-Wasserstein distance} between two persistence diagrams $B,B'\in\D$ is defined as:
\[
W_q(B,B') := \left[\inf_{\eta: B\rightarrow B'} \sum_{x\in B} \norm{x-\eta(x)}^q_\infty\right]^\frac{1}{q}.
\]
In particular:
\[
W_1(B,B') := \inf_{\eta: B\rightarrow B'} \sum_{x\in B} \norm{x-\eta(x)}_\infty.
\]

An important feature of persistent homology is its \emph{stability}. Intuitively, it indicates that small changes in the filtration imply small changes (for instance in Wasserstein metric) in the resulting persistence diagrams. Formally:

\medskip
\noindent {\bf Theorem} \citep{edelsbrunner2010computational} Let $\mathbb{X}$ be a finite cell complex and $f,g : \mathbb{X} \rightarrow \mathbb{R}$ filtering Lipshitz functions. Let $B$ and $B'$ be the PDs of $\mathbb{X}$ with filtration induced by $f$ and $g$ respectively. Then there exist constants $C$ and $k$ such that $W_1(B,B') \leq C || f-g ||_{\infty}^{1-k}$

\medskip

In this paper, we show stability with respect to 1-Wasserstein distance. Combined with the stability result described above, this indicates stability of bag-of-words representations with respect to the perturbation of initial data.

{\small
\bibliographystyle{ieee}
\bibliography{egbib}
}

\end{document}